\documentclass{article}

\usepackage[nonatbib, final]{neurips_2021}

\usepackage[utf8]{inputenc} 
\usepackage[T1]{fontenc}    
\usepackage{hyperref}       
\usepackage{url}  
\usepackage[utf8]{inputenc}
\usepackage{amsmath}
\usepackage{graphicx}
\usepackage{rotating}
\usepackage{tabulary}
\usepackage{algorithm}
\usepackage{algorithmic}
\usepackage[numbers]{natbib}
\usepackage{todonotes}
\usepackage{hyperref}
\usepackage{enumitem}
\usepackage{multirow}
\usepackage[thmmarks, thref, amsmath]{ntheorem}
\newtheorem*{lemma-nono}{Lemma}
\newtheorem*{theorem-nono}{Theorem}
\theoremstyle{plain}

\theoremheaderfont{\itshape}
\theorembodyfont{\upshape}
\theoremstyle{plain}

\theoremstyle{nonumberplain}
\theoremheaderfont{\scshape}
\theorembodyfont{\upshape}
\newcommand{\figref}[1]{Fig.~\ref{#1}}
\newcommand{\tabref}[1]{Tab.~\ref{#1}}
\newcommand{\secref}[1]{Sec.~\ref{#1}}
\newcommand{\AlgRef}[1]{Algorithm~\ref{#1}}
\newcommand{\equref}[1]{Equ.~(\ref{#1})}
\newcommand{\Appref}[1]{Appendix.~\ref{#1}}
\usepackage{minitoc}
\usepackage{subdef}
\usepackage{wrapfig,lipsum,booktabs}
\usepackage{subcaption}
\usepackage{microtype}
\theorempostwork{\setcounter{case}{0}}

\title{SIMILAR: Submodular Information Measures Based Active Learning In Realistic Scenarios}

\author{Suraj Kothawade \\ University of Texas at Dallas \\ \texttt{suraj.kothawade@utdallas.edu}
\And
Nathan Beck \\ University of Texas at Dallas \\ \texttt{nathan.beck@utdallas.edu}
\And
Krishnateja Killamsetty \\ University of Texas at Dallas \\ \texttt{krishnateja.killamsetty@utdallas.edu}
\And
Rishabh Iyer \\ University of Texas at Dallas \\ \texttt{rishabh.iyer@utdallas.edu}
}

\begin{document}
\maketitle
\doparttoc
\faketableofcontents

\begin{abstract}
    Active learning has proven to be useful for minimizing labeling costs by selecting the most informative samples. However, existing active learning methods do not work well in realistic scenarios such as imbalance or rare classes, out-of-distribution data in the unlabeled set, and redundancy. In this work, we propose \textsc{Similar} (\textbf{S}ubmodular \textbf{I}nformation \textbf{M}easures based act\textbf{I}ve \textbf{L}e\textbf{AR}ning), a unified active learning framework using recently proposed submodular information measures (SIM) as acquisition functions.  We argue that \textsc{Similar} not only works in standard active learning but also easily extends to the realistic settings considered above and acts as a \textit{one-stop} solution for active learning that is scalable to large real-world datasets. 
    Empirically, we show that \textsc{Similar} significantly outperforms existing active learning algorithms by as much as $\approx 5\% - 18\%$ in the case of rare classes and $\approx 5\% - 10\%$ in the case of out-of-distribution data on several image classification tasks like CIFAR-10, MNIST, and ImageNet. \textsc{Similar} is available as a part of the DISTIL toolkit: \url{https://github.com/decile-team/distil}.\looseness-1 
\end{abstract}

\section{Introduction}

\begin{wrapfigure}{R}{0.50\textwidth}
\includegraphics[width=0.50\textwidth]{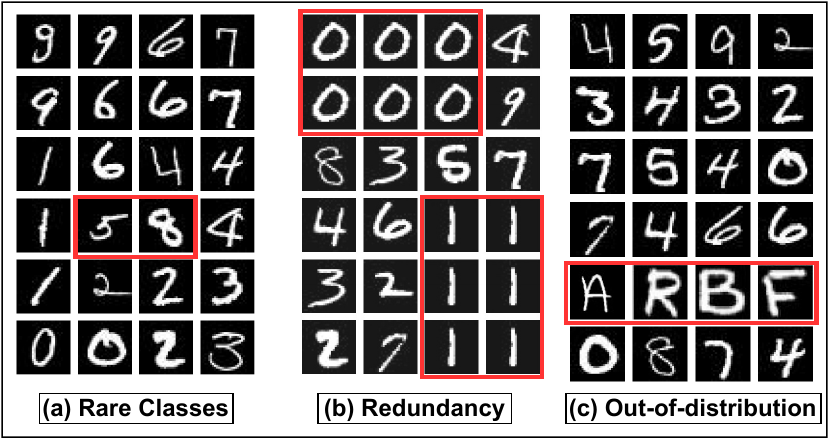}
\caption{Motivating scenarios for realistic active learning: (a) rare classes: digits 5 and 8 are rare; (b) redundancy: digits 0 and 1 are redundant; (c) out-of-distribution (OOD): letters A, R, B, F in digit classification. 
}
\label{fig:intro}
\end{wrapfigure}

Deep neural networks (DNNs) have had a lot of success in a wide variety of domains. However, they require large labeled datasets which are often taxing, time-consuming, and expensive to obtain. Active learning (AL) \cite{fine2002query, freund1997selective, schohn2000less, ash2019deep, campbell2000query} is a promising approach to solve this problem. It aims to select the most informative data points from an unlabeled dataset to be labeled in an adaptive manner with a human in the loop. The goal of AL is to achieve maximum accuracy of the model while minimizing the number of data points required to be labeled. 


Current AL methods have been tested in relatively simple, clean, and balanced datasets. However, real-world datasets are not clean and have a number of characteristics that makes learning from them challenging~\cite{chelba2013one, zhang2015character, zhu2015aligning, russakovsky2015imagenet, abu2016youtube, caesar2020nuscenes}. Firstly, these real-world datasets are \textit{imbalanced}, and some classes are \textit{very rare} (e.g., Fig~\ref{fig:intro}(a)). Examples of this imbalance are medical imaging domains where the \emph{cancerous} images are rare. 
Secondly, real-world data has a lot of \textit{redundancy} (e.g., Fig~\ref{fig:intro}(b)). This redundancy is more prominent in datasets that are created by sampling frames from videos (e.g., footage from a car driving on a freeway or surveillance camera footage). Thirdly, it is common to have \textit{out-of-distribution} (OOD) (e.g., Fig~\ref{fig:intro}(c)) data, where some part of the unlabeled data is not of concern to the task at hand. Given the amount of unlabeled data, it is not realistic to assume that these datasets can be cleaned manually; hence, it is the need of the hour to have active learning methods that are robust to such scenarios. We show that current AL approaches (including the state-of-the-art approach \textsc{Badge}~\cite{ash2019deep}) do not work well in the presence of the dataset biases described above. 
In this work, we address the following question: \textit{Can a machine learning model be trained using a single unified active learning framework that works for a broad spectrum of realistic scenarios?} As a solution, we propose \textsc{Similar}\footnote{\textbf{S}ubmodular \textbf{I}nformation \textbf{M}easures based act\textbf{I}ve \textbf{L}e\textbf{AR}ning}, a unified active learning framework which enables active learning for many realistic scenarios like rare classes, out-of-distribution (OOD) data, and redundancy.

\subsection{Related Work} \label{sec:relwork}
Active learning has enabled efficient training of complex deep neural networks by decreasing labeling costs. The most commonly used approach is to select the most uncertain items. Examples of uncertainty strategies include \textsc{Entropy}~\cite{settles2009active}, \textsc{Least Confidence}~\cite{wang2014new}, and \textsc{Margin}~\cite{roth2006margin}. One challenge of this approach is that all the samples within a batch can be potentially similar even though they are uncertain. 
To overcome this problem in batch active learning, many recent works have attempted to select diverse yet informative data points. \cite{wei2015submodularity, kaushal2019learning}  propose a simple approach: Filter a set of points using uncertainty sampling and then select a diverse subset from the filtered set. \cite{sener2017active} propose \textsc{Coreset}, which forms core-sets using greedy $k$-center clustering while maintaining the geometric arrangement. \textsc{Badge}~\cite{ash2019deep}, another recent approach, proposes to select data points corresponding to high-magnitude, diverse hypothesized gradients by using \textsc{k-means++} \cite{kmeansplus} initialization to distance from previously selected data points in the batch. Most existing AL approaches fail to ensure diversity across AL selection rounds and do not perform as well when there is a lot of redundancy. \citeauthor{sinha2019variational}~\cite{sinha2019variational} used a variational autoencoder (VAE)~\cite{kingma2013auto} to learn a feature space and an adversarial network~\cite{makhzani2015adversarial} to distinguish between labeled and unlabeled data points. However, their approach is computationally expensive and requires extensive hyperparameter tuning. Similarly, \textsc{BatchBALD}~\cite{kirsch2019batchbald} does not scale to larger batch sizes since their method would need a large number of Monte Carlo dropout samples to obtain a significant mutual information. Such limitations reduce the scope of applying these methods to realistic settings.\looseness-1 

Closely related to our work are two recently proposed works. The first is \textsc{Glister-Active}~\cite{killamsetty2020glister}, which formulates the AL acquisition function by maximizing the log-likelihood on a held-out validation set. This validation set could consist of examples from the rare classes or in-distribution examples. 
The second approach is the work of \citeauthor{gudovskiy2020deep}~\cite{gudovskiy2020deep}, who study AL for biased datasets using a self-supervised \textsc{Fisher} kernel and pseudo-label estimators. They address this problem by explicitly minimizing the KL divergence between training and validation sets via maximizing the \textsc{Fisher} kernel. Although their method shows promising results, they make multiple unrealistic assumptions: a) They use a \textit{large labeled validation set}, 
and b) they use feature representations from a model pretrained using unsupervised learning on a \textit{balanced} unlabeled dataset. In this work, we compare against both \textsc{Glister-Active} \cite{killamsetty2020glister} and \textsc{Fisher} \cite{gudovskiy2020deep} approaches in the more realistic setting of a small held-out validation set (smaller than the seed labeled set) and an imbalanced unlabeled set. Another work proposed a discrete optimization method for $k$-NN-type algorithms in the domain shift setting~\cite{berlind2015active}. However, their approach is limited to $k$-NNs.

This work utilizes submodular information measures (SIM) by~\cite{iyer2021submodular} and their extensions by~\cite{kaushal2021prism}. SIMs encompass submodular conditional mutual information (SCMI), which can then be used to derive submodular mutual information (SMI); submodular conditional gain (SCG); and submodular functions (SF). We discuss these functions in detail in \secref{sec:background}. \cite{kaushal2021prism} also studies these functions on the closely related problem of targeted data selection.\looseness-1

\subsection{Our Contributions}
The following are our main contributions: \textbf{1)} Given the limitations of existing approaches in handling active learning in the real world, we propose \textsc{Similar} (\secref{sec:unifiedAL}), a unified active learning framework that can serve as a comprehensive solution to multiple realistic scenarios. \textbf{2)} We treat SIM as a common umbrella for realistic active learning and study the effect of different function instantiations offered under SIM for various realistic scenarios. \textbf{3)} \textsc{Similar} not only handles standard active learning but also extends to a wide range of settings which appear in the real world such as rare classes, out-of-distribution (OOD) data, and datasets with a lot of redundancy. 
Finally, \textbf{4)} we empirically demonstrate the effectiveness of SMI-based measures for image classification (\secref{sec:exp}) in a number of realistic data settings including imbalanced, out-of-distribution, and redundant data. Specifically, in the case of imbalanced and OOD data, we show that \textsc{Similar} achieves improvements of more than 5 to 10\% on several image classification datasets.\looseness-1



\section{Background} \label{sec:background}
In this section, we enumerate the different submodular functions that are covered under SIM and the relationships between them. 

\noindent \textbf{Submodular Functions.}
 We let $\Ucal$ denote the \emph{unlabeled} set of $n$ data points $\Ucal = \{1, 2, 3,...,n \}$ and a set function $f:
 2^{\Ucal} \xrightarrow{} \mathbb{R}$. Formally, a function $f$ is submodular \cite{fujishige2005submodular} if for $x \in \Ucal$, $f(\Acal \cup x) - f(\Acal )\geq f(\Bcal \cup x) - f(\Bcal)$, $\forall \Acal \subseteq \Bcal \subseteq \Ucal$ and $x \notin \Bcal$. For a set $\Acal \subseteq \Ucal$, $f(\Acal)$ provides a real-valued score for $\Acal$. In the context of batch active learning, this is the score of an acquisition function $f$ on batch $\Acal$.  
 Submodularity is particularly appealing because it naturally occurs in real world applications~\cite{tohidi2020submodularity,bach2011learning,bach2019submodular,iyer2015submodular} and also admits a constant factor $1-\frac{1}{e}$~\cite{nemhauser1978analysis} for cardinality constraint maximization. 
 Additionally, variants of the greedy algorithm maximize a submodular function in \textit{near-linear time}~\cite{mirzasoleiman2015lazier}.\looseness-1 

\noindent \textbf{Submodular Mutual Information (SMI).}
Given sets $\Acal, \Qcal \subseteq \Ucal$, the SMI \cite{levin2020online,iyer2021submodular} is defined as $I_f(\Acal; \Qcal) = f(\Acal) + f(\Qcal) - f(\Acal \cup \Qcal)$. Intuitively, \textit{SMI models the similarity between $\Qcal$ and $\Acal$}, and maximizing SMI will select points \emph{similar} to $\Qcal$ while being diverse. $\Qcal$ here is the query set.

\noindent \textbf{Submodular Conditional Gain (SCG).} Given sets $\Acal, \Pcal \subseteq \Ucal$, the SCG $f(\Acal | \Pcal)$ is the gain in function value by adding $\Acal$ to $\Pcal$. Thus, $f(\Acal | \Pcal) = f(\Acal \cup \Pcal) - f(\Pcal)$ \cite{iyer2021submodular}. Intuitively, SCG models how different $\Acal$ is from $\Pcal$, and maximizing SCG functions will select data points \emph{not similar to the points in $\Pcal$} while being diverse. We refer to $\Pcal$ as the conditioning set.

\noindent \textbf{Submodular Conditional Mutual Information (SCMI).} 
Given sets $\Acal, \Qcal, \Pcal \subseteq \Ucal$, the SCMI is defined as $I_f(\Acal; \Qcal | \Pcal) = f(\Acal \cup \Pcal) + f(\Qcal \cup \Pcal) - f(\Acal \cup \Qcal \cup \Pcal) - f(\Pcal)$. Intuitively, SCMI \emph{jointly models the similarity between $\Acal$ and $\Qcal$ and their dissimilarity with $\Pcal$}. 

\begin{wraptable}{r}{7.5cm}
\small{
\begin{tabular}{|l|l|l|}
\hline
\textbf{Function} & \textbf{Setting} & \textbf{Realistic Scenario} \\ \hline
Submodular        & $\Qcal \leftarrow \Ucal, \Pcal \leftarrow \emptyset$                & Standard AL                 \\
SMI               & $\Qcal \leftarrow \Qcal, \Pcal \leftarrow \emptyset$                & Imbalance, OOD              \\
SCG               & $\Qcal \leftarrow \emptyset, \Pcal \leftarrow \Pcal$                & Redundancy                  \\
SCMI              & $\Qcal \leftarrow \Qcal, \Pcal \leftarrow \Pcal$                & OOD                      \\ \hline  
\end{tabular}
\caption{Relationship between SIM and their applications to realistic scenarios by choices of $\Qcal$ and $\Pcal$.}
\vspace{-2ex}
\label{tab:relSIM}}
\end{wraptable}

\noindent \textbf{Relationship between SIM}
The relationship between the above measures is the key component that unifies our AL framework~\cite{iyer2021submodular,kaushal2021prism}. The unification comes from the rich modeling capacity of SCMI: $I_f(\Acal; \Qcal | \Pcal)$ where $\Qcal, \Pcal \subseteq \Ucal$. This facilitates a \textit{single} acquisition function that can be applied to \textit{multiple} scenarios. Concretely, the submodular function $f$ can be obtained by setting $\Qcal \leftarrow \Ucal$ and $\Pcal \leftarrow \emptyset$. 
Next, the SMI can be obtained by setting $\Qcal \leftarrow \Qcal$ and $\Pcal \leftarrow \emptyset$, while we obtain SCG by setting $\Qcal \leftarrow \emptyset$, $\Pcal \leftarrow \Pcal$.   We summarize the relationships between SIM in \tabref{tab:relSIM}.\looseness-1  

\noindent \textbf{Instantiations of SIM.}
The formulations for Facility Location (\textsc{Fl}), Graph Cut (\textsc{Gc}) and Log Determinant (\textsc{Logdet}) are as in \cite{iyer2021submodular, kaushal2021prism} and we adapt them as acquisition functions for batch active learning.
We use two variants for \textsc{Fl}: \textsc{Flqmi}, which models pairwise similarities of \textit{only the query set} $\Qcal$ to the unlabeled dataset, and \textsc{Flvmi}, which additionally considers the pairwise similarities within the unlabeled dataset $\Ucal$. The SCG and SCMI expressions corresponding to \textsc{Fl} are referred as \textsc{Flcg} and \textsc{Flcmi}, respectively (see row 1 in \tabref{tab:smi_inst} and 2b). For \textsc{LogDet}, we refer to the SMI, SCG and SCMI expressions as \textsc{Logdetmi}, \textsc{Logdetcg} and \textsc{Logdetcmi}, respectively (see row 5 in \tabref{tab:smi_inst} and row 2 in Tab. 2b). Similarly, the SMI and SCG expressions are respectively referred to as \textsc{Gcmi} and \textsc{Gccg} for \textsc{Gc} (see row 3 in \tabref{tab:smi_inst} and 2b). For notation in Tab. 2, the pairwise similarity matrix $S$ between items in sets $\Acal$ and $\Bcal$ is denoted as $S_{\Acal, \Bcal}$. Also, we denote $S_{ij}$ as the $(i,j)$ entry of $S$.\looseness-1 
\vspace{-1ex}

\begin{table}[!htb]
    \caption{Instantiations of SIM. Note how the relationships in
    \tabref{tab:relSIM} can be applied to SCMI instantiations to obtain SMI and SCG instantiations.
    }
    \label{tab:SIM_inst}
    \begin{subtable}{.4\linewidth}
      \centering
        \caption{Instantiations of SMI functions.}
        \label{tab:smi_inst}
        \begin{tabular}{|c|c|c|}
        \hline
        \textbf{SMI} & \textbf{$I_f(\Acal;\Qcal)$} \\ \hline
        \scriptsize{FLVMI}             & \scriptsize{$\sum\limits_{i \in \Ucal}\min(\max\limits_{j \in \Acal}S_{ij},  \max\limits_{j \in \Qcal}S_{ij})$}                \\
        \scriptsize{FLQMI}             & \scriptsize{$\sum\limits_{i \in \Qcal} \max\limits_{j \in \Acal} S_{ij} + $ $ \sum\limits_{i \in \Acal} \max\limits_{j \in \Qcal} S_{ij}$}                \\
        \scriptsize{GCMI}              & \scriptsize{$2 \sum\limits_{i \in \Acal} \sum\limits_{j \in \Qcal} S_{ij}$}                \\
        \scriptsize{LOGDETMI}          & \scriptsize{$\log\det(S_{\Acal}) -\log\det(S_{\Acal} -$} \\ & \scriptsize{$ S_{\Acal,\Qcal}S_{\Qcal}^{-1}S_{\Acal,\Qcal}^T)$}           \\ \hline
        \end{tabular}
    \end{subtable}%
    \begin{subtable}{.6\linewidth}
      \centering
        \caption{Instantiations of SCG and SCMI functions.}
        \label{tab:scg_scmi_inst}
        \begin{tabular}{|l|l|}
        \hline
        \textbf{SCG} & \textbf{$f(\Acal|\Pcal)$} \\ \hline
        \scriptsize{FLCG}       & \scriptsize{$\sum\limits_{i \in \Ucal} \max(\max\limits_{j \in \Acal} S_{ij}-$ $ \max\limits_{j \in \Pcal} S_{ij}, 0)$}                \\
        \scriptsize{LogDetCG}       & \scriptsize{$\log\det(S_{\Acal} - S_{\Acal,\Pcal}S_{\Pcal}^{-1}S_{\Acal,\Pcal}^T)$} \\
        \scriptsize{GCCG}   & \scriptsize{$f(\Acal) - 2 \sum\limits_{i \in \Acal, j \in \Pcal} S_{ij}$}                \\ \hline
        \end{tabular}
        \vspace{0.5ex}
        
        \begin{tabular}{|l|l|}
        \hline
        \textbf{SCMI} & \textbf{$I_f(\Acal;\Qcal|\Pcal)$} \\ \hline
        \scriptsize{FLCMI}       & \scriptsize{$\sum\limits_{i \in \Ucal} \max(\min(\max\limits_{j \in \Acal} S_{ij},$ $ \max\limits_{j \in \Qcal} S_{ij})$} \scriptsize{$-  \max\limits_{j \in \Pcal} S_{ij}, 0)$}                \\\scriptsize{
        LogDetCMI}   & \scriptsize{$\log \frac{\det(I - S_{\Pcal}^{-1}S_{\Pcal, \Qcal} S_{\Qcal}^{-1}S_{\Pcal, \Qcal}^T)}{\det(I - S_{\Acal \cup \Pcal}^{-1} S_{\Acal \cup \Pcal, Q} S_{\Qcal}^{-1} S_{\Acal \cup \Pcal, Q}^T)}$} \\ \hline               
        \end{tabular}
    \end{subtable} 
\end{table}

\section{\textsc{Similar}: Our Unified Active Learning Framework} \label{sec:unifiedAL}
In this section, we propose a unified active learning framework \textsc{Similar}, which uses SIMs to address the limitations of the current work (see \secref{sec:relwork}). We show that \textsc{Similar} can be effectively applied to a broad range of realistic scenarios and thus acts as \textit{one-stop} solution for AL.

The basic idea behind our framework is to exploit the relationship between the SIMs (\tabref{tab:relSIM}) such that it can be applied to any real-world dataset. Particularly, we use the formulation of SCMI and appropriately choose a query set $\Qcal$ and/or a conditioning set $\Pcal$ depending on the scenario at hand. Towards this end, we use the inspiration from \cite{ash2019deep} where they select data points based on diverse gradients. The SIM functions (see \tabref{tab:SIM_inst}) are instantiated using similarity kernels computed using pairwise similarities $S_{ij}$ between the gradients of the current model. Specifically, we define $S_{ij} = \langle \nabla_{\theta} \Hcal_i(\theta), \nabla_{\theta} \Hcal_j(\theta) \rangle$, where $\Hcal_i(\theta) = \Hcal(x_i, y_i, \theta)$ is the loss on the $i$th data point. Similar to~\cite{wei2015submodularity,ash2019deep}, we use hypothesized labels for computing the gradients, and the corresponding similarity kernels. The hypothesized label for each data point is assigned as the class with the maximum probability. We then optimize a SCMI function: 
\begin{align}\label{eq:SCMI-al}
\max_{\Acal \subseteq \Ucal, |\Acal| \leq B} I_f(\Acal; \Qcal | \Pcal)    
\end{align}
with appropriate choices of query set $\Qcal$ and conditioning set $\Pcal$. In the context of batch active learning, $\Acal$ is the batch and $B$ is the budget (batch size in AL). We present our unified AL framework in \AlgRef{algo:unifiedAL} and illustrate the choices of query and conditioning set for realistic scenarios in \figref{fig:choice}.

\begin{algorithm}
\begin{algorithmic}[1]
\REQUIRE Initial Labeled set of data points: $\Lcal$, large unlabeled dataset: $\Ucal$, Loss function $\Hcal$ for learning model $\Mcal$, batch size: $B$, number of selection rounds: $N$ \\
\FOR{selection round $i = 1:N$}
\STATE Train model $\Mcal$ with loss $\Hcal$ on the current labeled set $\Lcal$ and obtain parameters $\theta$
\STATE Using model parameters $\theta_i$, compute gradients using hypothesized labels $\{\nabla_{\theta} \mathcal H(x_j, \hat{y_j}, \theta), \forall j \in \Ucal\}$ and obtain a similarity matrix $X$.
\STATE Instantiate a submodular function $f$ based on $X$.
\STATE $\Acal_i \leftarrow \mbox{argmax}_{\Acal \subseteq \Ucal, |\Acal | \leq B}  I_f(\Acal; \Qcal | \Pcal)$  (Optimize SCMI with an appropriate choice of $\Qcal$ and $\Pcal$, see \tabref{tab:relSIM})
\STATE Get labels $L(\Acal_i)$ for batch $\Acal_i$ and $\Lcal \leftarrow \Lcal \cup L(\Acal_i)$, $\Ucal \leftarrow \Ucal - \Acal_i$
\ENDFOR
\STATE Return trained model $\Mcal$ and parameters $\theta$.
\end{algorithmic}
\caption{\textsc{Similar}: Unified AL Framework}
\label{algo:unifiedAL}
\end{algorithm}

In the scenarios below, we will discuss how this paradigm can provide a unified view of active learning, handle aspects like standard active learning (\secref{sec:standardAL}), rare classes and imbalance (\secref{sec:rareClsAL}), redundancy (\secref{sec:redundancyAL}) and, OOD/outliers in the unlabeled data (\secref{sec:OODAL}).  

\subsection{Standard Active Learning} \label{sec:standardAL}
We refer to standard active learning for ideal scenarios when there is no imbalance, redundancy or OOD data in the unlabeled dataset. In such cases, there is no requirement for having a query set and conditioning set. Hence, given a SCMI function $I_f(\Acal; \Qcal | \Pcal)$, we get $I_f(\Acal; \Qcal | \Pcal) = f(\Acal)$ by setting $\Qcal \leftarrow \Ucal$ (the unlabeled dataset) and $\Pcal \leftarrow \emptyset$. In a nutshell, the standard diversified active learning setting can be seen as a special case of our proposed unified AL framework (\equref{eq:SCMI-al}) by choosing $\Qcal, \Pcal$ as above. Note that this approach is very similar and closely related to  \textsc{Badge}~\cite{ash2019deep}, where the authors also choose points based on diverse gradients. Furthermore, the authors discuss the use of Determinantal Point Processes (DPP) \cite{kulesza2012determinantal} for sampling, and this is very similar to maximizing log-determinants. In the supplementary paper, we compare the choice of different submodular functions for AL.\looseness-1

\subsection{Rare Classes} \label{sec:rareClsAL}
\begin{wrapfigure}{R}{0.50\textwidth}
\vspace{-5ex}
\includegraphics[width=0.50\textwidth]{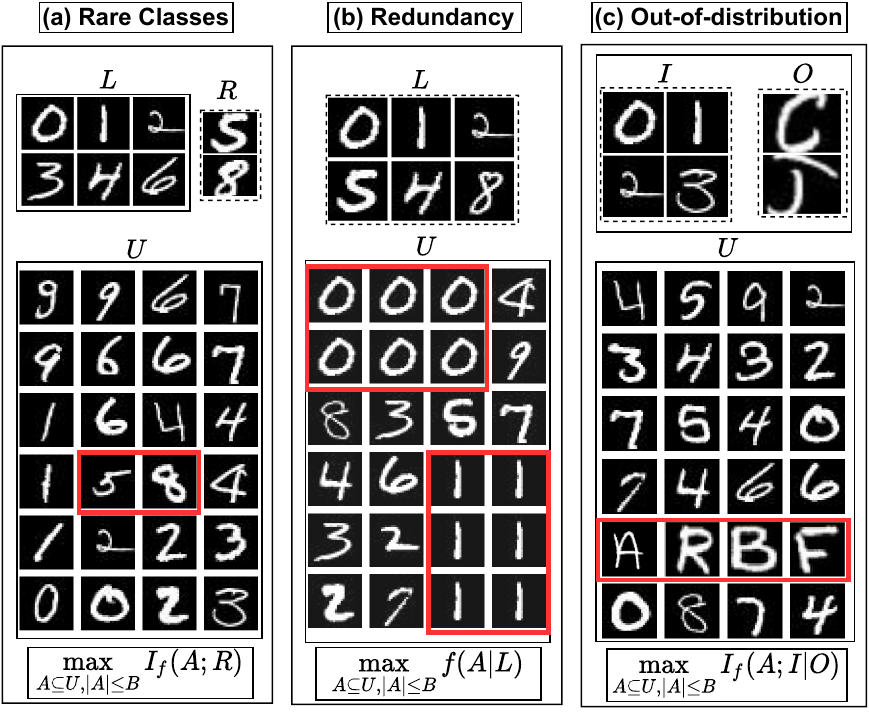}
\caption{An illustration of realistic scenarios where \textsc{Similar} is applied with appropriate choices of query and conditioning sets: a) \textsc{Similar} finds rare digits $5,8 \in \Ucal$, by optimizing the SMI function $I_f(\Acal; \Rcal)$ with $\Rcal$ containing $5,8$ as \textit{queries}, b) select samples from $\Ucal$ which are diverse among themselves and also diverse w.r.t those in $\Lcal$ by optimizing $f(\Acal | \Lcal)$ (here, we want to \textit{avoid} digits $0,1 \in \Ucal$ altogether because they are present in $\Lcal$), c) select digits (in-distribution) and avoid alphabets (out-of-distribution) in $\Ucal$ by optimizing $I_f(\Acal; \Ical | \Ocal)$, where $\Ical$ are ID labeled points and $\Ocal$ are OOD points selected so far.
 \vspace{-5ex}
}
\label{fig:choice}
\end{wrapfigure}

A very common and naturally occurring scenario is that of imbalanced data. This imbalance is because some classes or attributes are naturally more frequently occurring than others in the real-world. For example, in a self-driving car application, there may be very few images of pedestrians at night on highways, or cyclists at night. Another example is medical imaging, where there are many rare yet important diseases (e.g., various forms of cancers), and it is often the case that non-cancerous images are much more than compared to the cancerous ones. While such classes are rare, it is also critical to be able to perform well in these classes. The problem with running standard active learning algorithms in such a case is that they may not sample too many data points from these rare classes, and as a result, the model continues to perform poorly on these classes. 
In such cases, we can create a (small) held-out set $\Rcal$ which contains data points from these rare classes, and try to encourage the AL by sampling more of these rare classes by maximizing the SMI function $I_f(\Acal; \Rcal)$:
\begin{align}\label{eq:smi-max-rare}
        \max_{\Acal \subseteq \Ucal, |\Acal| \leq B} I_f(\Acal ; \Rcal)
\end{align}
This setting is shown in \figref{fig:choice}(a). $\Rcal$ contains a small number of held-out examples of classes $5, 8$ which are rare, and the AL acquisition function is \equref{eq:smi-max-rare}. Note that this is exactly equivalent to maximizing the SCMI function with $\Qcal \leftarrow \Rcal$ and $\Pcal \leftarrow \emptyset$ (i.e. \equref{eq:SCMI-al} in Line 5 of \AlgRef{algo:unifiedAL}). Furthermore,  since the SMI functions naturally model query relevance and diversity, they will also try to pick a diverse set of data points which are relevant to $\Rcal$. Finally, we also point out that this setting was considered in~\cite{gudovskiy2020deep} where they use a \textsc{Fisher} kernel based approach to sample data points. Note that for this setting to be realistic, it is critical that the size of this validation set is very small -- \cite{gudovskiy2020deep} uses a much larger validation set which is not very realistic (e.g., $200\times$ our set, see Appendix~\ref{app:experimental} for more details).\looseness-1 
\subsection{Redundancy in Unlabeled Data} \label{sec:redundancyAL}
Another commonplace scenario is where we are dealing with a lot of redundancy -- \textit{e.g.,} frames sampled from a video, where subsequent frames are visually similar. In such cases, existing AL algorithms tend to pick data points that are semantically similar to the ones selected in some earlier batch. This is true even for the state-of-the-art AL algorithm \textsc{Badge}~\cite{ash2019deep} that attempts to enforce diversity, but only in the current batch of data points and not the already selected labeled set. To illustrate this, consider the scenario in \figref{fig:choice}(b). The digits $0, 1$ are redundant in the unlabeled set, and they are already present in the labeled set $\Lcal$. Algorithms which just focus on diversity in the current batch could fail at ensuring diversity across batches. To mitigate inter-batch redundancy, we use SCG acquisition function and condition upon the already labeled set $\Lcal$:
\begin{align} \label{eq:scg-max-redun}
    \max_{\Acal \subseteq \Ucal, |\Acal| \leq B} f(\Acal | \Lcal)
\end{align}
Notice that this is a special case of our proposed unified AL framework (\equref{eq:SCMI-al}) since the SCG function $f(\Acal | \Lcal)$ is basically a SCMI function with $\Qcal \leftarrow \emptyset$ and $\Pcal \leftarrow \Lcal$. 

\subsection{Out of Distribution Data} \label{sec:OODAL}
In real world scenarios, we often have out-of-distribution (OOD) data or irrelevant classes in the unlabeled set. Such OOD data is not useful for the given classification task at hand. Using an acquisition function that selects a lot of OOD data points will lead to a waste of labeling effort and time. This is because annotators have to spend time in filtering out OOD data points and discard them from the training dataset. To account for OOD data, we add an additional class called "OOD" in our model. Since the goal is to improve on in-distribution classes , we ignore the prediction for the OOD class at test time. For our AL acquisition function, we use the currently labeled OOD points $\Ocal$ as the conditioning set $\Pcal$, and the currently labeled in-distribution (ID) points $\Ical$ as the query set $\Qcal$. In other words, our acquisition function is to optimize:\looseness-1
\begin{align}\label{eq:SCMI-ood}
    \max_{\Acal \subseteq \Ucal, |\Acal| \leq B} I_f(\Acal; \Ical | \Ocal)
\end{align}
This is illustrated in \figref{fig:choice}(c), where the labeled set consists of six examples, four of them being ID data points (set $\Ical$) and two being OOD data points (set $\Ocal$). In \figref{fig:choice}(c), the ID data are digits (digit classification) and the OOD examples are alphabets. This SCMI based approach will naturally pick points "close" to the ID data while avoiding the OOD points. 

Another approach for designing the acquisition function is to not explicitly condition on the OOD data points. In other words, we can just optimize the SMI function:
\begin{align}\label{eq:smi-ood}
    \max_{\Acal \subseteq \Ucal, |\Acal| \leq B} I_f(\Acal; \Ical)
\end{align}
We contrast the choices of SCMI (\equref{eq:SCMI-ood}) and SMI (\equref{eq:smi-ood}) functions in our experiments. 

\subsection{Multiple Co-occurring Realistic Scenarios} \label{sec:multipleScenarios}
We can also apply \textsc{Similar} to datasets where more than one realistic scenarios are co-occurring. As illustrated in \tabref{tab:multipleSIM}, we can use the formulation of SCMI and make appropriate choices of $\Qcal$ and $\Pcal$ to tackle multiple realistic scenarios. 

\begin{table}[!ht]
\centering
\small{
\begin{tabular}{|l|l|l|}
\hline
\textbf{Function} & \textbf{Setting} & \textbf{Realistic Scenario} \\ \hline
$I_f(\Acal;\Rcal|\Ocal)$               & $\Qcal \leftarrow \Rcal, \Pcal \leftarrow \Ocal$                & Rare classes + OOD              \\
$ I_f(\Acal; \Rcal | \Lcal-\tilde{\Rcal})$               & $\Qcal \leftarrow \Rcal, \Pcal \leftarrow \Lcal-\tilde{\Rcal}$                & Rare classes + Redundancy              \\
$I_f(\Acal; \Ical|\Ocal \cup \Ical^{'})$               & $\Qcal \leftarrow \Ical, \Pcal \leftarrow \Ocal \cup \Ical^{'}$                & Redundancy + OOD                                  \\ \hline  
\end{tabular}
\vspace{1ex}
\caption{Choices for $\Qcal$ and $\Pcal$ for multiple co-occuring realistic scenarios}
\label{tab:multipleSIM}}
\end{table}

 \textbf{Rare classes and OOD:} We set $\Qcal \leftarrow \Rcal$ and $\Pcal \leftarrow \Ocal$ and maximize $I_f(\Acal; \Rcal|\Ocal)$. Intuitively, this function would pick points close to $\Rcal$ while avoiding the OOD points. In this scenario, we can also optimize an SMI function $I_f(A;R)$ if the data points belonging to the rare classes are not similar to the OOD data points, meaning that only searching for rare classes may suffice. Regardless, the SCMI approach above will further reinforce the avoidance of the OOD points.

\textbf{Rare classes and Redundancy: }We set $\Qcal \leftarrow \Rcal$ and $\Pcal \leftarrow \Lcal-\tilde{\Rcal}$. Here, $\tilde{\Rcal}$ is the subset of data points from the labeled set $\Lcal$ that belong to the rare classes. Intuitively, this function would pick points close to $\Rcal$ while avoiding points already in $\Lcal-\tilde{\Rcal}$, thereby avoiding redundant data. Just focusing on $\Rcal$ by optimizing $I_f(\Acal;\Rcal)$ is also a feasible option because rare classes are generally not redundant. As before, the SCMI approach will only reinforce the avoidance of redundant samples in any non-rare class instances selected.

\textbf{Redundancy and OOD: } This is a more challenging scenario than the ones above. We start with using the SCMI formulation for the OOD scenario, i.e., $I_f(\Acal;\Ical|\Ocal)$, where $\Ical$ is the set of ID samples and $\Ocal$ is the set of OOD samples. Optimizing this function will pick diverse in-distribution samples within a batch. For selecting diverse samples across different batches, we can tackle this by using an appropriate kernel for the conditioning set. For instance, consider the \textsc{Flcmi} function in \tabref{tab:SIM_inst}(b). On setting $\Pcal \leftarrow \Ocal \cup \Ical$, we can rewrite the \textsc{Flcmi} function by splitting the penalty term as follows: $\sum\limits_{i \in \Ucal} \max(\min(\max\limits_{j \in \Acal} S_{ij}, \max\limits_{j \in \Ical} S_{ij}) - \max(\max\limits_{j \in \Ocal} S_{ij}, \max\limits_{j \in \Ical} S'_{ij}), 0)$. 
While $S$ is computed using cosine similarity, we can compute $S'$ using an exponential kernel to magnify the value of $S'_{ij}$ using the exponent when $i$ and $j$ are very similar. This exponent is a hyperparameter which can be tuned to penalize selecting redundant samples from $I$ (denoted as $I^{'}$) in \tabref{tab:multipleSIM}.

\subsection{Realizing Realistic Scenarios in Applications}
In this section, we discuss a few insights on how these realistic scenarios can be realized. To begin with, the initial labeled set used in AL usually follows the distribution of the unlabeled set. The statistics of this set can be used to identify rare classes. If the initial seed set is small, the rare classes/OOD data points can be realized after a few rounds of standard AL. Until such scenarios are discovered, standard AL can be done using a diversity-based acquisition function like the log determinant (\textsc{Logdet}). For production-level models, they go through a test deployment phase. During this phase, systematically recurring errors are often found. An example is of undetected bicycles at night in an object detector (false negatives). Such recurring failure cases can be due to rare classes in the labeled set. Moreover, we as users often know whether there are rare classes or if there is redundancy from domain knowledge.  For instance, in the biomedical domain, images of cancer cells are typically rarer than ones of non-cancer cells because cancer inherently is a rare disease. 

\subsection{Scalability and Computational Aspects of \textsc{Similar}} \label{sec:scalability}
\textbf{Computational Complexity: } The computational complexity of the different SMI functions are determined by (1) the kernel computation time, and (2) the time complexity of the greedy algorithm. All functions considered here are graph based functions and require computing a kernel matrix. The \textsc{Logdet} functions (\textsc{Logdet}, \textsc{Logdetmi}, \textsc{Logdetcg}, \textsc{Logdetcmi}), some \textsc{Fl} functions (\textsc{Fl}, \textsc{Flvmi}, \textsc{Flcmi}), and GC, \textsc{Gcmi} all require the $n \times n$ similarity matrix ($n = |\Ucal|$ is the number of unlabeled points) which entails a complexity of $O(n^2)$ to construct the similarity kernel. Once constructed, the complexity of the greedy algorithm for \textsc{LogDet} class of functions is roughly $O(B^3n)$~\cite{chen2018fast}, while the complexity of the greedy algorithm with \textsc{Fl}, \textsc{Flvmi}, and \textsc{Flcmi} is $O(Bn^2)$~\cite{iyer2019memoization,iyer2015submodular}($B$ is the batch size). Different from others, \textsc{Flqmi} does not require computing a $n \times n$ kernel, but only a $n \times q$ kernel (where $q=|\Qcal|$ is the number of query points). Correspondingly, the complexity of the greedy algorithm with \textsc{Flqmi} is $O(nqB)$, and is linear in $n$. In \Appref{app:computational_aspect}, we provide a detailed summary of the complexity of different SF, SMI, SCG, and SCMI functions. 

\textbf{Partition Trick: } The deal with the high $O(n^2)$ of the \textsc{Logdet}, \textsc{Gc}, and some of the \textsc{Fl} variants (except \textsc{Flqmi}), we also propose the following partitioning algorithm: We randomly split the unlabeled set $\Ucal$ into $p$ partitions $\Ucal_1, \cdots, \Ucal_p$, and we then define the corresponding function (SF, SMI, SCMI, SCG) on each of the partitions and independently optimize them. In each partition, we select $B/p$ points. The complexity of this reduces from $O(n^2)$ to $O(n^2/p)$ and with an appropriate choice of $p$, we can significantly reduce the computational complexity. We use this in our ImageNet experiments (see \secref{sec:expRareCls}), and observe that our approaches continue performing well while being more scalable. We provide more details on partitioning in \Appref{app:computational_aspect}.\looseness-1

\textbf{Last Layer Gradients: } Deep models have numerous parameters leading to very high dimensional gradients. Since our kernel matrix is computed using the cosine similarity of gradients, this becomes intractable for most models. To solve this problem, we use last-layer gradient approximation by representing data points using last layer gradients. \textsc{Badge} \cite{ash2019deep}, \textsc{Coreset} \cite{sener2017active} and \textsc{Glister} \cite{killamsetty2020glister} are other baselines that also use this approximation.  Using this representation, we compute a pairwise cosine similarity matrix to instantiate acquisition functions in \textsc{Similar} (see lines 3,4 in \AlgRef{algo:unifiedAL}).\looseness-1

\begin{figure*}
\centering
\includegraphics[width = 14cm, height=1cm]{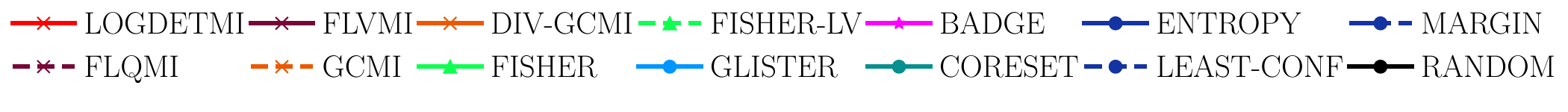}
\centering
\hspace*{-0.6cm}
\begin{subfigure}[t]{0.33\textwidth}
\includegraphics[width = \textwidth]{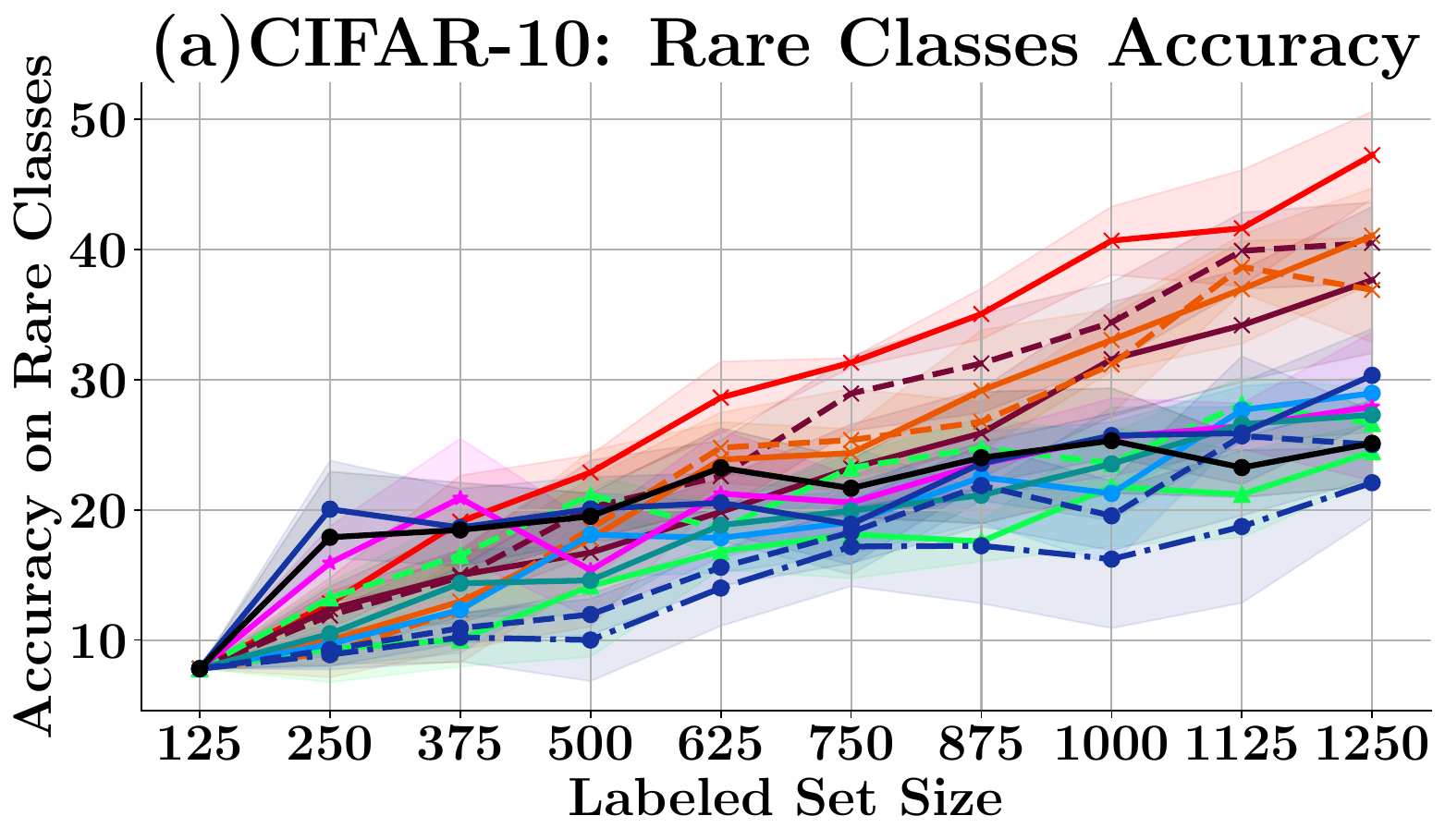}
\end{subfigure} 
\begin{subfigure}[t]{0.33\textwidth}
\includegraphics[width = \textwidth]{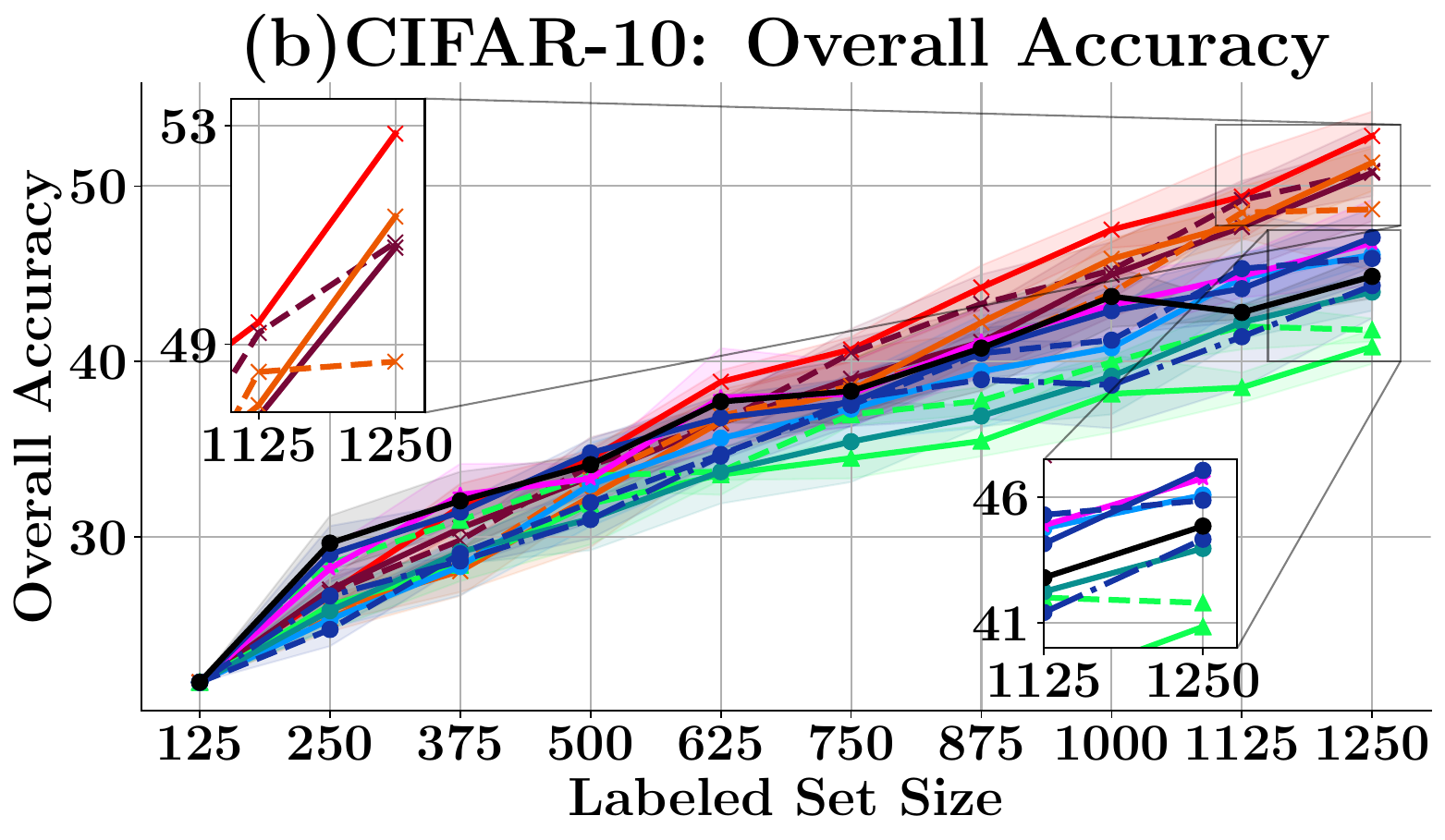}
\end{subfigure}
\begin{subfigure}[t]{0.33\textwidth}
\includegraphics[width = \textwidth]{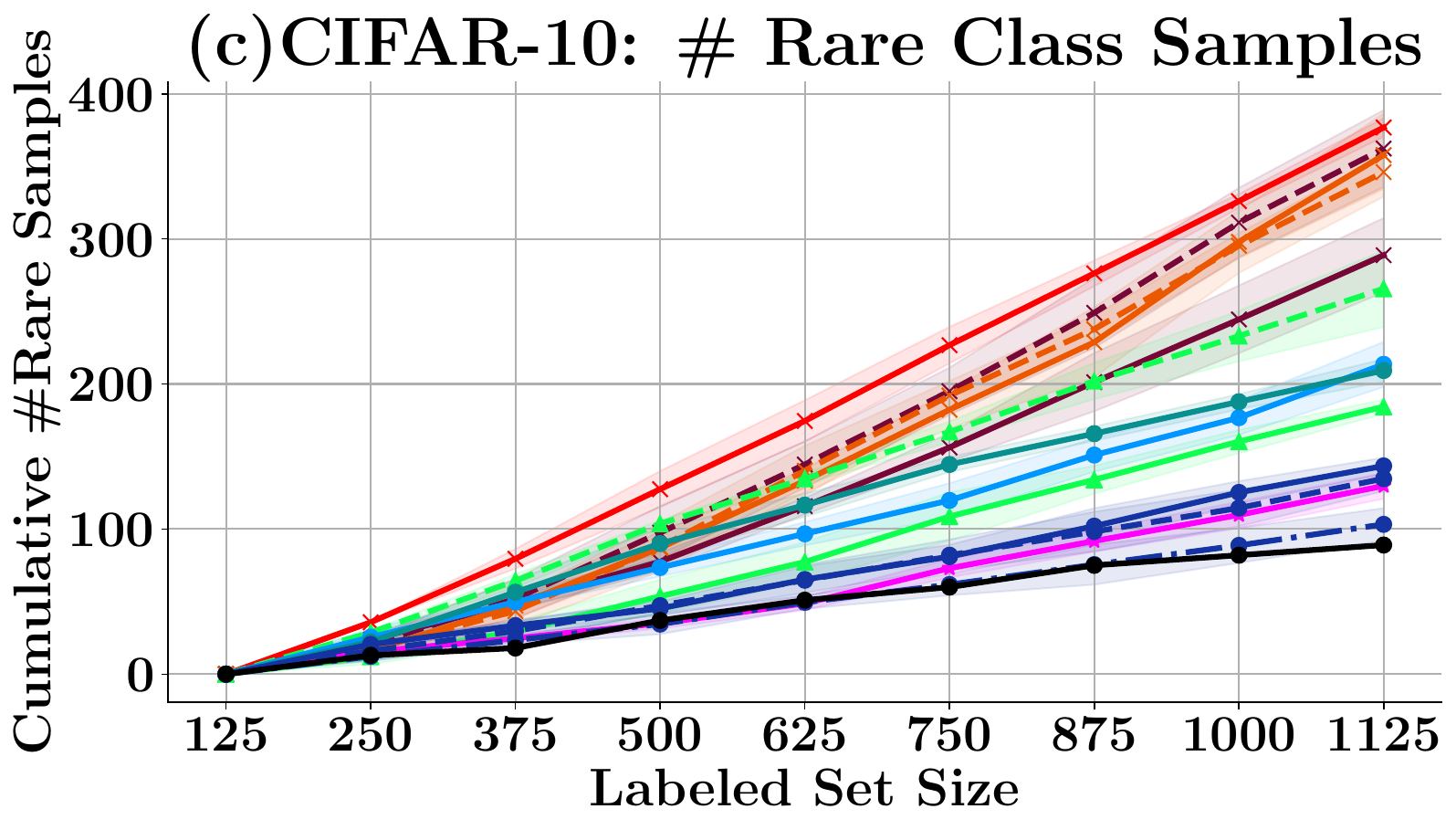}
\end{subfigure}
\begin{subfigure}[b]{0.33\textwidth}
\includegraphics[width = \textwidth]{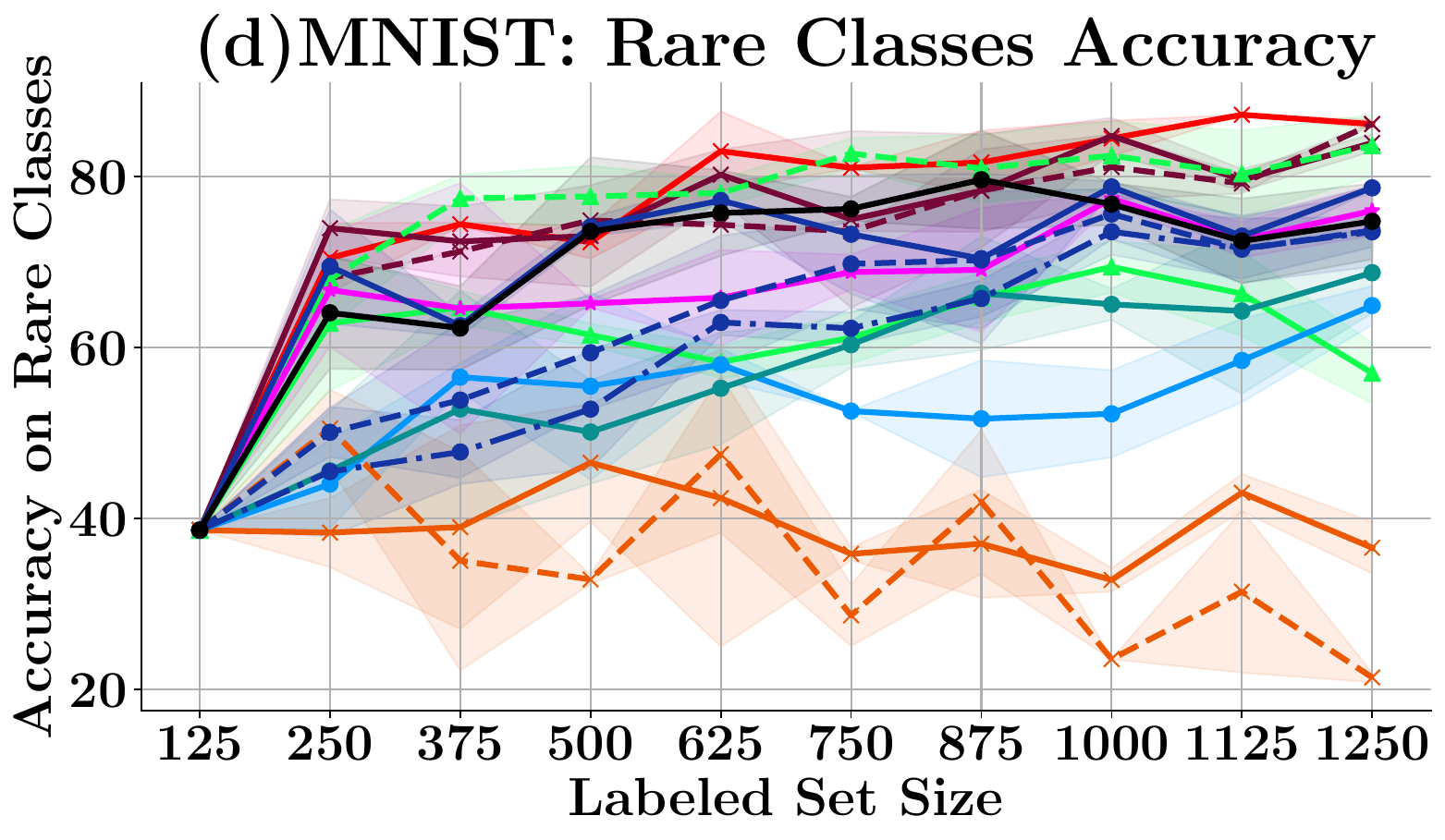}
\end{subfigure}
\begin{subfigure}[t]{0.33\textwidth}
\includegraphics[width = \textwidth]{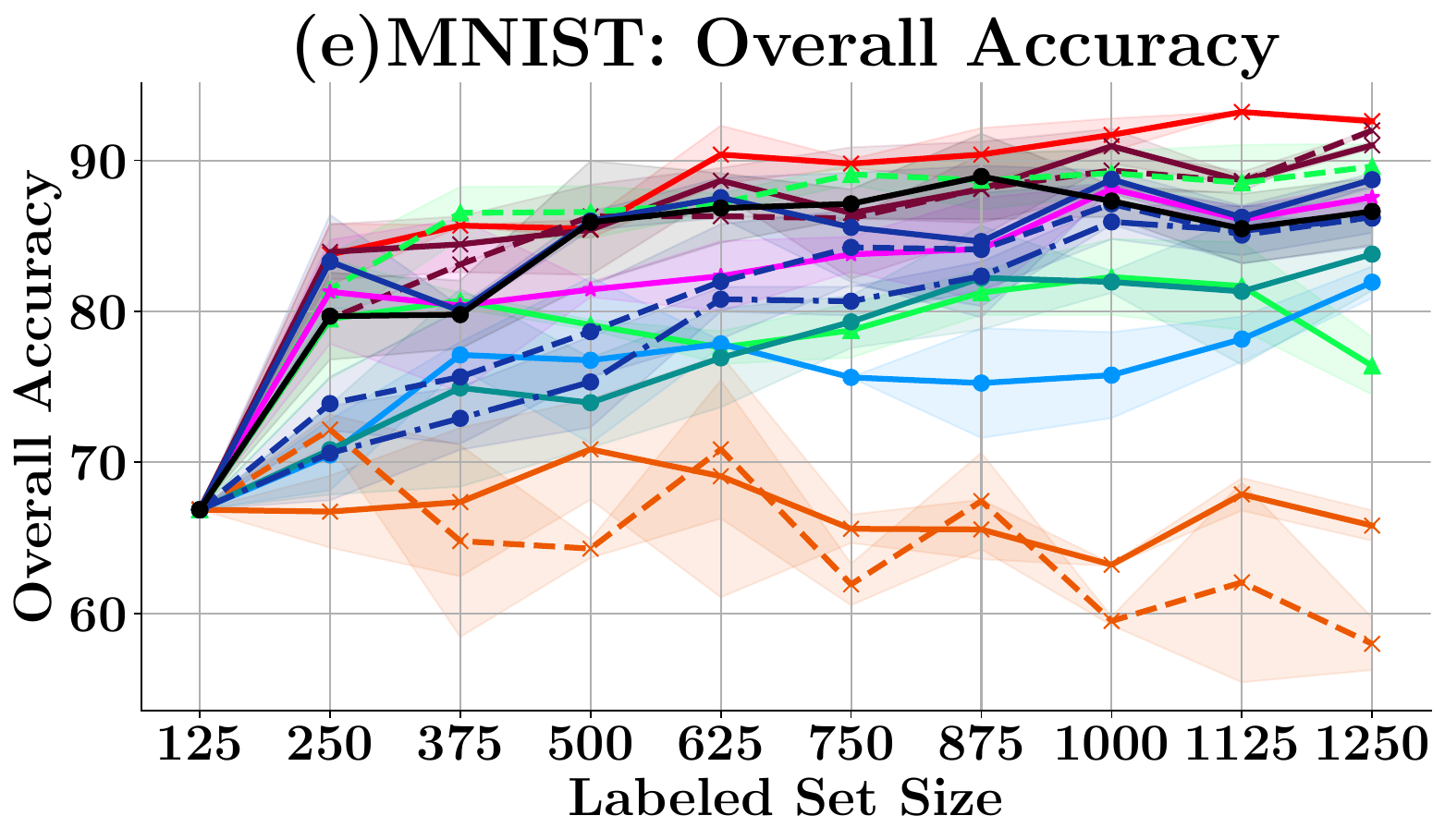}
\end{subfigure} 
\begin{subfigure}[t]{0.325\textwidth}
\includegraphics[width = \textwidth]{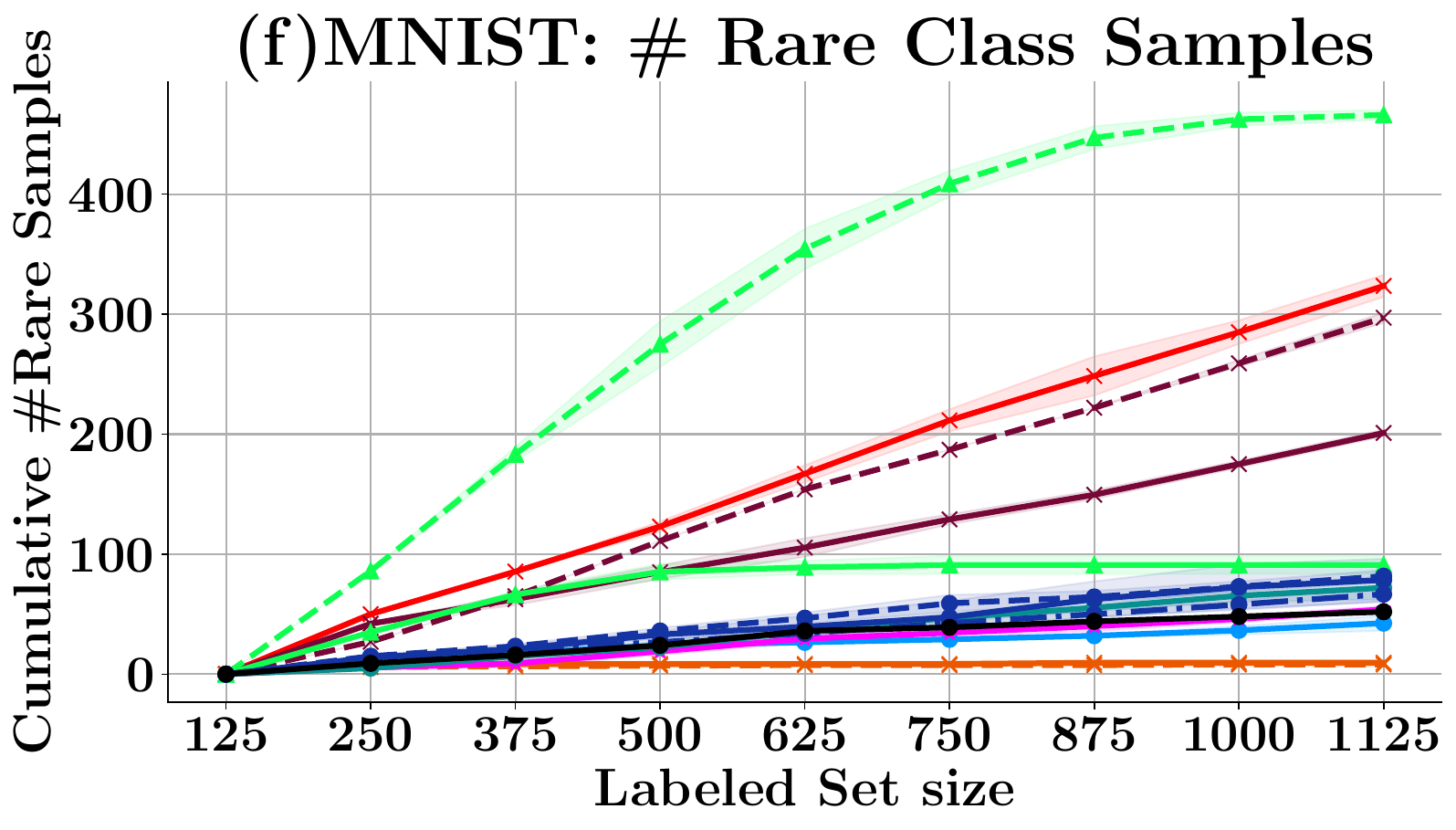}
\end{subfigure}
\begin{subfigure}[b]{0.33\textwidth}
\includegraphics[width = \textwidth]{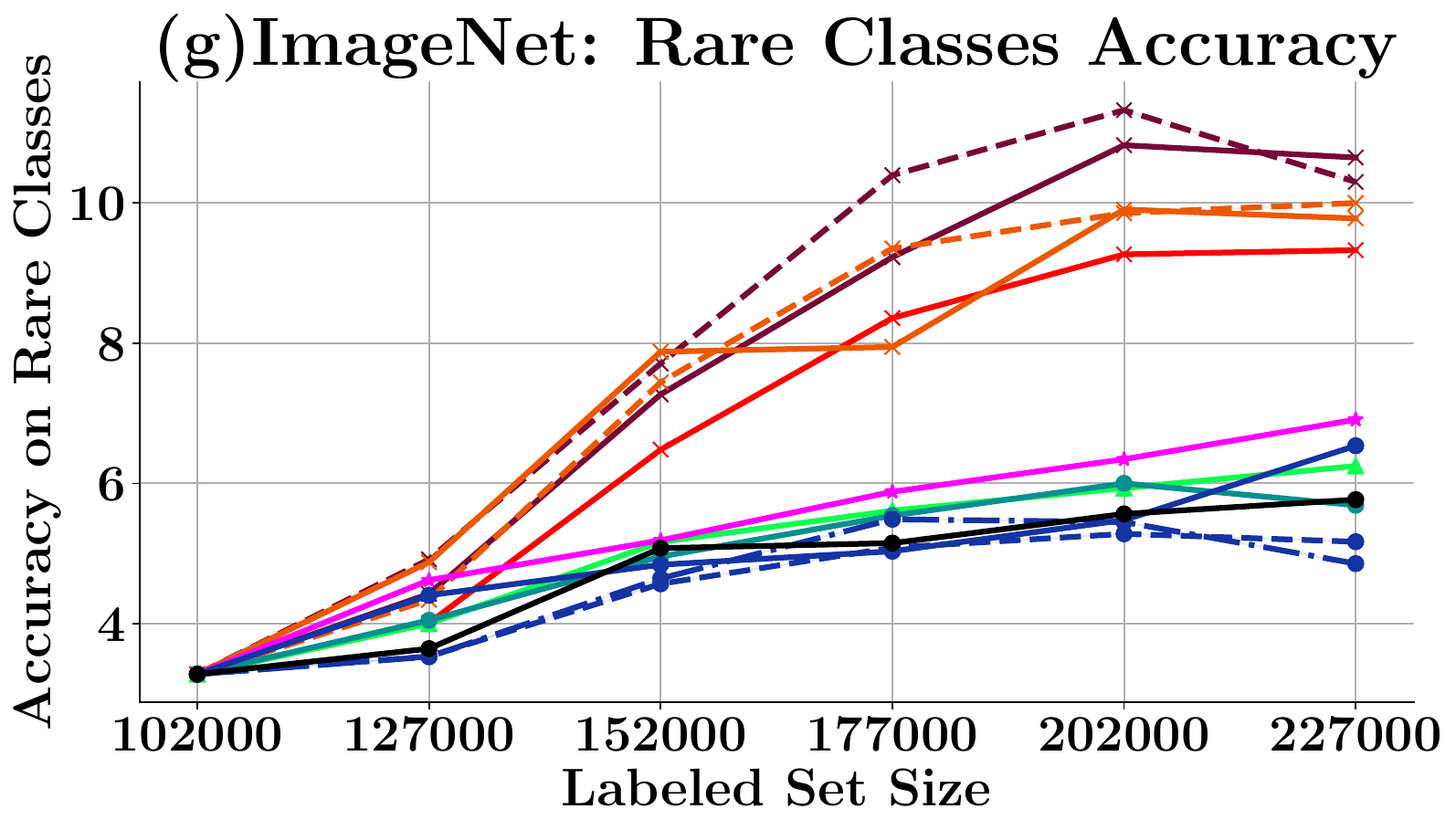}
\end{subfigure}
\begin{subfigure}[t]{0.33\textwidth}
\includegraphics[width = \textwidth]{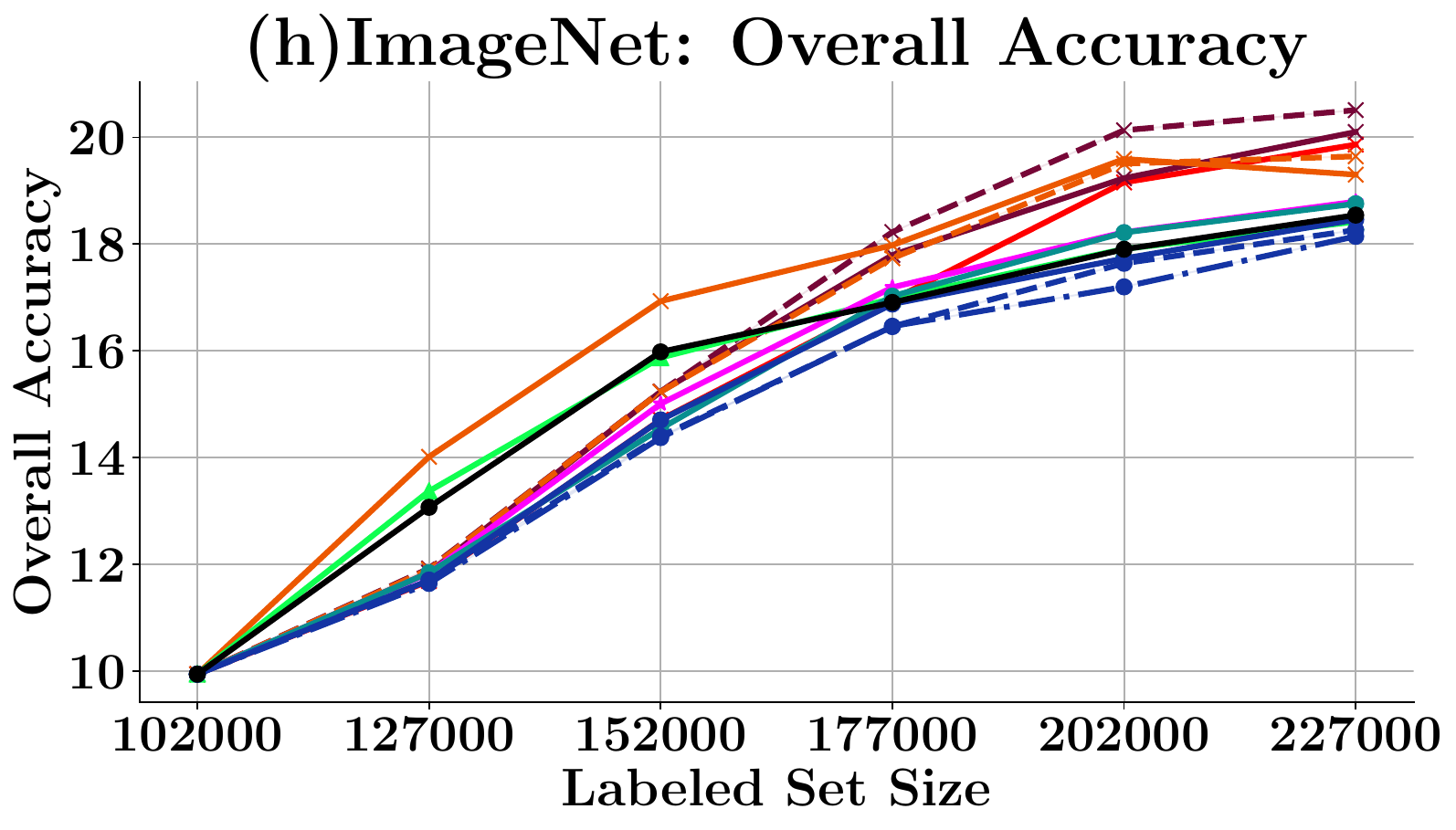}
\end{subfigure} 
\begin{subfigure}[t]{0.325\textwidth}
\includegraphics[width = \textwidth]{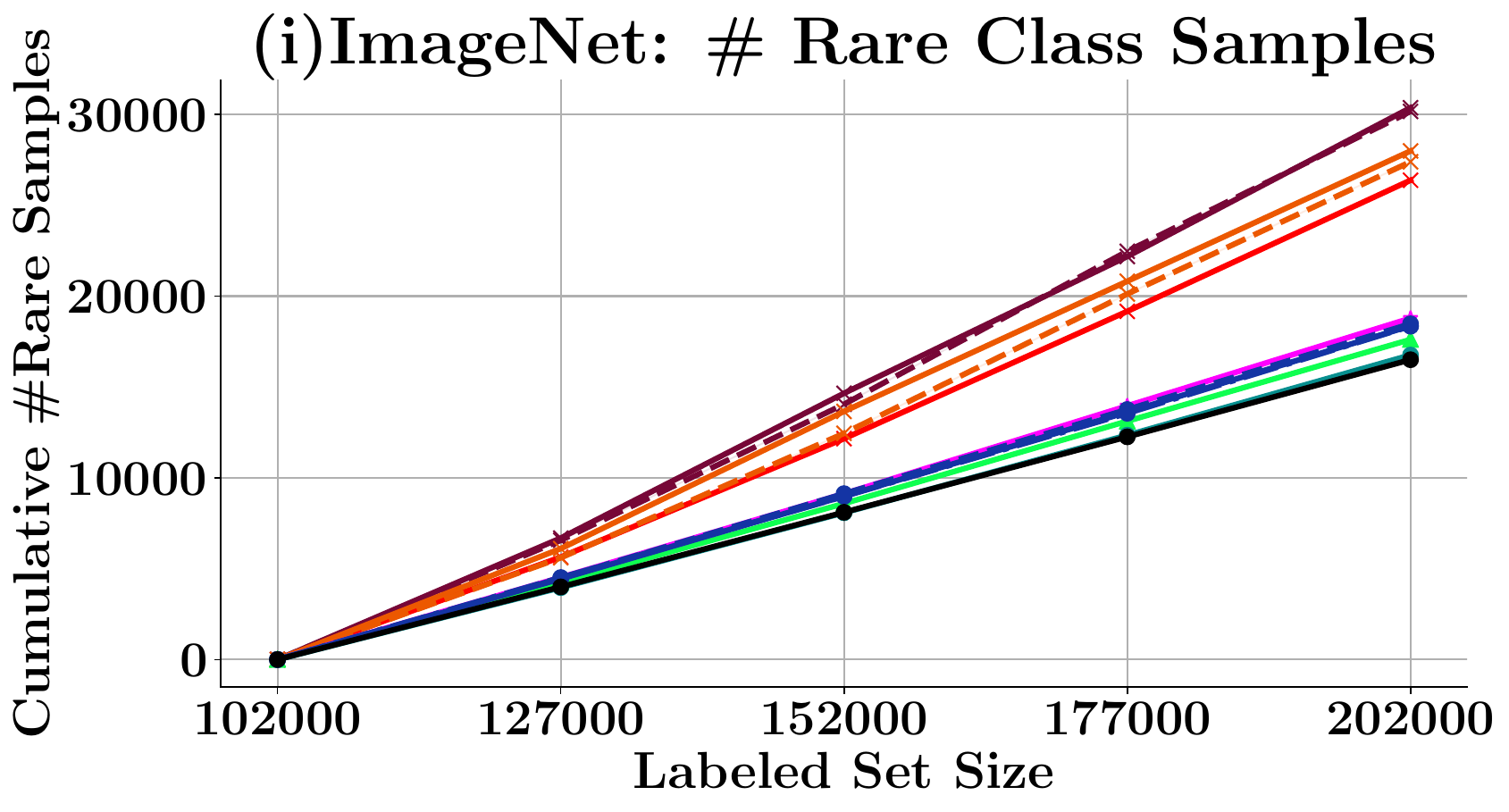}
\end{subfigure}
\caption{Active Learning with rare classes on CIFAR-10 (top row), MNIST (middle row), and ImageNet (bottom row). Left side plots (a,d,g) are rare class accuracies, center plots (b,e,h) are overall test accuracies, right plots (c,f,i) are a number of rare class samples selected. The SMI functions (specifically \textsc{Logdetmi}, \textsc{Flqmi}) outperform other baselines by more than 10\% on the rare classes.
\vspace{-1ex}
}
\label{fig:res_rarecls}
\end{figure*}

\section{Experimental Results}
\label{sec:exp}
In this section, we empirically evaluate the effectiveness of \textsc{Similar} on a wide range of scenarios like rare classes (\secref{sec:expRareCls}), redundancy (\secref{sec:expRedundancy}) and out-of-distribution (\secref{sec:expOOD}). We do so by comparing the accuracy and selections of various SCMI based acquisition functions with existing AL approaches. Using these experiments, we cover the issues with the current AL methods and show that these issues can be mitigated by using a unified implementation using SCMI with appropriate choices of query and/or conditioning sets. Although this section focuses on realistic scenarios, we also study \textsc{Similar} in a standard active learning setting and show that it performs at par with current AL methods (see \Appref{app:standard_al}). Furthermore, we present some experiments on a real-world medical dataset in \Appref{app:MedicalDatasetRes} and some experiments on multiple co-occurring realistic scenarios (\secref{sec:multipleScenarios}) in \Appref{app:resMultipleScenarios}. \looseness-1

\textbf{Baselines in all scenarios:}
We compare SCMI based functions against several methods. Particularly, we compare against: (1) three uncertainty based AL algorithms: i)\textsc{Entropy:} Selects the top $B$ data points with the highest \textit{entropy}~\cite{settles2009active}, ii) \textsc{Margin:} Select the bottom $B$ data points that have the least difference in the confidence of first and the second most probable labels~\cite{roth2006margin}, iii)\textsc{Least-Conf:}  Select $B$ samples with the smallest predicted class probability~\cite{wang2014new}, (2) state-of-the-art diversity based algorithms: iv) \textsc{Badge} \cite{ash2019deep} v) \textsc{Glister} \cite{killamsetty2020glister} vi) \textsc{Coreset} \cite{sener2017active} which are all discussed in section \secref{sec:relwork}, and, 3) \textsc{Random:} Select $B$ samples randomly. Additionally, in the rare classes scenario, we compare against \textsc{Fisher} \cite{gudovskiy2020deep} which is also discussed in \secref{sec:relwork}.

\textbf{Datasets, model architecture and experimental setup:}
 We apply our framework to CIFAR-10 \cite{krizhevsky2009learning} and MNIST \cite{lecun2010mnist} classification tasks. Additionally, we also evaluate our method on down sampled $32 \times 32$ ImageNet-2012 \cite{russakovsky2015imagenet} for the rare classes setting (\secref{sec:expRareCls}). Due to the lack of test split on ImageNet, we used the validation split for evaluation. In the sections below, we discuss the individual splits for $\Lcal$, $\Ucal$, $\Rcal$, $\Ical$, and $\Ocal$ in each realistic scenario. To ensure that all the selection algorithms that we are studying are given fair and equal treatment across all realistic scenarios, we use a common training procedure and hyperparameters. We use standard augmentation techniques like random crop, horizontal flip followed by data normalization except for MNIST which does not use horizontal flip to preserve labels. For training, we use an SGD optimizer with an initial learning rate of 0.01, the momentum of 0.9, and a weight decay of 5e-4. We decay the learning rate using cosine annealing \cite{loshchilov2016sgdr} for each epoch. On all datasets except MNIST, we train a ResNet18 \cite{he2016deep} model, while on MNIST we train a LeNet \cite{lecun1989backpropagation} model. For all the experiments in a particular scenario (rare classes, redundancy and OOD), we start with an identical initial model $\Mcal$ and initial labeled set $\Dcal$. 
 We reinitialize the model parameters at the beginning of every selection round using Xavier initialization and train the model until either the training accuracy reaches 99\% or the epoch count reaches 150. We run each experiment $3 \times$ on CIFAR-10 and MNIST and $1 \times$ on ImageNet and provide error bars (std deviation). All experiments were run on a V100 GPU. 
 For more details on the experimental setup, baselines, and datasets see \Appref{app:experimental}.\looseness-1
\vspace{-1ex}

\subsection{Rare Classes} \label{sec:expRareCls}
\textbf{Custom dataset:} Following \cite{gudovskiy2020deep, killamsetty2020glister}, we simulate these rare classes by creating a class imbalance. We initialize the batch active learning experiments by creating a custom dataset which is a subset of the full dataset with the same marginal distribution. Given that $\Ccal$ consists of data points from the imbalanced classes and $\Dcal$ consists of data points from the balanced classes, we create an initial labeled set $\Lcal$ such that $|\Dcal_\Lcal| = \rho |\Ccal_\Lcal|$ and an unlabeled set $|\Dcal_\Ucal| = \rho |\Ccal_\Ucal|$, where $\rho$ is the imbalance factor. We use a small and clean validation/query set $\Rcal$ containing data points from the imbalanced classes ($\approx 3$ data points per imbalanced class). We create an imbalance in CIFAR-10 using $5$ random classes, $\rho=10$ and for MNIST we create an imbalance using the same classes as in \cite{gudovskiy2020deep} $(5 \cdots 9)$ and use $\rho=20$. For both datasets: $|\Ccal_\Lcal| + |\Dcal_\Lcal| = 125$, $|\Ccal_\Ucal| + |\Dcal_\Ucal| = 16.5K$, $B=125$ (AL batch size) and, $|\Rcal| = 25$ (size of the held out rare instances). For MNIST, we also present the results for $B=25$ and $\rho=100$ in the supplementary. On ImageNet, we randomly select $500$ classes out of 1000 classes for imbalance and $\rho=5$ such that $|\Ccal_\Lcal| + |\Dcal_\Lcal| = 102K$, $|\Ccal_\Ucal| + |\Dcal_\Ucal| = 664K$, $B=25K$ and, $|\Rcal| = 2.5K$. These data splits are chosen to simulate a low initial accuracy on the rare classes and at the same time maintain the imbalance factor in the labeled and unlabeled datasets. 

\textbf{Results:} The results are shown in \figref{fig:res_rarecls}. We observe that SMI based functions not only consistently outperform uncertainty based methods (\textsc{Entropy}, \textsc{Least-Conf} and \textsc{Margin}) but also all the state-of-the-art diversity based methods (\textsc{Badge}, \textsc{Glister}, \textsc{Coreset}) by \textbf{$\approx 5-10\%$} in terms of overall accuracy and \textbf{$\approx 10-18\%$} in terms of average accuracy on rare classes (see \figref{fig:res_rarecls}a, 3d, 3g). The reason for the same can be seen in \figref{fig:res_rarecls}c, 3f, 3i which illustrates that they fail to pick an adequate number of examples from the rare classes. Evidently, \textsc{Flqmi} and \textsc{Logdetmi} which balance between diversity and relevance perform better than \textsc{Gcmi} which only models relevance. Furthermore, \textsc{Div-Gcmi} which is a linear combination of \textsc{Gcmi} and a diversity term performs consistently worse, which suggest that a naive combination of the two may not be as effective. This suggests the need of SMI based acquisitions functions (\equref{eq:smi-max-rare}) with richer modeling capabilities like \textsc{Flqmi} and \textsc{Logdetmi} within \textsc{Similar}.  
Furthermore, all SMI based functions also outperform the \textsc{Fisher} kernel based method when the validation set is small and realistic, \emph{i.e.,} $|\Rcal|=25$. Since, \cite{gudovskiy2020deep} use a very large validation set in their experiments, we try their method \textsc{Fisher-Lv} with a \textbf{$40 \times$} larger validation set of size 1000 (which is \textit{not practical}) and observe a comparable performance with the SMI functions which use a small validation set. Furthermore, we see that \textsc{Fisher-Lv} actually picks significantly larger number of rare class instances in MNIST, but yet is comparable in performance of \textsc{Flqmi} and \textsc{Logdetmi}. This suggests that both these methods select higher quality and diverse rare class instances. We observe that the \textsc{Gc} SMI variants( \textsc{Gcmi} and \textsc{Div-Gcmi}) do not perform well on MNIST classification. Finally, we point out in the case of ImageNet, \textsc{Flqmi} performs the best and outperforms \textsc{Flvmi} and \textsc{Logdetmi} -- this is because we do not need to do the partition trick for \textsc{Flqmi} since it is already linear in time complexity. For \textsc{Flvmi} and \textsc{Logdetmi}, we set the number of partitions $p = 50$ for ImageNet. Finally, we do a pairwise $t$-test to compare the performance of the algorithms (\Appref{app:exp_rarecls}) and observe that the \emph{SMI functions (and particularly \textsc{Flvmi} and \textsc{Logdetmi}) statistically significantly outperform all AL baselines.} 

\vspace{-1.5ex}
\subsection{Redundancy} \label{sec:expRedundancy}
\textbf{Custom dataset:}
To simulate a realistic redundancy scenario we create a custom dataset by duplicating $20\%$ of the unlabeled dataset $10 \times$. For CIFAR-10, the number of unique points in the unlabeled set $|\Ucal|=5K$, the initial labeled set $|\Lcal|=500$, $B=500$, whereas for MNIST $|\Ucal|=500$, $|\Lcal|=50$ and $B=50$. For MNIST, we also present the results for $5 \times$ and $20 \times$ in the \Appref{app:exp_redundancy}.

\begin{figure*}
\centering
\includegraphics[width = 14cm, height=1cm]{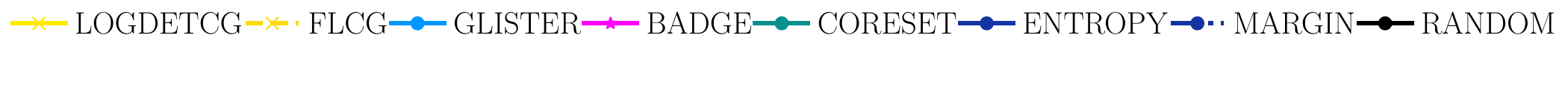}
\centering
\hspace*{-0.6cm}
\begin{subfigure}[t]{0.33\textwidth}
\includegraphics[width = \textwidth]{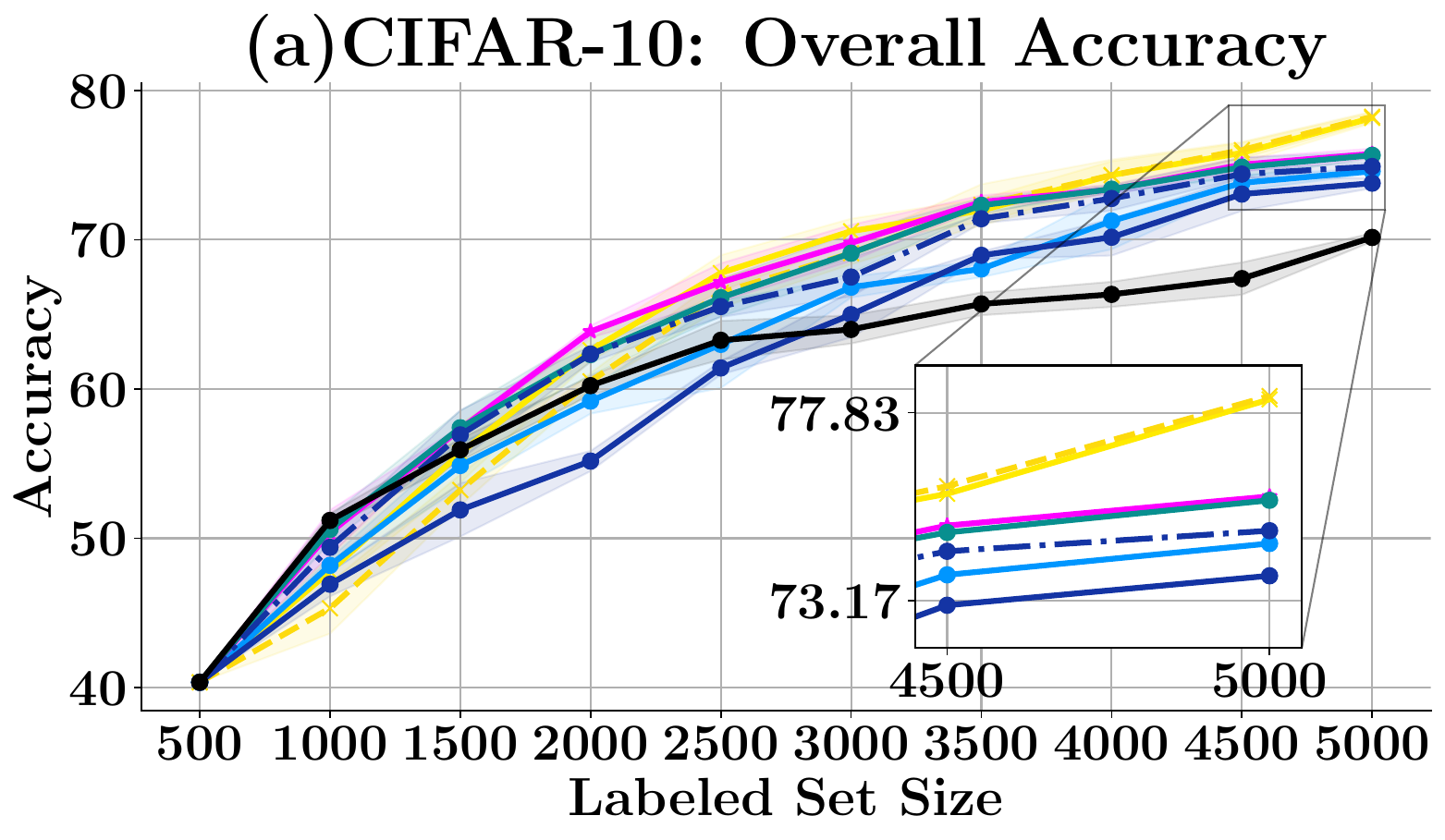}
\end{subfigure}
\begin{subfigure}[t]{0.33\textwidth}
\includegraphics[width = \textwidth]{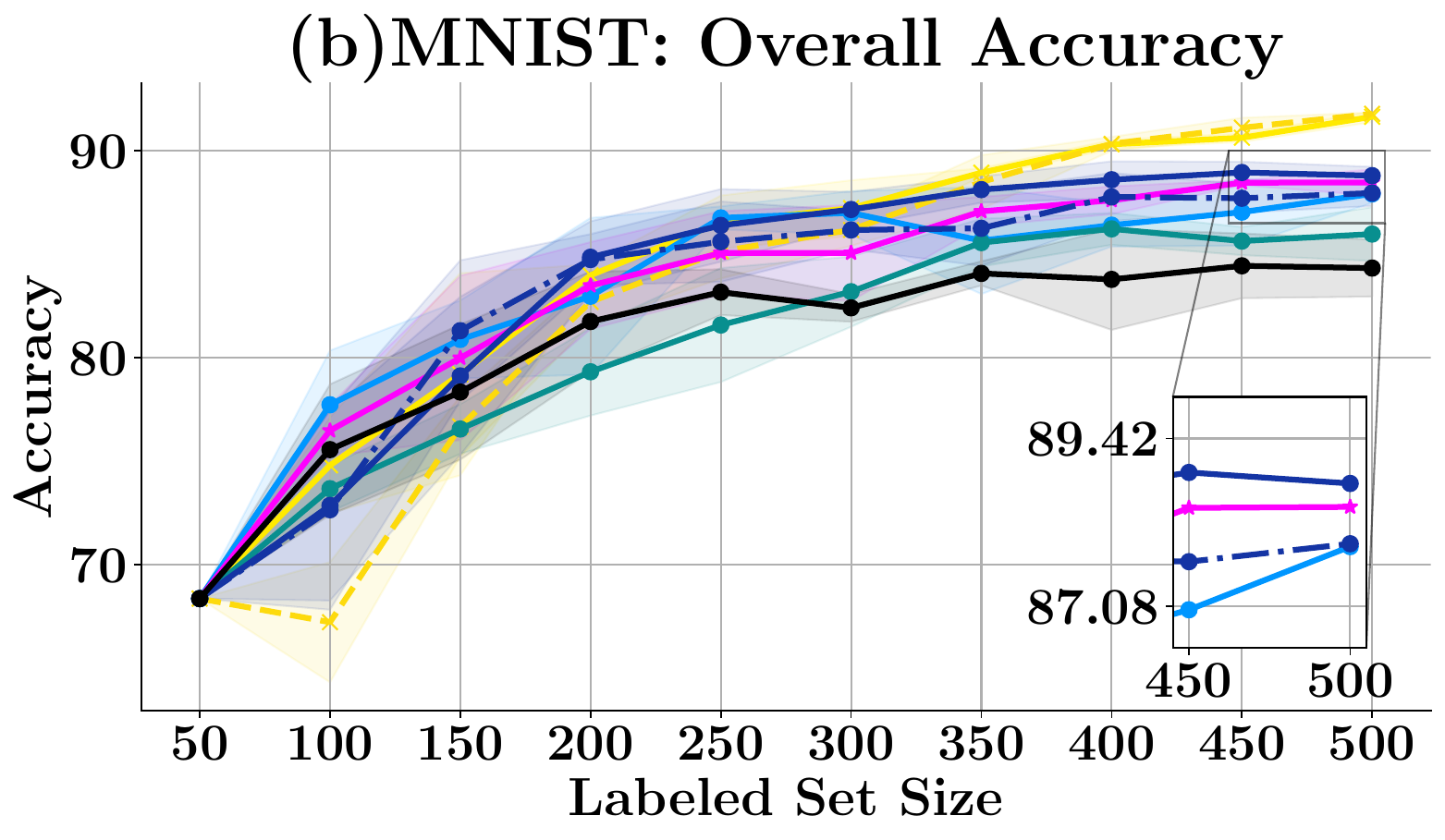}
\end{subfigure}
\begin{subfigure}[t]{0.333\textwidth}
\includegraphics[width = \textwidth]{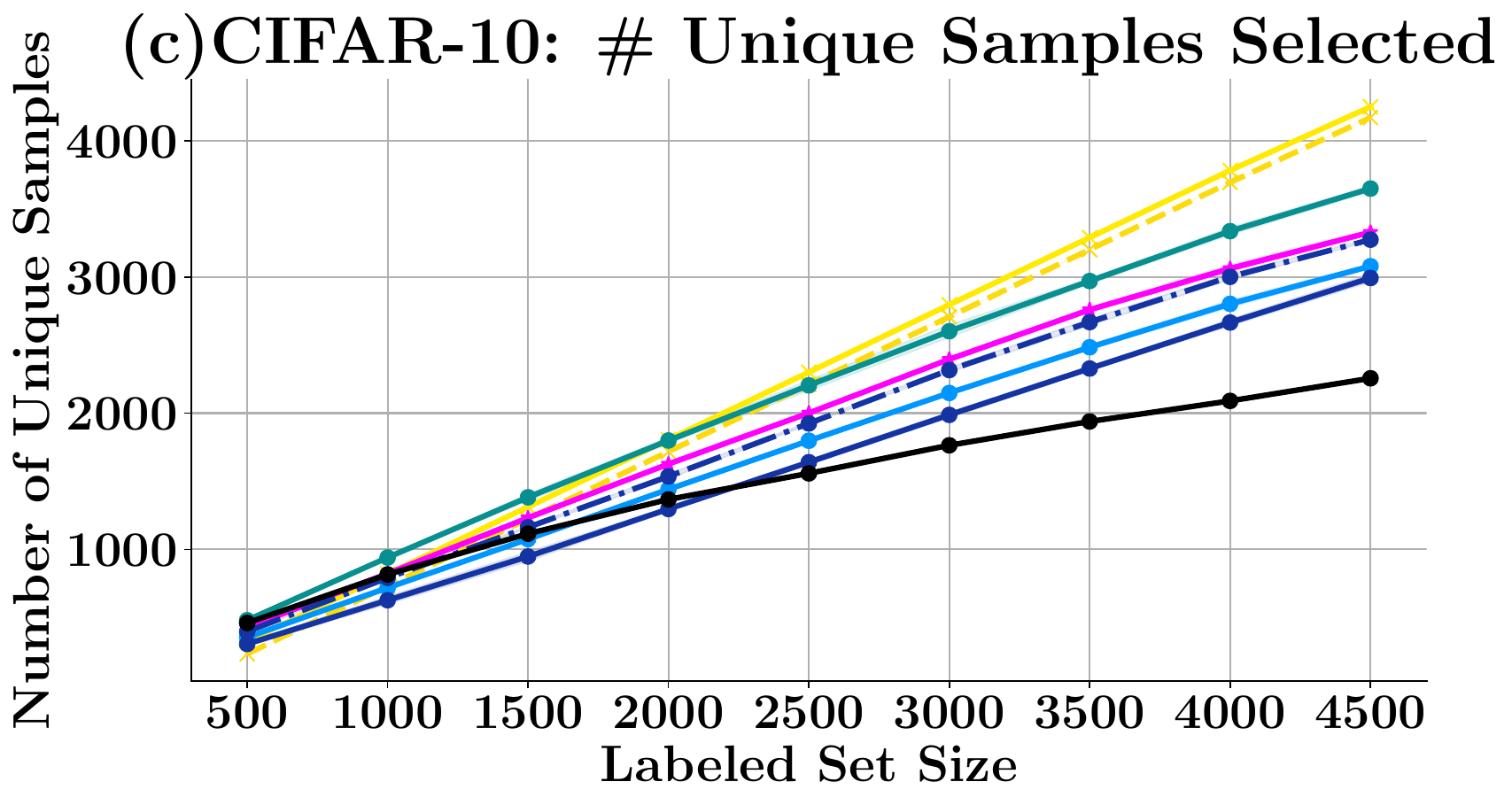}
\end{subfigure}

\caption{Active Learning under $10\times$ redundancy for CIFAR-10 and MNIST. The CG functions (\textsc{Logdetcg, Flcg}) pick more unique points and outperform existing algorithms including \textsc{Badge}.
\vspace{-1ex}
}
\label{fig:res_redundancy}
\end{figure*}

\textbf{SCG vs Baselines:}
As expected, the diversity and uncertainty based methods outperform random. Importantly, we observe that the SCG functions (\textsc{Flcg} and \textsc{Logdetcg}) significantly outperform all baselines by \textbf{$\approx 3-5\%$} towards the end as the conditioning gets stronger with increase in $\Lcal$ (see \figref{fig:res_redundancy}a, 4b). This implies that simply relying on model parameters for diversity and/or uncertainty is not sufficient and that conditioning on the updated labeled set $\Lcal$ (\equref{eq:scg-max-redun}) is required in batch active learning. In \figref{fig:res_redundancy}c we show that SCG based acquisition functions select significantly more unique data points than other baselines. We also perform a pairwise t-test (\Appref{app:exp_redundancy}), to prove that the SCG functions consistently and statistically significantly outperform \textsc{Badge} and other baselines.\looseness-1 
\vspace{-1ex}

\subsection{Out-Of-Distribution} \label{sec:expOOD}
\textbf{Custom dataset:} We simulated a scenario where we convert the classification problem in CIFAR-10 and MNIST to a $8$-class classification, where the first $8$ classes represent the set $\Ical_F$ of in-distribution (ID) data points and the last $2$ represent the set $\Ocal_F$ of out-of-distribution(OOD) data points. The initial labeled set $\Lcal$ \emph{consists only of ID points}, i.e. $\Ocal_F \cap \Lcal = \emptyset$. The unlabeled set is simulated to reflect a realistic and somewhat extreme setting where the unlabeled ID data points $|\Ical_F|$ is much smaller than the unlabeled OOD data points $|\Ocal_F|$. Additionally, we also assume we have a very small validation set of ID points $\Ical_V$. For CIFAR-10: $|\Lcal|=1.6K$, $|\Ical_F|=4K$, $|\Ocal_F|=10K$, $|\Ical_V|=40$, $B=250$ whereas for MNIST which is a relatively simpler task, we use a smaller initial labeled sets 
and keep the unlabeled sets of the same size: $|\Lcal|=40$, $|\Ical_F|=400$, $|\Ocal_F|=10K$, $|\Ical_V|=16$, $B=20$. Recall that our algorithm uses ID set $\Ical$ (initialized to $\Ical_V$) and OOD set $\Ocal$ which we build as follows. Every time our selection approach selects a set $\Acal$, we update $\Ical = \Ical \cup (\Acal \cap \Ical_F)$ and $\Ocal = \Ocal \cup (\Acal \cap \Ocal_F)$, i.e. we augment the ID and OOD points in $\Acal$ to the sets $\Ical$ and $\Ocal$ respectively.\looseness-1

\begin{figure*}
\centering
\includegraphics[width = 14cm, height=1cm]{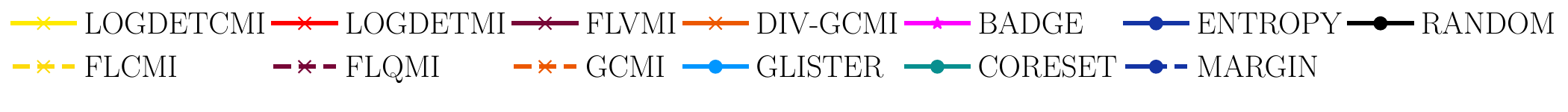}
\centering
\hspace*{-0.6cm}
\begin{subfigure}[t]{0.33\textwidth}
\includegraphics[width = \textwidth]{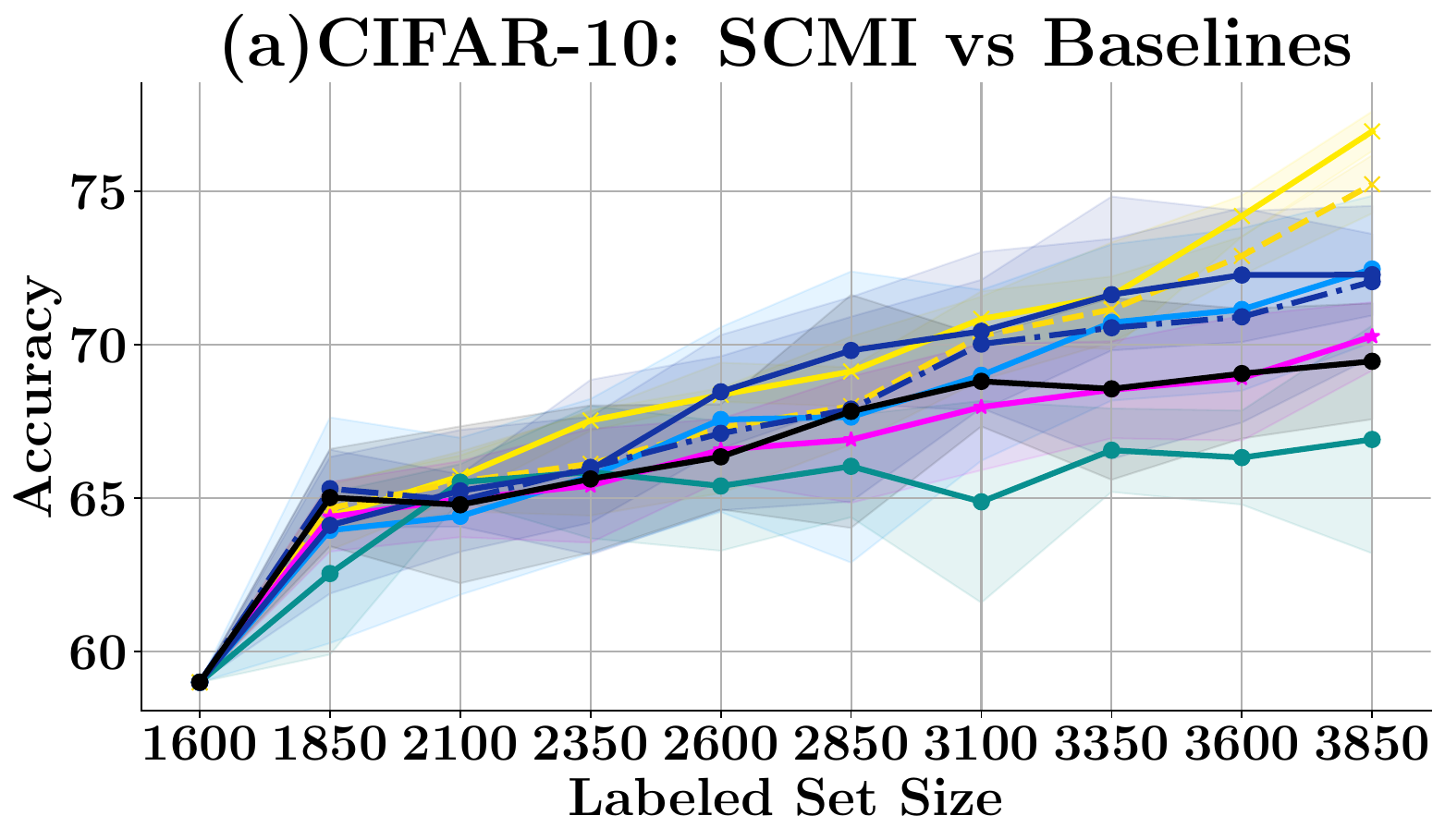}
\end{subfigure} 
\begin{subfigure}[t]{0.33\textwidth}
\includegraphics[width = \textwidth]{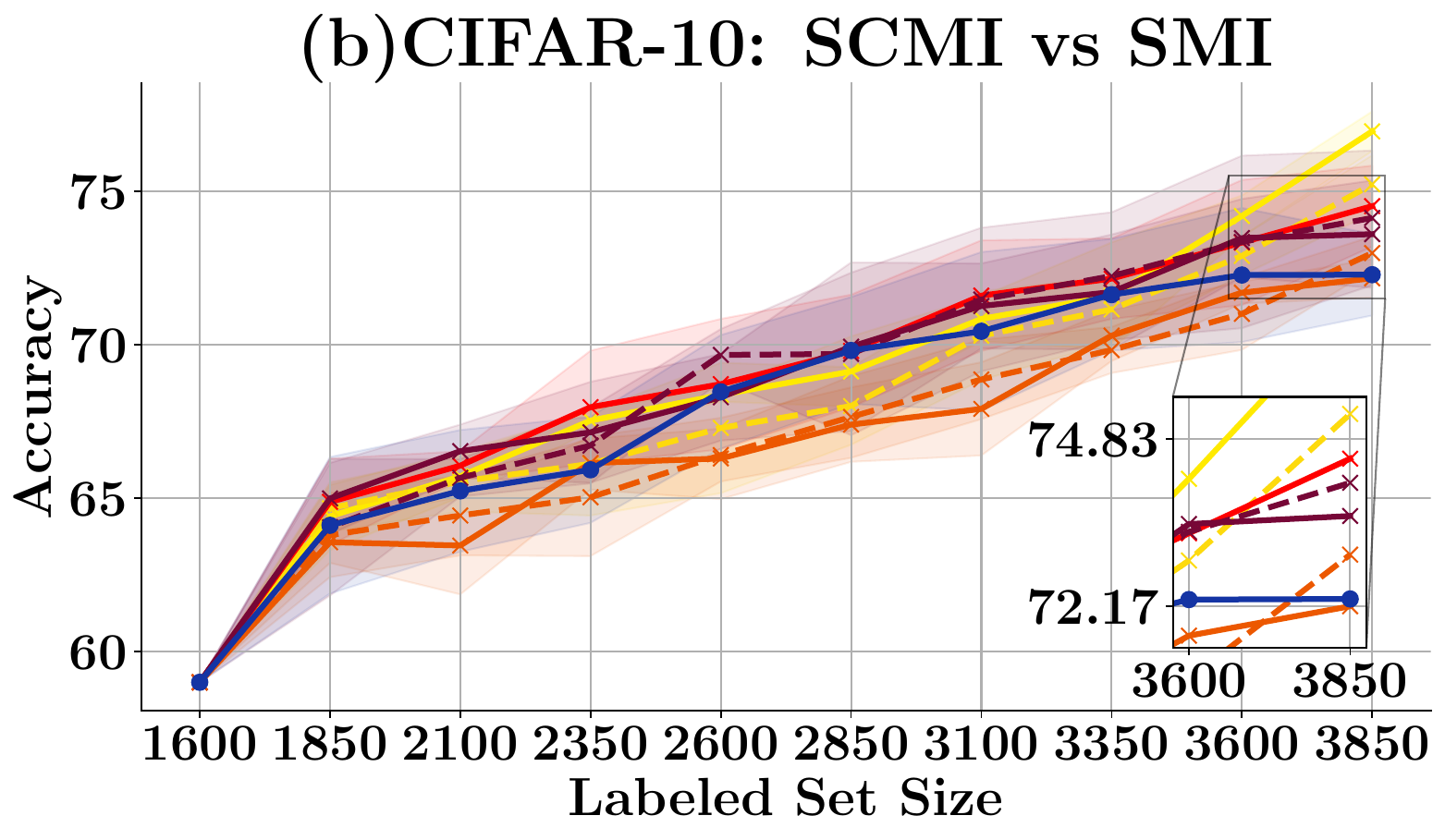}
\end{subfigure}
\begin{subfigure}[b]{0.325\textwidth}
\includegraphics[width = \textwidth]{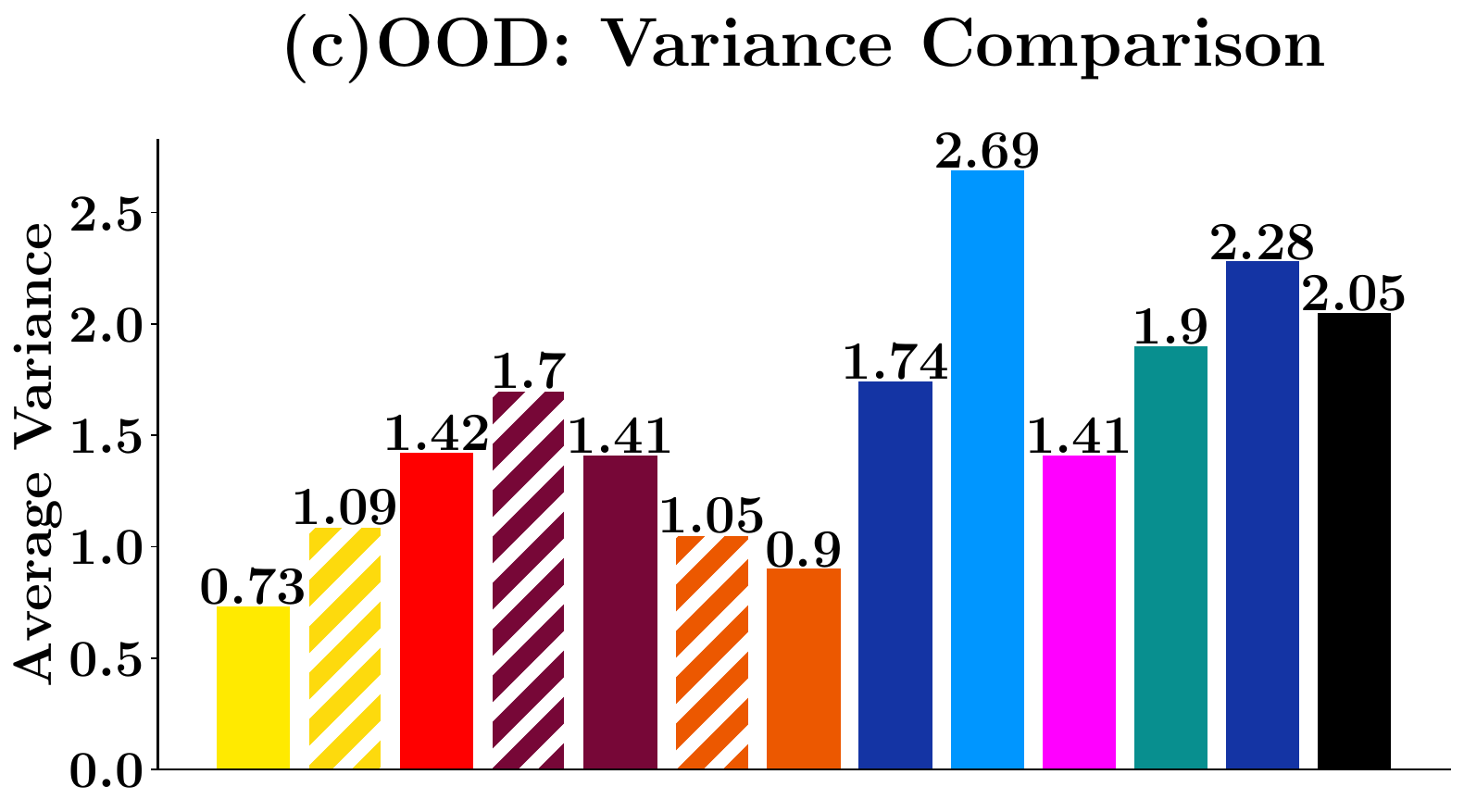}
\end{subfigure}

\begin{subfigure}[t]{0.33\textwidth}
\includegraphics[width = \textwidth]{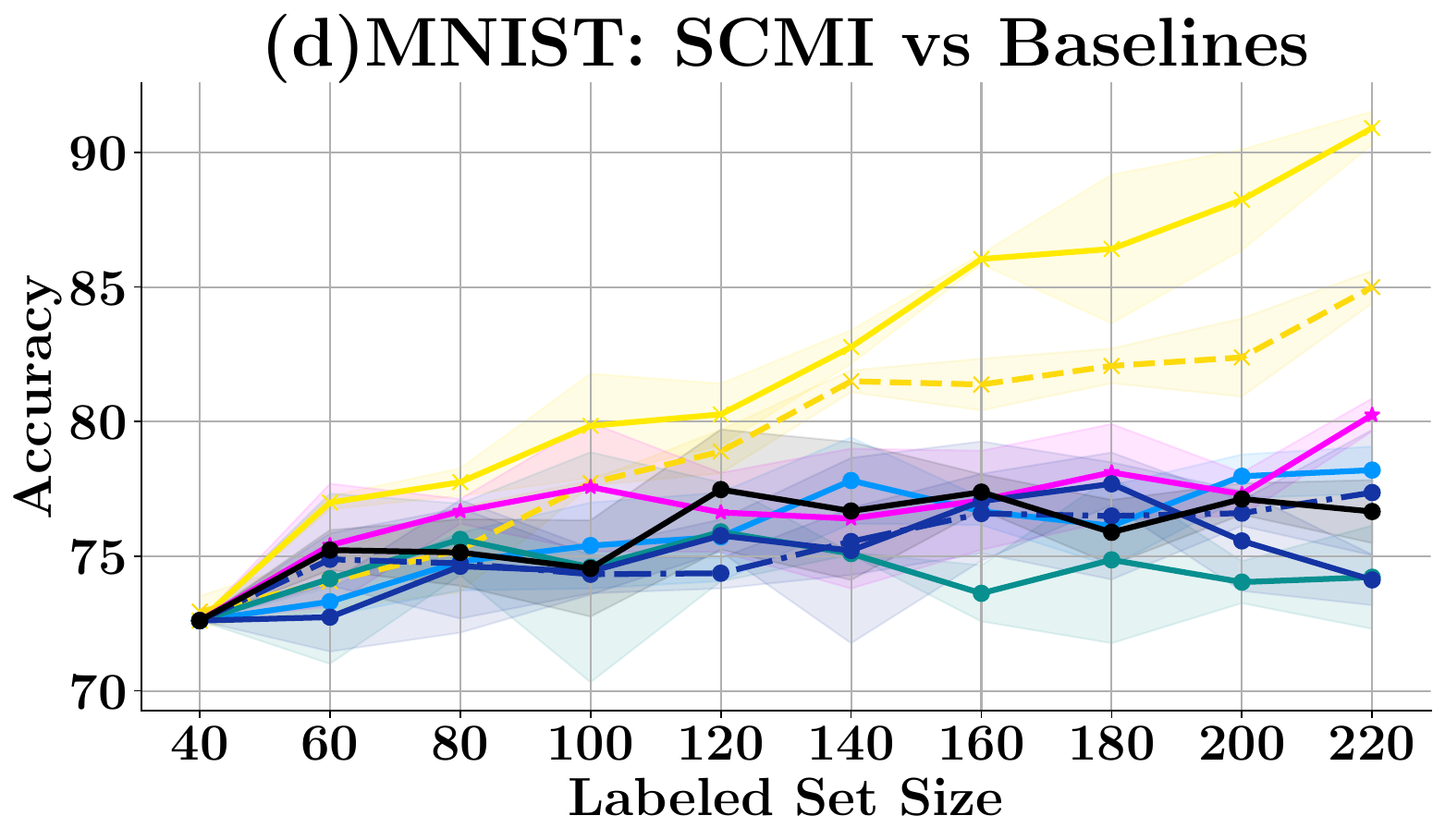}
\end{subfigure} 
\begin{subfigure}[t]{0.33\textwidth}
\includegraphics[width = \textwidth]{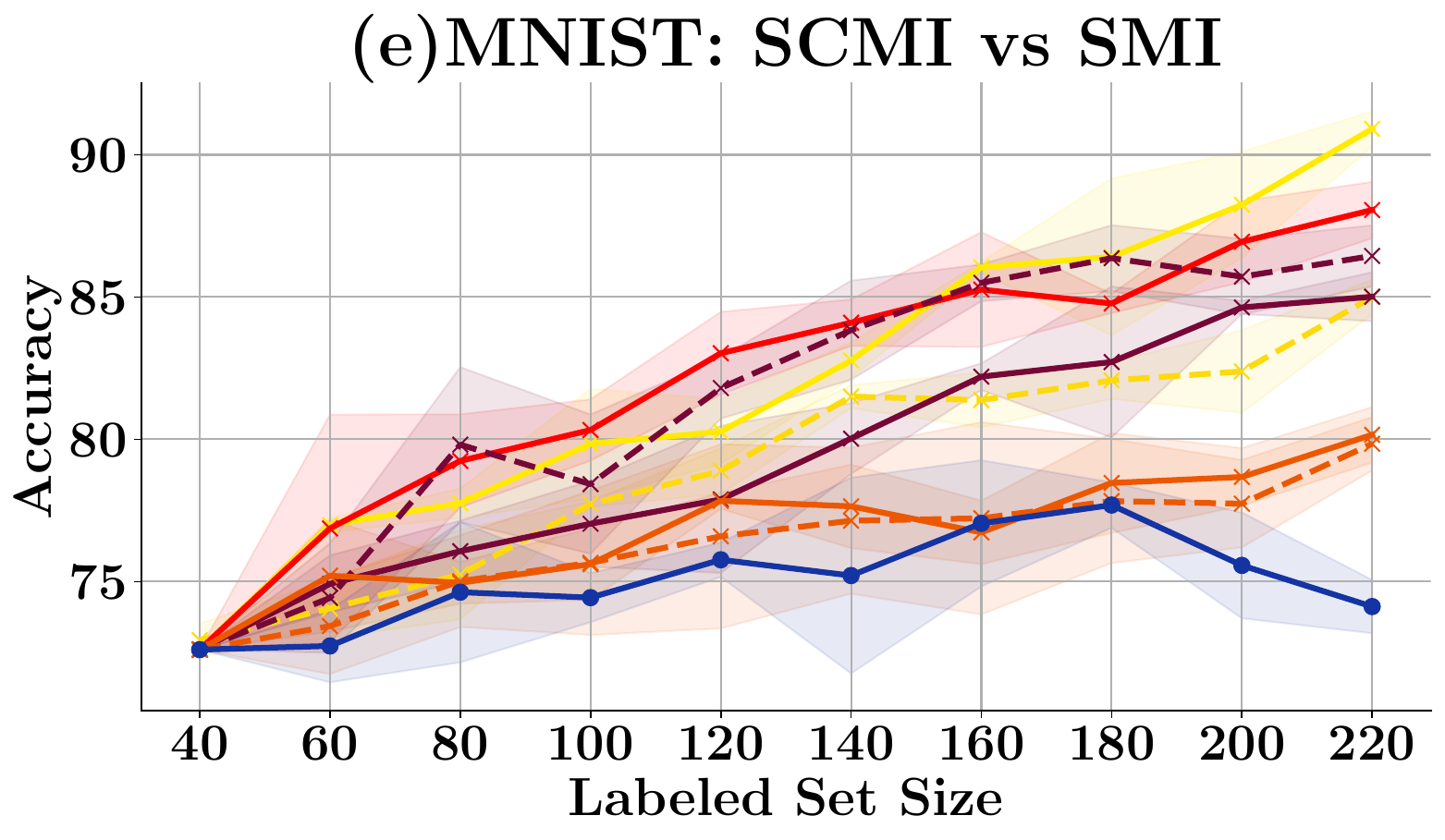}
\end{subfigure}
\begin{subfigure}[t]{0.32\textwidth}
\includegraphics[width = \textwidth]{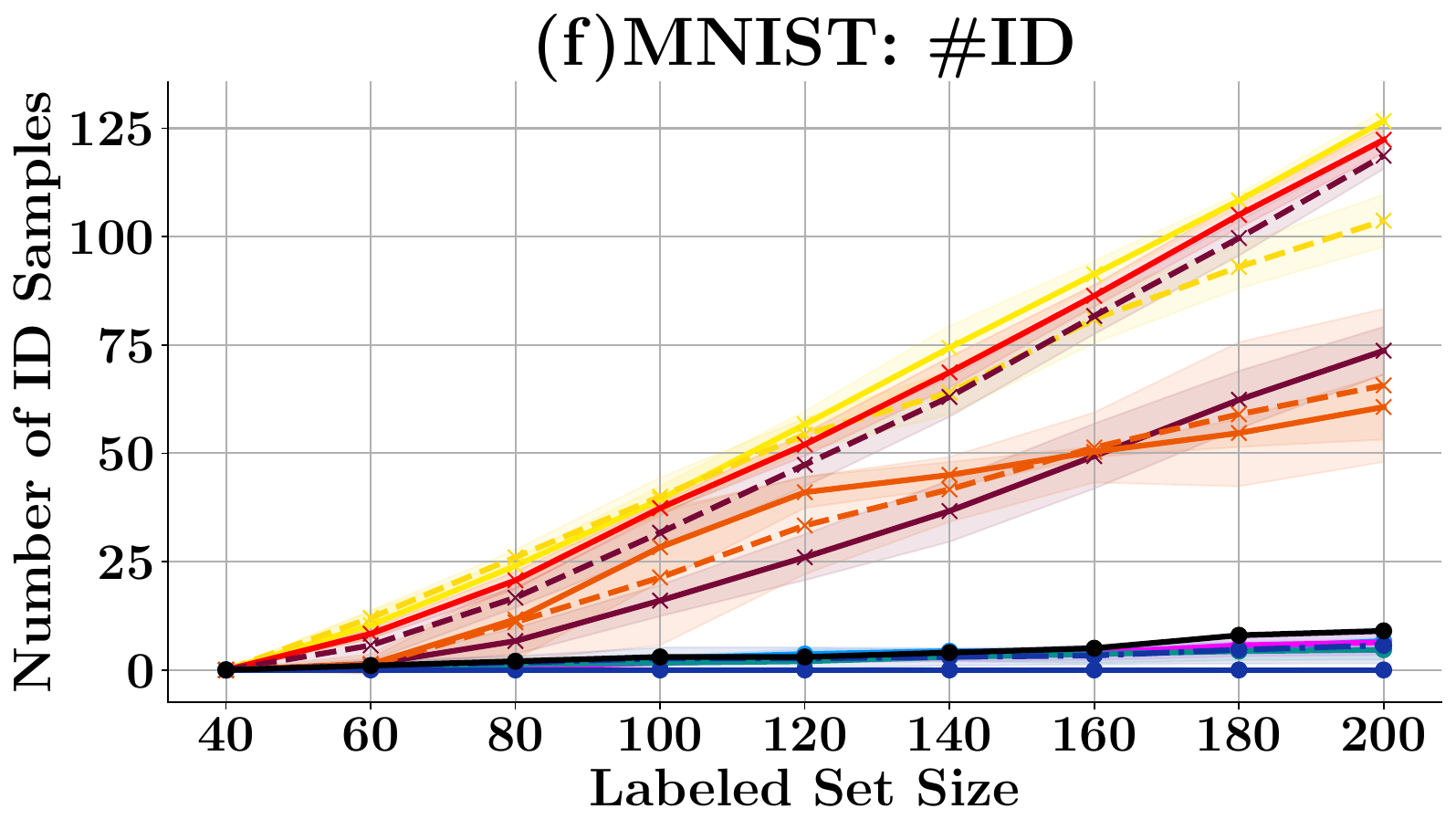}
\end{subfigure}
\caption{Active Learning with OOD data in unlabeled set. Top row: CIFAR-10 results for (a) SCMI vs Baselines, (b) SCMI vs SMI, and (c) variance comparison of different baselines, bottom row: MNIST results for (d) SCMI vs Baselines, (e) SCMI vs SMI, and (f) Number of ID points selected. We see that, i) the SCMI functions consistently outperform the baselines by $5\% - 10\%$, ii) SCMI functions outperform the corresponding SMI functions for later rounds, and (iii) SCMI functions have the least variance compared to the rest, showing that they are more robust in performance.\vspace{-3ex}}
\label{fig:res_ood}
\end{figure*}

\textbf{SCMI vs Baselines:}
Since we care about the predictive performance of the ID classes, we report the ID classes accuracy. We see that SCMI based acquisition functions significantly outperform existing AL approaches by \textbf{$\approx 5 - 10\%$} (see \figref{fig:res_ood}a, 5d). We also observe that existing acquisition functions have a high variance, which is undesirable in real-world deployment scenarios where deep models are being continuously developed. Our SCMI based acquisition functions (\textsc{Logdetcmi} and \textsc{Flcmi}) show the lowest variance in training (see  \figref{fig:res_ood}c). This reinforces the need of having a framework like \textsc{Similar} that facilitates query and conditioning sets. 

\textbf{SCMI vs SMI:}
We compare SCMI functions against SMI functions to study the effect of conditioning and observe that the SCMI functions are comparable to the SMI functions initially but in the later selection rounds of active learning, the SCMI functions consistently outperform SMI functions. In particular, we see an improvement of $2-3\%$ as the conditioning becomes stronger (see \figref{fig:res_ood}b, 5e). We also observe the SCMI tends to select more ID points than SMI and other baselines (see \figref{fig:res_ood}f), and SCMI functions have a lower variance overall compared to even the SMI functions (\figref{fig:res_ood}c).
\vspace{-2ex}

\section{Conclusion}
\vspace{-1ex}
In this paper, we proposed a unified active learning framework \textsc{Similar} using the submodular information functions. We showed the applicability of the framework in three realistic scenarios for active learning, namely rare classes, redundancy, and out of distribution data. In each case, we observed that the functions in \textsc{Similar} significantly outperform existing baselines in each of these tasks. Our real-world experiments on MNIST, CIFAR-10, and ImageNet show that many of the SIM functions (specifically the \textsc{Logdet} and \textsc{Fl} variants) yield  $\approx 5\% - 18\%$ gain compared to existing baselines, particularly in the rare class scenario and $\approx 5\% - 10\%$ OOD scenarios. The main limitations of our work is the dependence on good representations to compute similarity. A potential negative societal impact of this work is the use of \textsc{Similar} to perpetuate certain biases through a malicious use of the query and conditioning set. We discuss this in more detail in \Appref{app:imp_lim}. 



\section*{Acknowledgments and Disclosure of Funding}
This work is supported by the National Science Foundation under Grant No. IIS-2106937, a startup grant from UT Dallas, and by a Google and Adobe research award. Any opinions, findings, and conclusions or recommendations expressed in this material are those of the authors and do not necessarily reflect the views of the National Science Foundation, Google or Adobe.

\bibliography{main}
\newpage

\appendix

\setcounter{page}{1}

\begin{center}
\part{Supplementary Material for SIMILAR: Submodular Information Measures Based Active Learning In Realistic Scenarios} 
\end{center}
\parttoc 
\newpage

\section{Computational Aspects of SIM Functions in \textsc{Similar}}\label{app:computational_aspect}

\subsection{Computational complexity for selection using each function in SMI and baselines}

        

Below, we provide a detailed analysis of the complexity of creating and optimizing the different SIM functions. Denote $|\Xcal|$ as the size of set $\Xcal$. Also, let $|\Ucal| = n$ (the ground set size, which is the size of the unlabeled set in this case). In the main paper, we provided the high-level intuition of the complexity, ignoring the terms of $|\Pcal|$ and $|\Qcal|$ since they would be typically much smaller than the number of unlabeled points $n$. For completeness, we provide the detailed complexity below:
\begin{itemize}[leftmargin=*]
    \item \textbf{Facility Location: } We start with FLVMI. The complexity of creating the kernel matrix is $O(n^2)$. The complexity of optimizing it is $\tilde{O}(n^2)$ (using memoization~\cite{iyer2019memoization})\footnote{$\tilde{O}$: Ignoring log-factors} if we use the stochastic greedy algorithm~\cite{mirzasoleiman2015lazier} and $O(n^2k)$ with the naive greedy algorithm. The overall complexity is $\tilde{O}(n^2)$.
    For FLQMI, the cost of creating the kernel matrix is $O(n|\Qcal|)$, and the cost of optimization is also $\tilde{O}(n|\Qcal|)$ (with naive greedy, it is $O(nB |\Qcal|)$). The complexity of FLCG is $O([n + |\Pcal|]^2)$ to compute the kernel matrix and $\tilde{O}(n^2)$ for optimizing (using the stochastic greedy algorithm). Finally, for FLCMI, the complexity of computing the kernel matrix is $O([n + |\Qcal| + |\Pcal|]^2)$, and the complexity of optimization is $\tilde{O}(n^2)$.
    \item \textbf{Log-Determinant: } We start with LogDetMI. The complexity of the kernel matrix computation (and storage) is $O(n^2)$. The complexity of optimizing the LogDet function using the stochastic greedy algorithm is $\tilde{O}(B^2 n)$, so the overall complexity is $\tilde{O}(n^2 + B^2n)$. For LogDetCG, the complexity of computing the matrix is $O([n + |\Pcal|]^2$, and the complexity of optimization is $\tilde{O}([B + |\Pcal|]^2 n)$. For the LogDetCMI function, the complexity of computing the matrix is $O([n + |\Pcal| + |\Qcal|]^2$, and the complexity of optimization is $\tilde{O}([B + |\Pcal| + |\Qcal|]^2 n)$. 
    \item \textbf{Graph-Cut: } Finally, we study GC functions. For GCMI, we require a $O(n|\Qcal|)$ kernel matrix, and the complexity of the stochastic greedy algorithm is also $\tilde{O}(n|\Qcal|)$. Finally, for GCCG, the complexity of creating the kernel matrix is $O(n|^2 + n|\Pcal|)$, and the complexity of the stochastic greedy algorithm is $\tilde{O}(n^2 + n|\Pcal|)$. 
\end{itemize}
We end with a few comments. First, most of the complexity analysis above is with the stochastic greedy algorithm~\cite{mirzasoleiman2015lazier}. If we use the naive or lazy greedy algorithm, the worst-case complexity is a factor $B$ larger. Secondly, we ignore log-factors in the complexity of stochastic greedy since the complexity is actually $O(n\log 1/\epsilon)$, which achieves an $1 - 1/e - \epsilon$ approximation. Finally, the complexity of optimizing and constructing the FL, LogDet, and GC functions can be obtained from the CG versions by setting $\Pcal = \emptyset$.

\subsection{Details on Partitioning Approach}
In some of our experiments, we choose to partition the unlabeled set into chunks in order to meet the scale of the dataset used in that experiment. This is because many of the techniques (specifically LogDet functions, FLVMI, FLCG, FLCMI, GCCG) all have $O(n^2)$ space complexity. For $n$ in the range of a few million to a few billion data points (which is not uncommon in big-data applications today), we need to scale our algorithms to be linear in $n$ and not quadratic. For this, we propose a simple partitioning approach where the unlabeled data is chunked into $p$ partitions. In this strategy, we perform unlabeled instance acquisition on each chunk using a proportional fraction of the full AL batch size. The most notable example of the use of our partitioning strategy is in our down-sampled ImageNet experiment. By performing AL acquisition on the full unlabeled set, almost all AL strategies exhaust the available compute resources. Hence, to execute most of our AL strategies, we partitioned the unlabeled set into 50 equally sized chunks, so each partition has around 10k to 20k instances. As $n$ grows, the number of partitions would also grow so that $n/p$ is roughly constant. The complexity of most approaches discussed above would then be $O(n^2/p)$ ($O(n^2/p^2)$ for each chunk, repeated $p$ times), and if $n/p = r$ is a constant, then the complexity $O(nr)$ would be linear in $n$. We then acquire a number of unlabeled instances from each chunk whose ratio with the full AL batch size is equal to the ratio between the chunk size and the full unlabeled set. The acquired instances from each chunk are then combined to form the full acquired set of unlabeled instances.

\section{More Details on Experimental Setup, Datasets, and Baselines}\label{app:experimental}

\subsection{Datasets description in each scenario}
We used various standard datasets -- namely, MNIST, CIFAR10, and ImageNet -- to demonstrate the effectiveness and robustness of \textsc{Similar}. We also provide additional experiments on SVHN in sections below. We use standard sources for all datasets. As previously mentioned, we perform our experiments on a down-sampled version of ImageNet. Beyond the fact that each image is now $32\times32$, the data set is otherwise identical. Moreover, we find that the provided validation set is often used as the test set in most evaluations on down-sampled ImageNet. The down-sampled ImageNet training set can be procured \href{https://image-net.org/data/downsample/Imagenet32_train.zip}{{\color{blue} here}}, and the validation set can be found \href{https://image-net.org/data/downsample/Imagenet32_val.zip}{{\color{blue} here}}. Note that associated licenses for all datasets apply.

\paragraph{Rare classes setting:}In \tabref{tab:numpoints-rarecls}, we show the exact initial splits used in our experiments for the rare classes scenario. In CIFAR-10, ImageNet, and SVHN, we use randomly chose half the number of classes as imbalanced and the other half as balanced. Following \cite{gudovskiy2020deep}, we chose classes $(5, \cdots 9)$ as imbalanced classes in MNIST. We use an AL batch size of $125$ for the CIFAR-10, MNIST and SVHN datasets. We use the same data setting for the CIFAR-10 and SVHN datasets with an imbalance factor $\rho=20$. The results for SVHN are in \Appref{app:exp_rarecls}. For MNIST, we additionally show results for $\rho=100$ in \Appref{app:exp_rarecls}. Due to the scale of down-sampled ImageNet and the natural imbalance present in its full training set, we adopt a different dataset splitting strategy. Following \cite{gudovskiy2020deep}, we randomly chose 500 classes (half) as rare classes. Our train set is initialized as having 34 examples per rare class and 170 examples per normal class. Our validation set contains 5 examples per class, making it balanced. The unlabeled set is created to have 1 rare example for every 5 normal examples. In all, our initialization leads our initial train set, validation set, and unlabeled set to have approximately 100k, 5k, and 660k points, respectively. We use an AL batch size of 25k points, and we use the same training conditions as before. However, we perform AL selection by dividing the unlabeled set into chunks (partitions), selecting a proportionate fraction of the AL batch size from each. In this case, we divide the unlabeled set into 50 to 100 partitions (determined by compute limitations) and perform selection on each partition.


\begin{table}[h]
\centering
\begin{tabular}{|l|r|r|r|r|}
\hline
Dataset                                                                  & \multicolumn{1}{l|}{\begin{tabular}[c]{@{}l@{}}Imbalance \\ factor ($\rho$)\end{tabular}} & \multicolumn{1}{l|}{\begin{tabular}[c]{@{}l@{}}Labeled\\ (per class)\end{tabular}} & \multicolumn{1}{l|}{\begin{tabular}[c]{@{}l@{}}Valid\\ (per class)\end{tabular}} & \multicolumn{1}{l|}{\begin{tabular}[c]{@{}l@{}}Unlabeled\\ (per class)\end{tabular}} \\ \hline
\multirow{2}{*}{\begin{tabular}[c]{@{}l@{}}CIFAR-10\\ SVHN\end{tabular}} & \multirow{2}{*}{20}                                                              & 3                                                                                  & 5                                                                                & 150                                                                                  \\ \cline{3-5} 
                                                                         &                                                                                  & 22                                                                                 & 5                                                                                & 3000                                                                                 \\ \hline
\multirow{4}{*}{MNIST}                                                   & \multirow{2}{*}{20}                                                              & 3                                                                                  & 5                                                                                & 200                                                                                  \\ \cline{3-5} 
                                                                         &                                                                                  & 22                                                                                 & 5                                                                                & 4000                                                                                 \\ \cline{2-5} 
                                                                         & \multirow{2}{*}{100}                                                             & 3                                                                                  & 5                                                                                & 40                                                                                   \\ \cline{3-5} 
                                                                         &                                                                                  & 22                                                                                 & 5                                                                                & 4000                                                                                 \\ \hline
\end{tabular}
\caption{Number of data points for each dataset in the rare classes scenario. For CIFAR-10, MNIST, and SVHN, we use 5 balanced classes and 5 imbalanced classes. In the main paper, we show experiments for $\rho=20$. In \Appref{app:exp_rarecls}, we show experiments for $\rho=100$.}
\label{tab:numpoints-rarecls}
\end{table}

\paragraph{Redundancy setting:} In \tabref{tab:numpoints-redundancy}, we show the exact initial splits used in our experiments for the redundancy scenario. For CIFAR-10 and SVHN, we use the same setting. Since MNIST classification is a relatively simpler problem, we use one tenth of the data points used in the CIFAR-10 setting. For all datasets, we create the unlabeled dataset by duplicating $20\%$ of the unlabeled dataset RF $\times$. We denote RF as the redundancy factor. For instance, we consider 5000 unique points and duplicate $20\%$ of them $10 \times$ in CIFAR-10. This gives us $(5000 \times 0.2 \times 10=10000)$ duplicated points and $(5000-(5000 \times 0.2 \times 10)=4000)$ original points for a total of $(10000+4000=14000)$ points.

\begin{table}[h]
\centering
\begin{tabular}{|l|r|l|l|}
\hline
Dataset        & \multicolumn{1}{l|}{\begin{tabular}[c]{@{}l@{}}Total Unique\\ Points\end{tabular}} & \begin{tabular}[c]{@{}l@{}}Fraction of points\\ duplicated\end{tabular} & \begin{tabular}[c]{@{}l@{}}Number of \\ duplicated points\end{tabular} \\ \hline
CIFAR-10, SVHN & 5000                                                                               & 20\%                                                                    & 5000*0.2*RF                                                            \\ \hline
MNIST          & 500                                                                                & 20\%                                                                    & 500*0.2*RF                                                             \\ \hline
\end{tabular}
\caption{Number of data points for each dataset in the redundancy scenario. RF here is the redundancy factor. In the main paper, we show experiments for RF=$10 \times$. In \Appref{app:exp_redundancy}, we show experiments for RF=$5 \times$ and RF=$20 \times$.}
\label{tab:numpoints-redundancy}
\end{table}

\paragraph{Out-of-distribution setting: } In \tabref{tab:numpoints-ood}, we show the exact initial splits used in our experiments for the out-of-distribution scenario. In all datasets, we chose the first 8 classes to be in-distribution (ID) and the last 2 classes to be out-of-distribution (OOD). Initially, the labeled set consists of only ID points. The unlabeled set is designed to reflect a realistic setting with high number of OOD points. For CIFAR-10, we use 200 points per ID class in the labeled set and 500 points per ID class, 5000 points per OOD class in the unlabeled set. This gives us an initial labeled set of size $200 \times 8=1600$ and an initial unlabeled set of size $500 \times 8 + 5000 \times 2= 14000$. We make the task slightly challenging for MNIST by further decreasing the number of ID points in the unlabeled dataset as shown in \tabref{tab:numpoints-ood}.  

\begin{table}[h]
\centering
\begin{tabular}{|l|l|r|r|r|}
\hline
Dataset                   &            & \multicolumn{1}{l|}{\begin{tabular}[c]{@{}l@{}}Labeled\\ (per class)\end{tabular}} & \multicolumn{1}{l|}{\begin{tabular}[c]{@{}l@{}}Valid\\ (per class)\end{tabular}} & \multicolumn{1}{l|}{\begin{tabular}[c]{@{}l@{}}Unlabeled\\ (per class)\end{tabular}} \\ \hline
\multirow{2}{*}{CIFAR-10} & ID points  & 200                                                                                & 5                                                                                & 500                                                                                  \\ \cline{2-5} 
                          & OOD points & 0                                                                                  & 0                                                                                & 5000                                                                                 \\ \hline
\multirow{2}{*}{MNIST}    & ID points  & 5                                                                                  & 2                                                                                & 50                                                                                   \\ \cline{2-5} 
                          & OOD points & 0                                                                                  & 0                                                                                & 5000                                                                                 \\ \hline
\end{tabular}
\caption{Number of data points for each dataset in the out-of-distribution scenario.}
\label{tab:numpoints-ood}
\end{table}

\subsection{Experimental setup}
We ran experiments using an SGD optimizer with an initial learning rate of 0.01, a momentum of 0.9, and a weight decay of 5e-4. We decay the learning rate via cosine annealing \cite{loshchilov2016sgdr} for each epoch.  For MNIST, we use the LeNet model \cite{lecun1989backpropagation}. For all other datasets, we use ResNet18 model \cite{he2016deep}. For each round of active learning, we train until the accuracy reaches 99\% or the epoch count reaches 150. We run all our experiments on a single V100 GPU.

\subsection{Details on computation of penalty matrix}

The penalty matrices computed in this paper follow the strategy used in~\cite{ash2019deep}. In their strategy, a penalty matrix is constructed for each dataset-model pair. Each cell $(i,j)$ of the matrix reflects the fraction of training rounds that AL with selection algorithm $i$ has higher test accuracy than AL with selection algorithm $j$ with statistical significance. As such, the average difference between the test accuracies of $i$ and $j$ and the standard error of that difference are computed for each training round. A two-tailed $t$-test is then performed for each training round: If $t>t_\alpha$, then $\frac{1}{N_{train}}$ is added to cell $(i,j)$. If $t<-t_\alpha$, then $\frac{1}{N_{train}}$ is added to cell $(j,i)$. Hence, the full penalty matrix gives a holistic understanding of how each selection algorithm compares against the others: A row with mostly high values signals that the associated selection algorithm performs better than the others; however, a column with mostly high values signals that the associated selection algorithm performs worse than the others. As a final note, \cite{ash2019deep} takes an additional step where they consolidate the matrices for each dataset-model pair into one matrix by taking the sum across these matrices, giving a summary of the AL performance for their entire paper that is fairly weighted to each experiment. We present the penalty matrices for each of the settings in the sections below.

\subsection{Licensing details}
\paragraph{Datasets.} Our experiments with \textsc{Similar} utilize the following datasets.

\begin{itemize}
    \item \href{https://www.cs.toronto.edu/~kriz/cifar.html}{{\color{blue}CIFAR-10}}~\cite{krizhevsky2009learning}: MIT License
    \item \href{http://yann.lecun.com/exdb/mnist/}{{\color{blue}MNIST}}~\cite{lecun2010mnist}: Creative Commons Attribution-Share Alike 3.0
    \item \href{http://ufldl.stanford.edu/housenumbers/}{{\color{blue}SVHN}}~\cite{netzer2011reading}: CC0 1.0 Public Domain
    \item \href{https://www.image-net.org/about.php}{{\color{blue}ImageNet}}~\cite{russakovsky2015imagenet}: Custom (Research, Non-Commercial)
\end{itemize}

\paragraph{Repositories.} Our experiments utilize contributions from existing code repositories. Specifically, we utilize the DISTIL repository for AL baselines. We utilize the Fisher Kernel Self-Supervision repository in our usages of \textsc{Fisher} and its variants. We extensively use PyTorch, and we utilize the CORDS repository in our gradient computations. To summarize, the following repositories are used, and their licenses from their original sources are also provided:

\begin{itemize} 
    \item \href{https://pytorch.org/}{{\color{blue}PyTorch}}~\cite{NEURIPS2019_9015}: Modified BSD
    \item \href{https://github.com/decile-team/distil}{{\color{blue}DISTIL}}: MIT License
    \item \href{https://github.com/decile-team/cords}{{\color{blue}CORDS}}: MIT License
    \item \href{https://github.com/gudovskiy/al-fk-self-supervision}{{\color{blue}Fisher Kernel Self-Supervision}}~\cite{gudovskiy2020deep}: (None Listed)
    \item \href{https://github.com/JordanAsh/badge}{{\color{blue}\textsc{Badge}}}~\cite{ash2019deep}: None Listed
\end{itemize}

\subsection{Baselines and Code}
For all baselines, we use code either from existing libraries and codebases or from the authors. For \textsc{Badge}~\cite{ash2019deep}, we use the code from the authors\footnote{\url{https://github.com/JordanAsh/badge}}. Similarly, for the \textsc{Fisher} baseline, we use the code from the authors\footnote{\url{https://github.com/gudovskiy/al-fk-self-supervision}}. For the other methods like entropy sampling, \textsc{Coreset}, etc., we use DISTIL\footnote{\url{https://github.com/decile-team/distil}}, which implements most of the state-of-the-art standard AL approaches building upon the respective authors code. 

\begin{figure}[ht]
\centering
\includegraphics[width =0.7\textwidth]{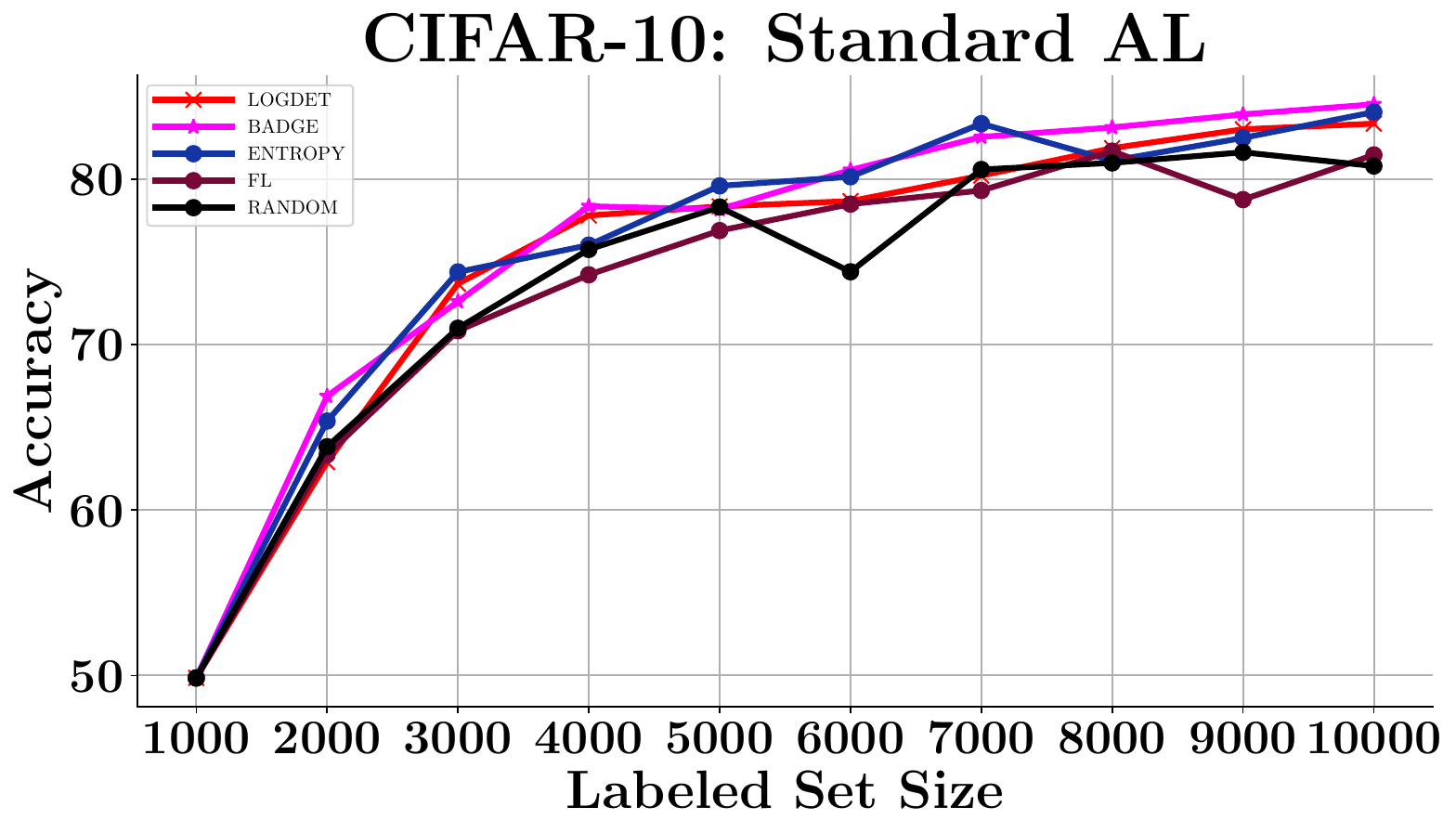}
\caption{Comparison of submodular functions with baselines in a standard active learning setting.
}
\label{standard-al}
\end{figure}

\begin{figure}[ht]
\centering
\includegraphics[width = 14cm, height=1cm]{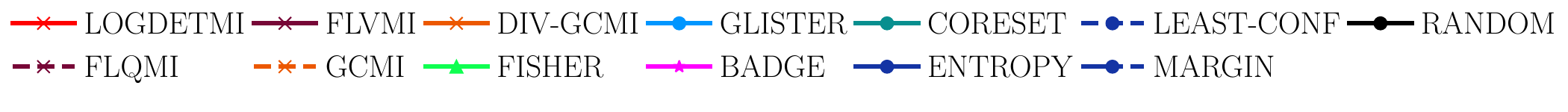}
\centering
\hspace*{-0.6cm}
\begin{subfigure}[t]{0.33\textwidth}
\includegraphics[width = \textwidth]{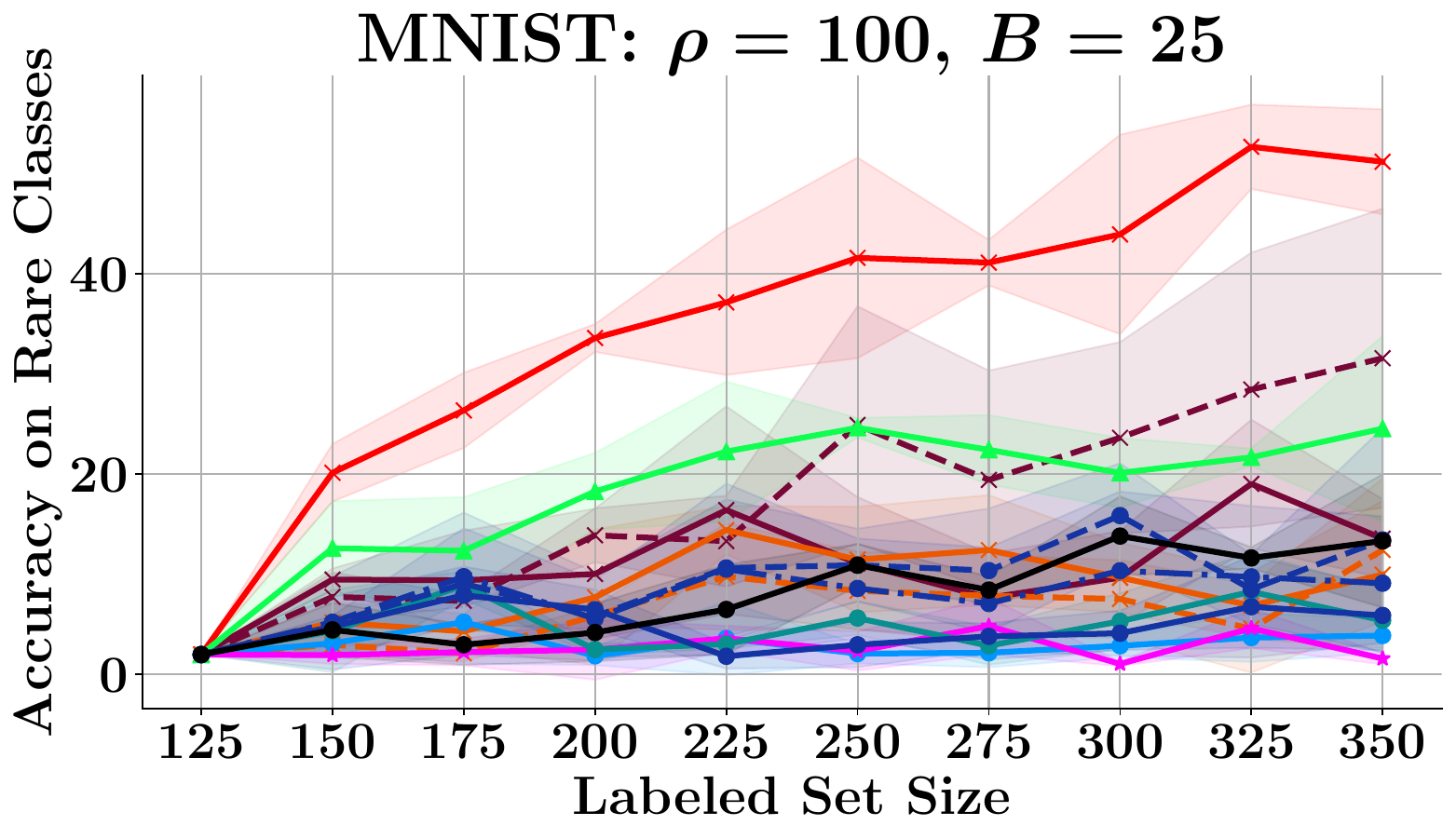}
\end{subfigure} 
\begin{subfigure}[t]{0.33\textwidth}
\includegraphics[width = \textwidth]{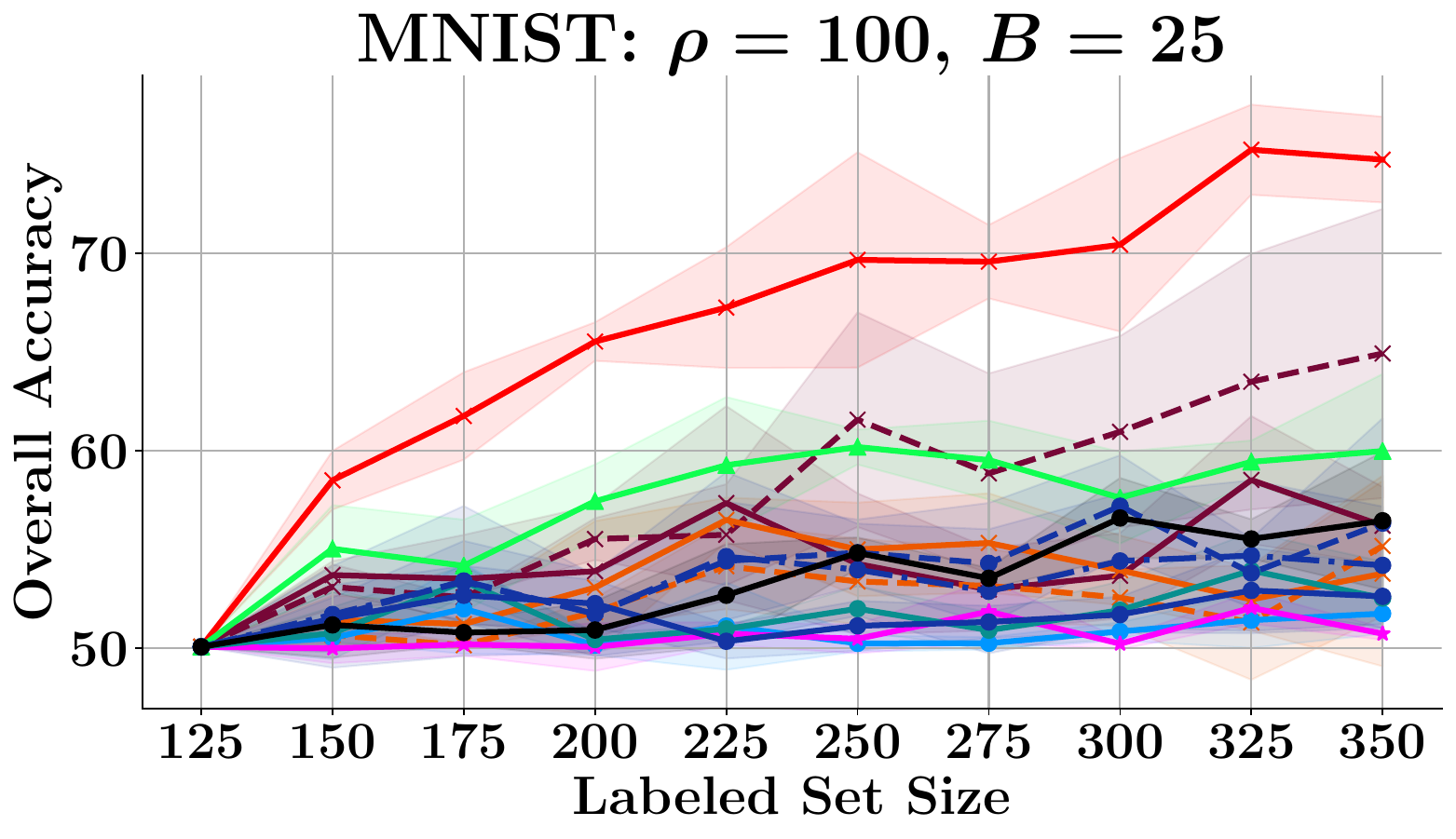}
\end{subfigure}
\begin{subfigure}[t]{0.33\textwidth}
\includegraphics[width = \textwidth]{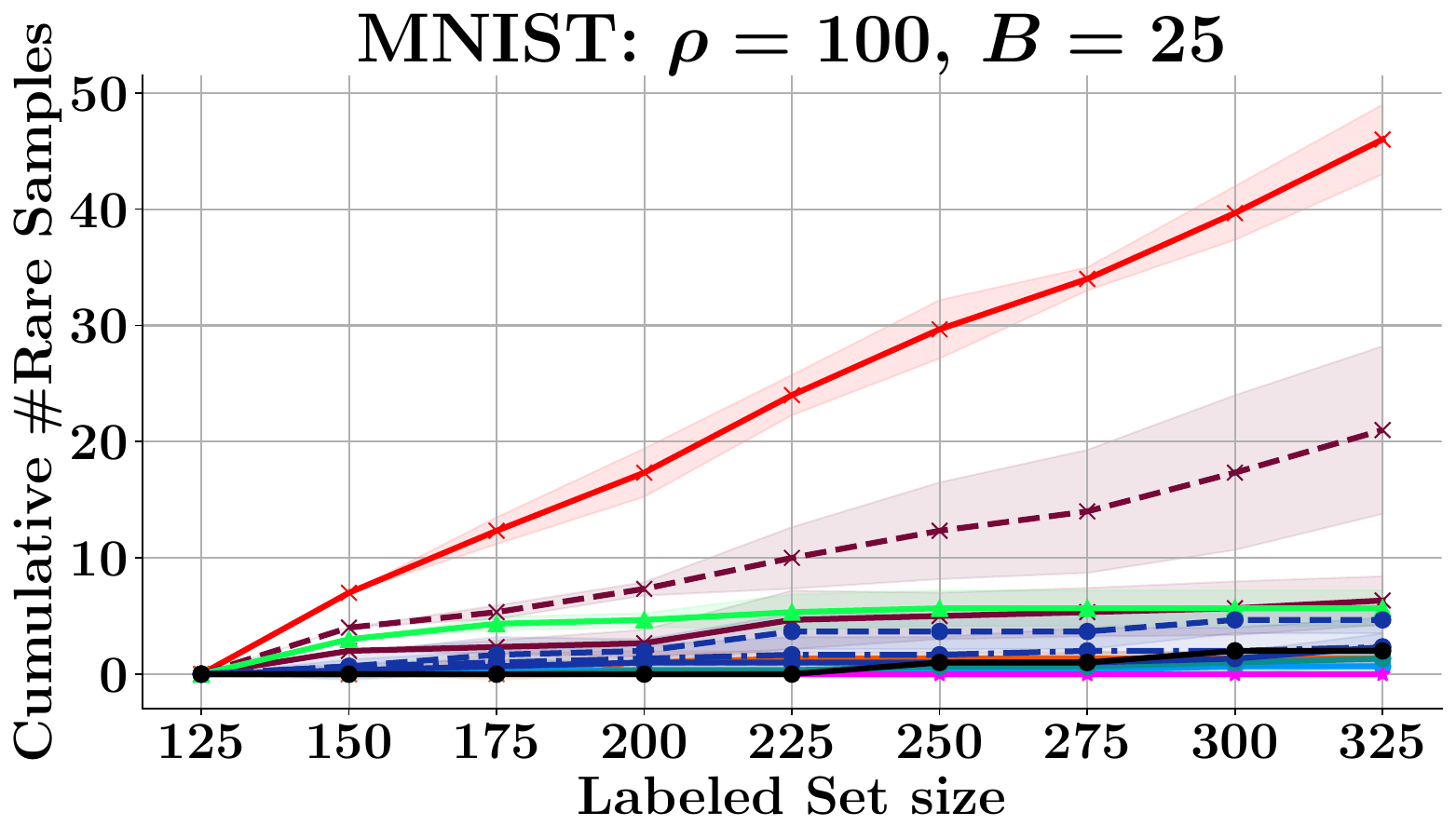}
\end{subfigure}
\begin{subfigure}[b]{0.33\textwidth}
\includegraphics[width = \textwidth]{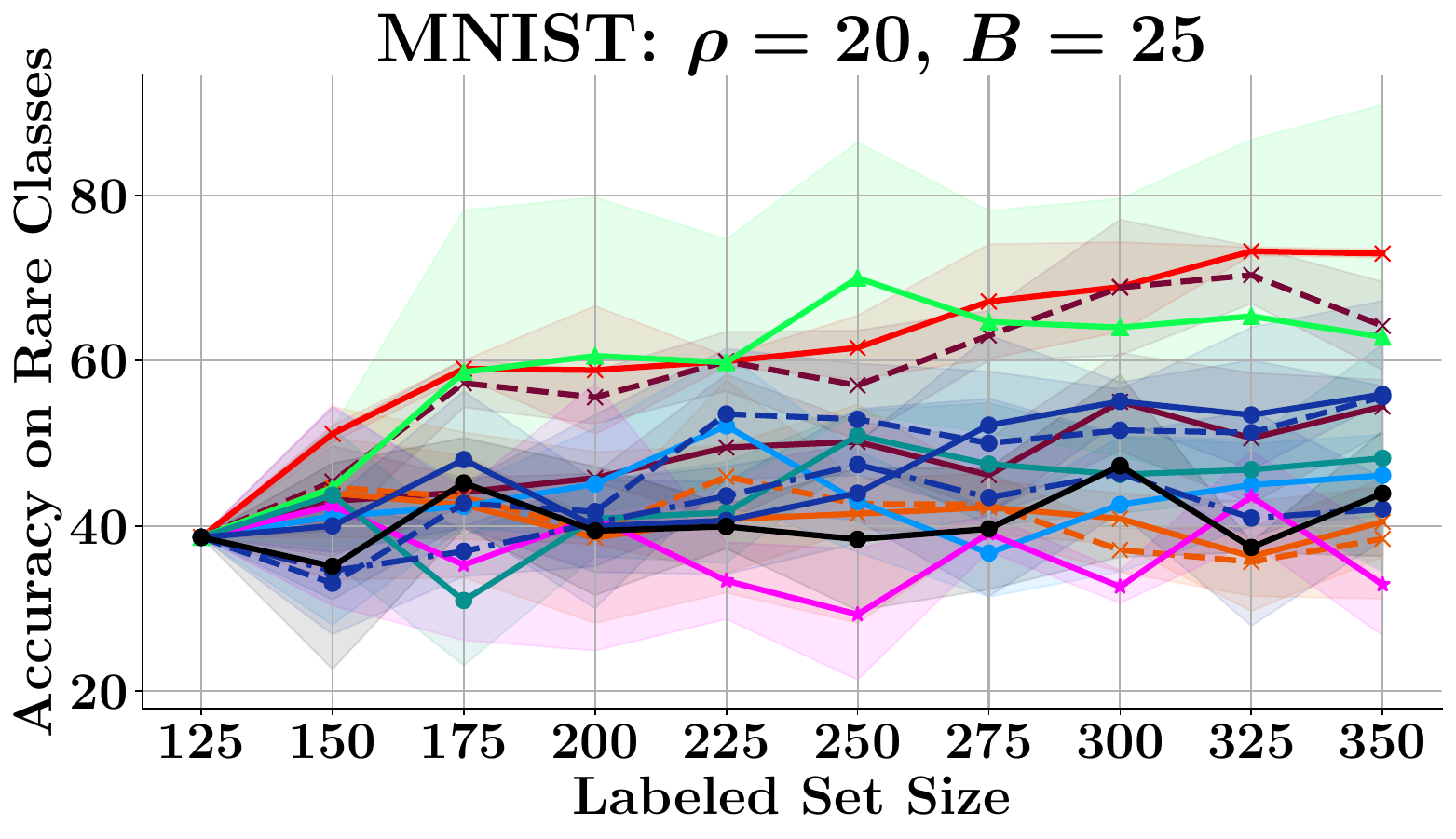}
\end{subfigure}
\begin{subfigure}[t]{0.33\textwidth}
\includegraphics[width = \textwidth]{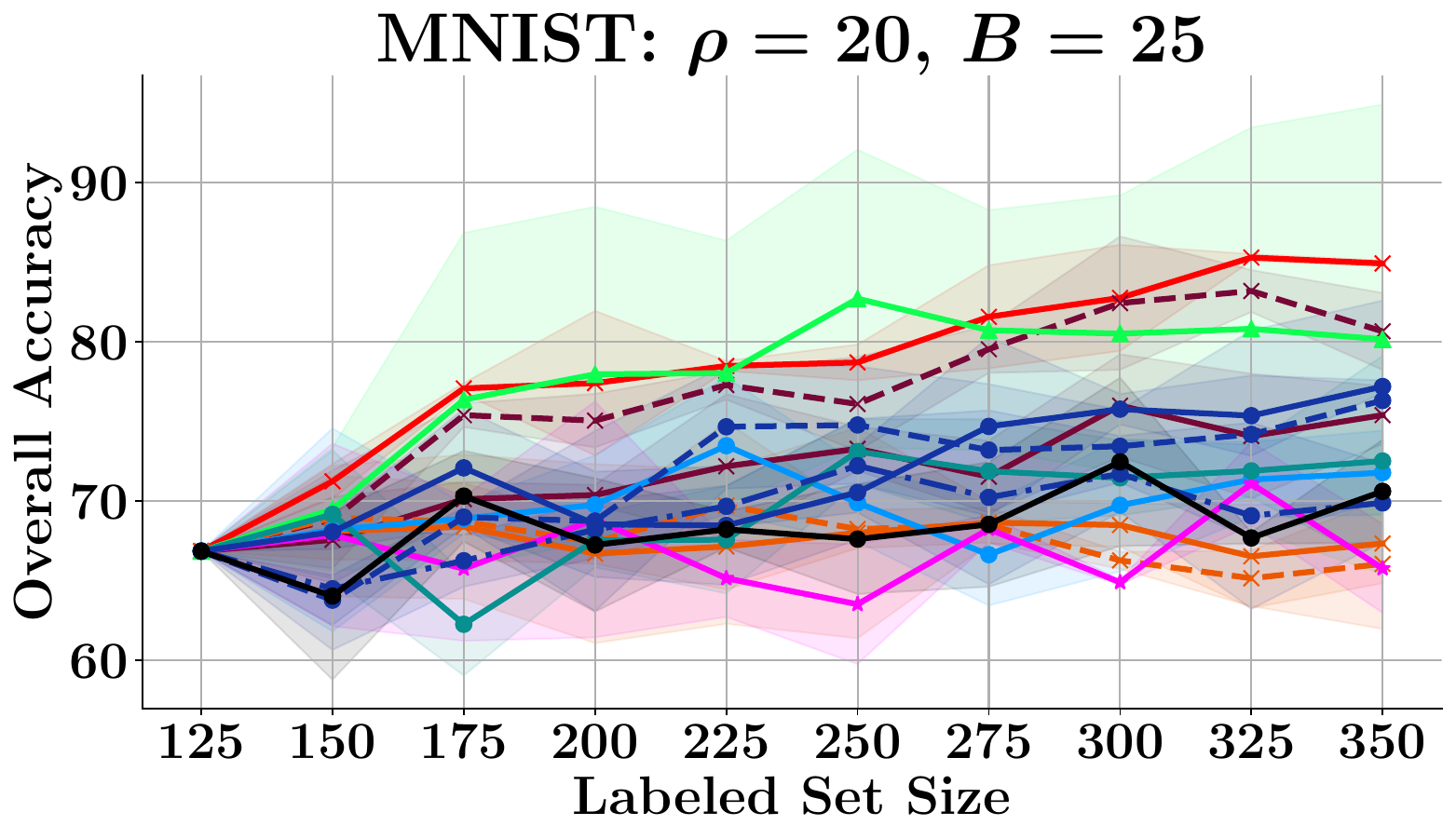}
\end{subfigure} 
\begin{subfigure}[t]{0.325\textwidth}
\includegraphics[width = \textwidth]{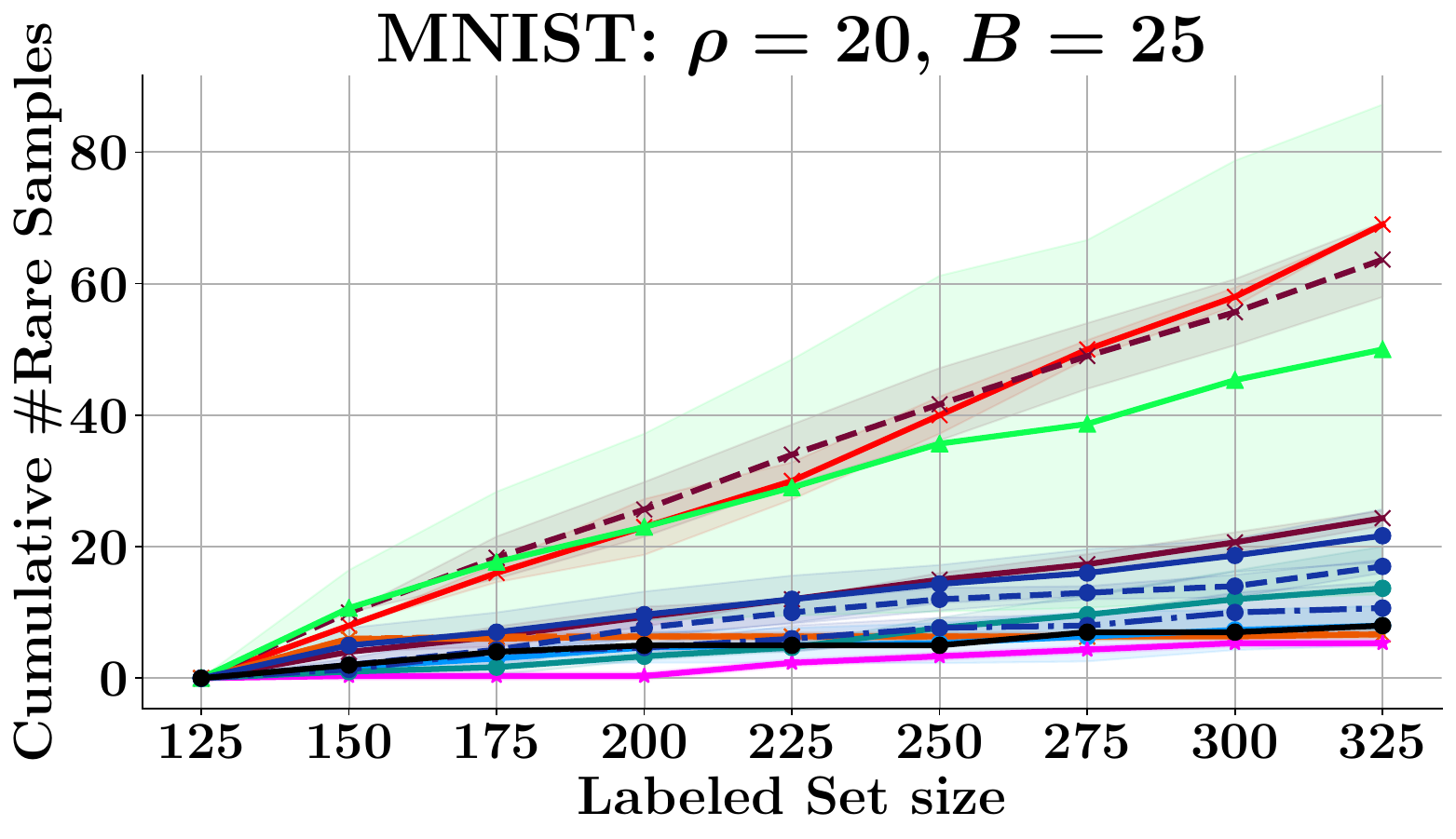}
\end{subfigure}
\begin{subfigure}[t]{0.325\textwidth}
\includegraphics[width = \textwidth]{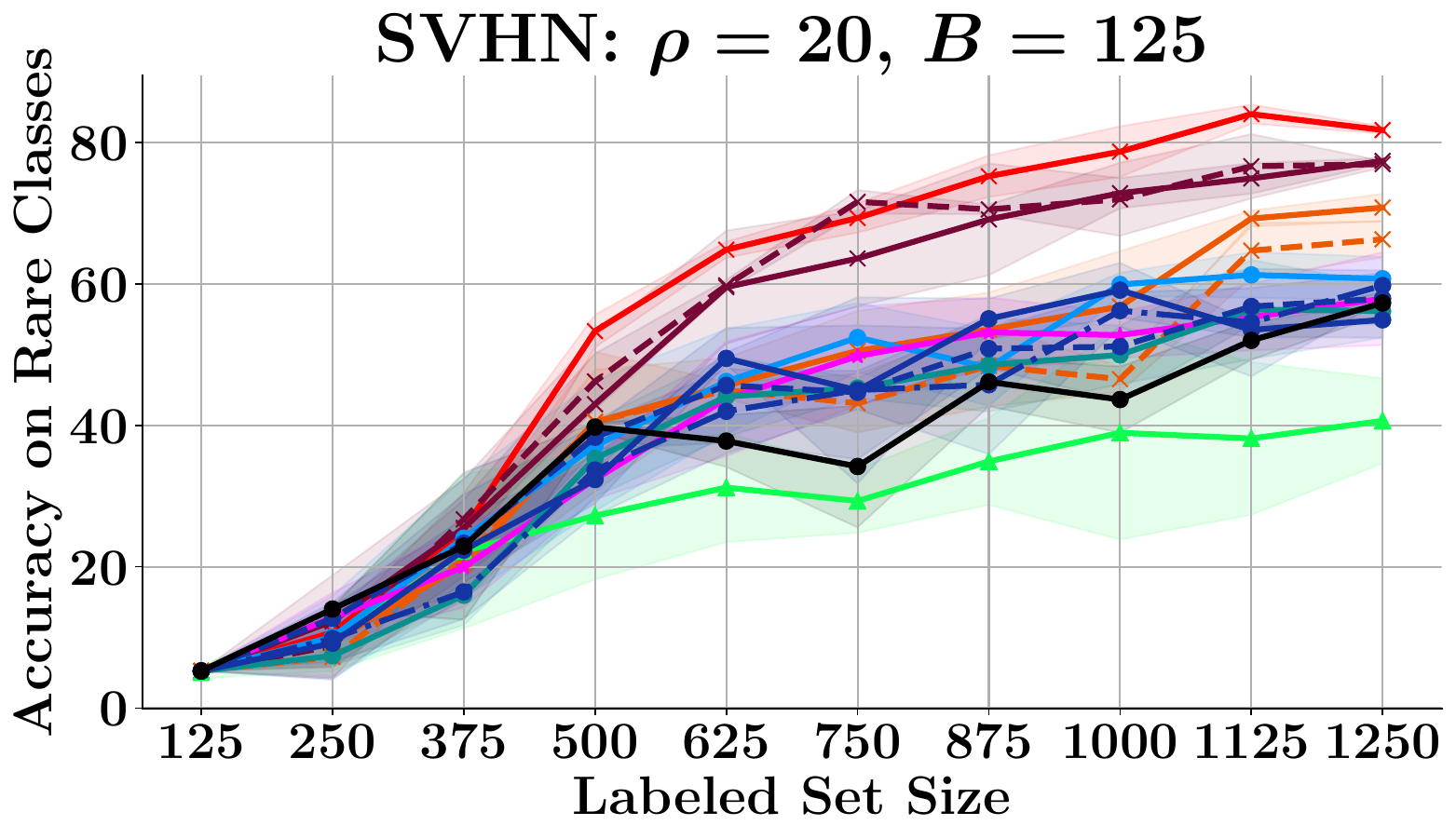}
\end{subfigure}
\begin{subfigure}[t]{0.325\textwidth}
\includegraphics[width = \textwidth]{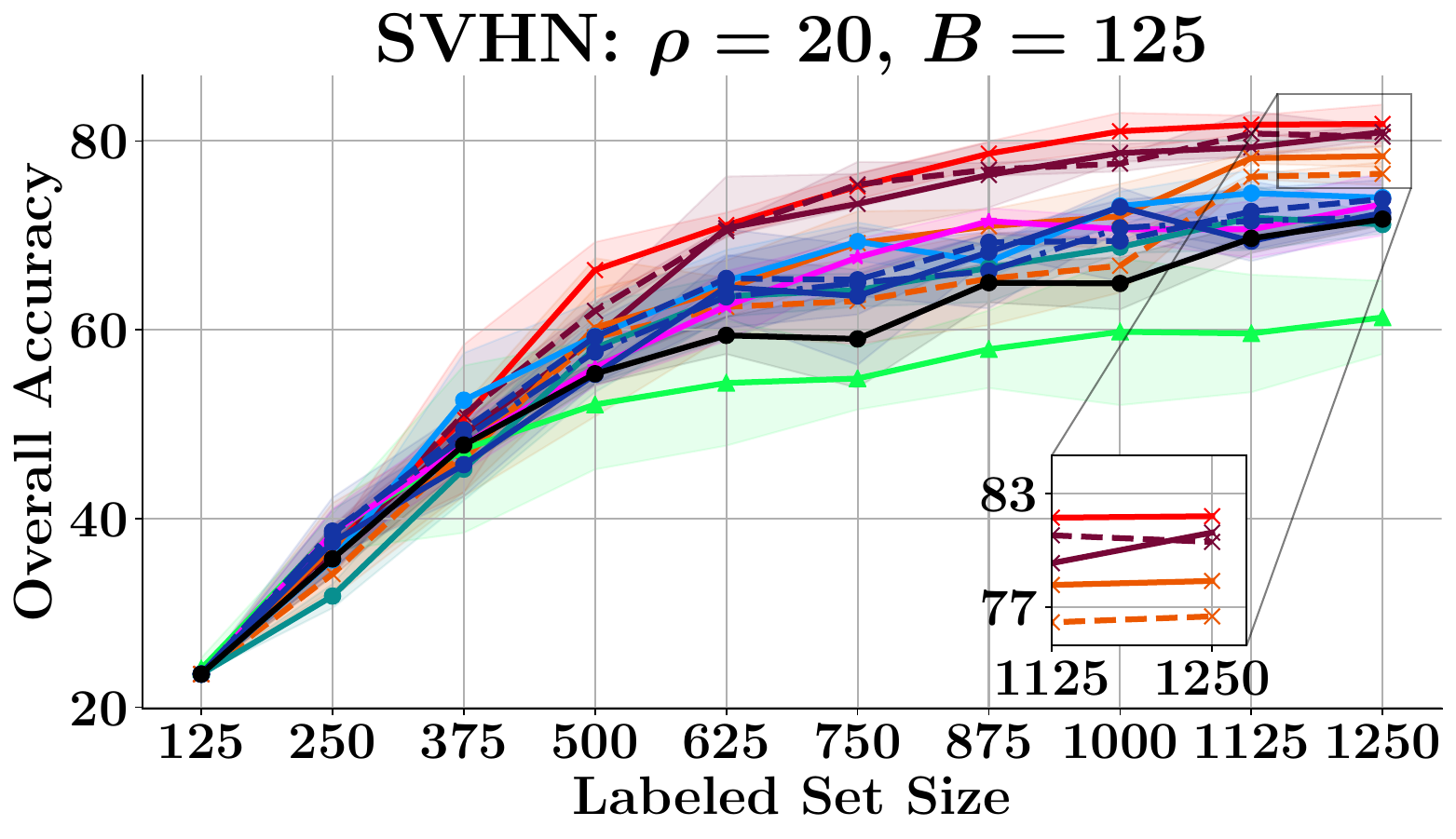}
\end{subfigure}
\begin{subfigure}[t]{0.325\textwidth}
\includegraphics[width = \textwidth]{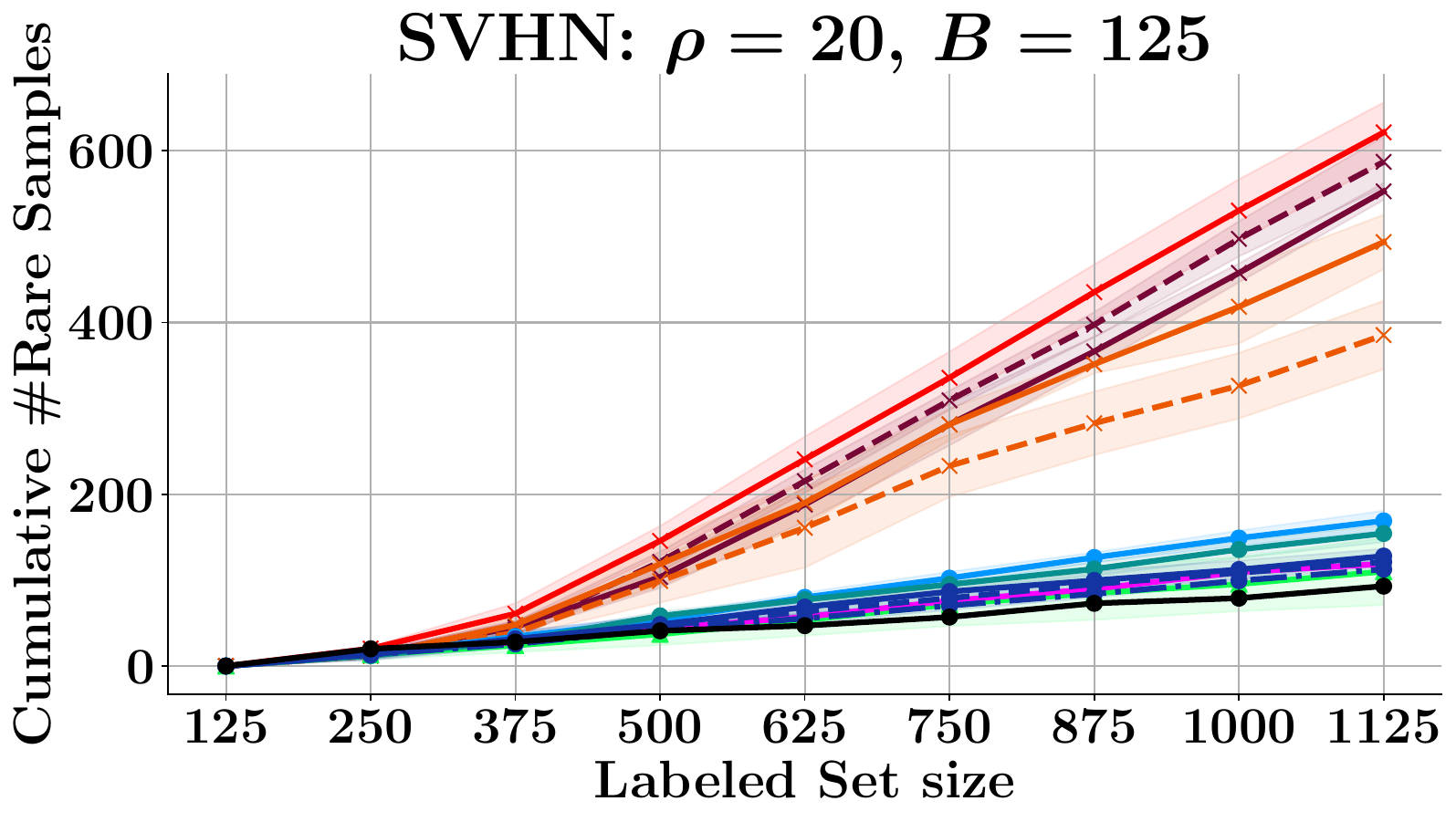}
\end{subfigure}
\caption{Additional experiments on MNIST and SVHN for active learning with rare classes. \textbf{Top row:} MNIST $\rho=100, B=25$, \textsc{Logdetmi} outperforms other methods even in extreme imbalance, with a large gap in accuracy, followed by \textsc{Flqmi}. \textbf{Middle row:} MNIST $\rho=20, B=25$, \textsc{Logdetmi} and \textsc{Flqmi} outperform all baselines in the later rounds of AL. \textbf{Bottom row:} SVHN $\rho=20, B=125$, All SMI methods significantly outperform other baselines.  
\vspace{-1ex}}
\label{fig:supp_res_rarecls}
\end{figure}

\begin{figure}[ht]
\centering
\includegraphics[width = 0.6\textwidth]{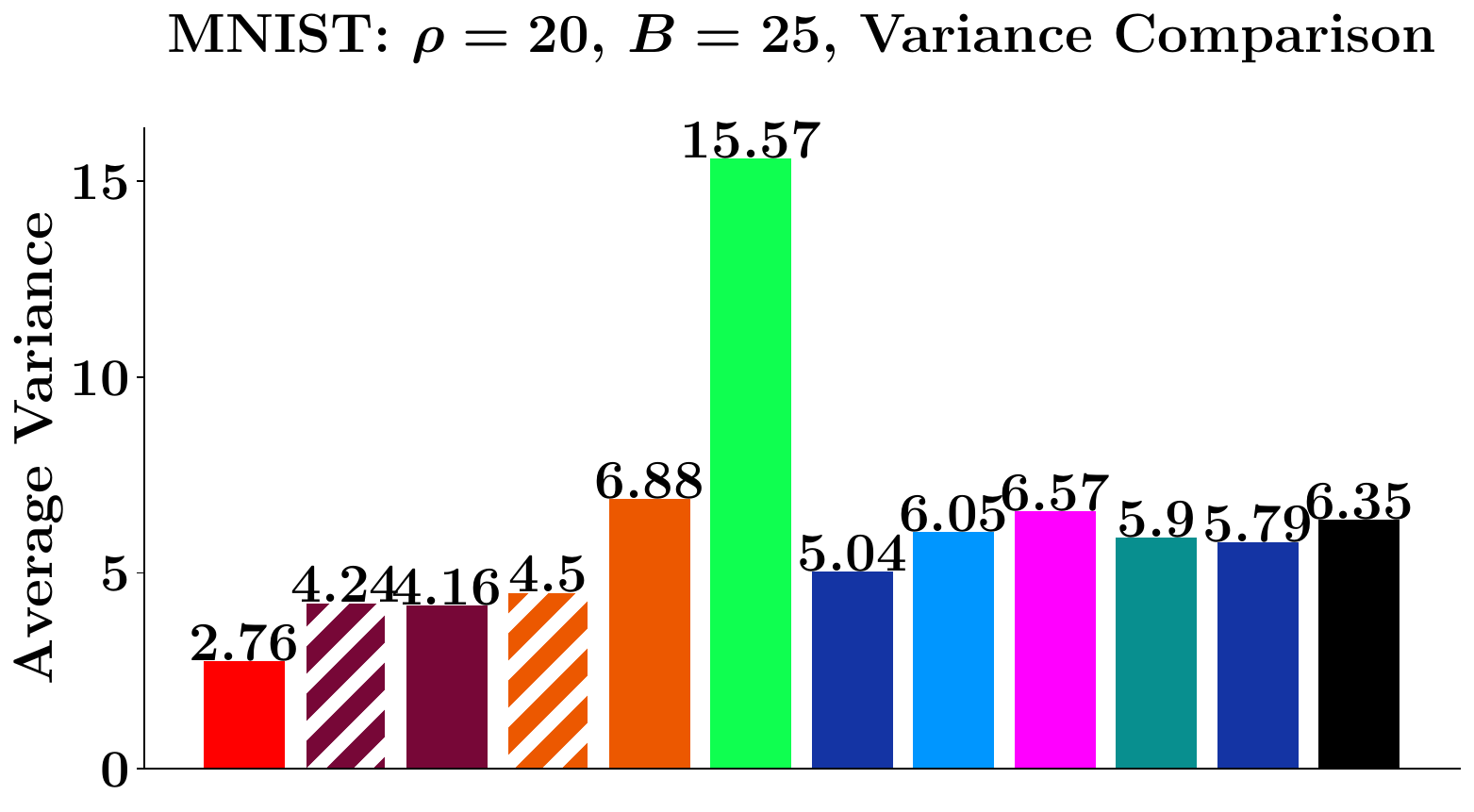}
\caption{Variance comparison on the rare classes scenario for MNIST $\rho=20, B=25$ (Middle row in \figref{fig:supp_res_rarecls}). \textsc{Fisher} has $\approx 5 \times$ variance in comparison with the SMI methods. The figure shares the same legend as \figref{fig:supp_res_rarecls}.
}
\label{fig:supp_res_rarecls_var}
\end{figure}

\begin{figure}[!ht]
\centering
\includegraphics[width =0.7\textwidth]{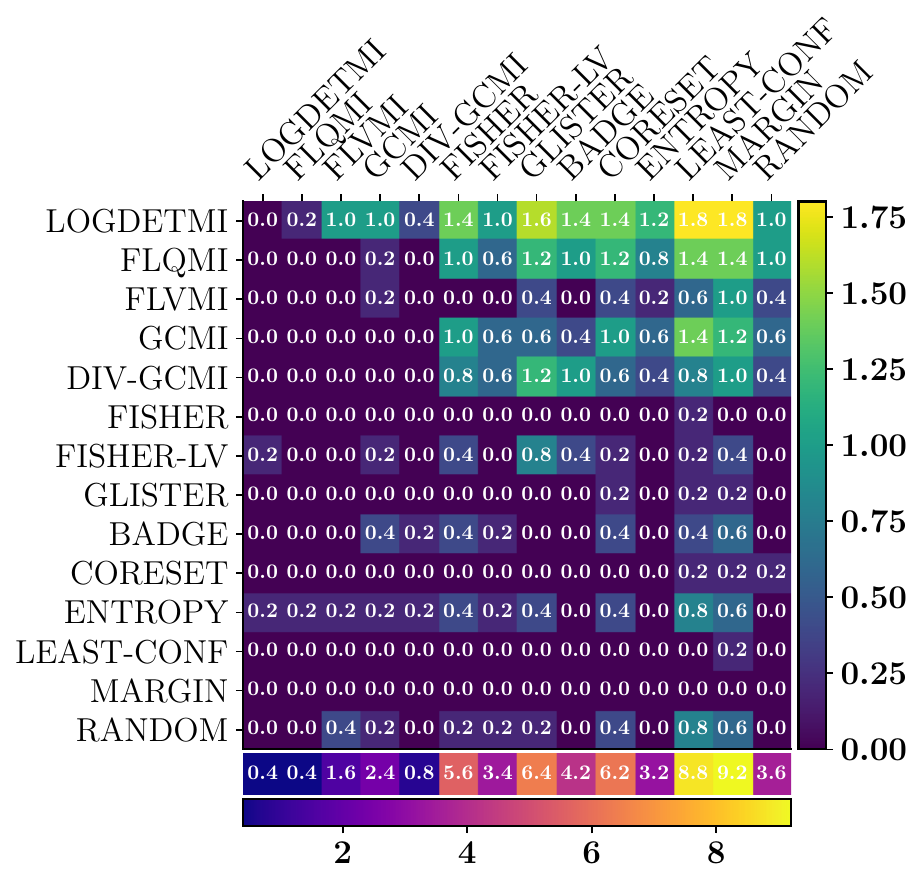}
\includegraphics[width = 0.7\textwidth]{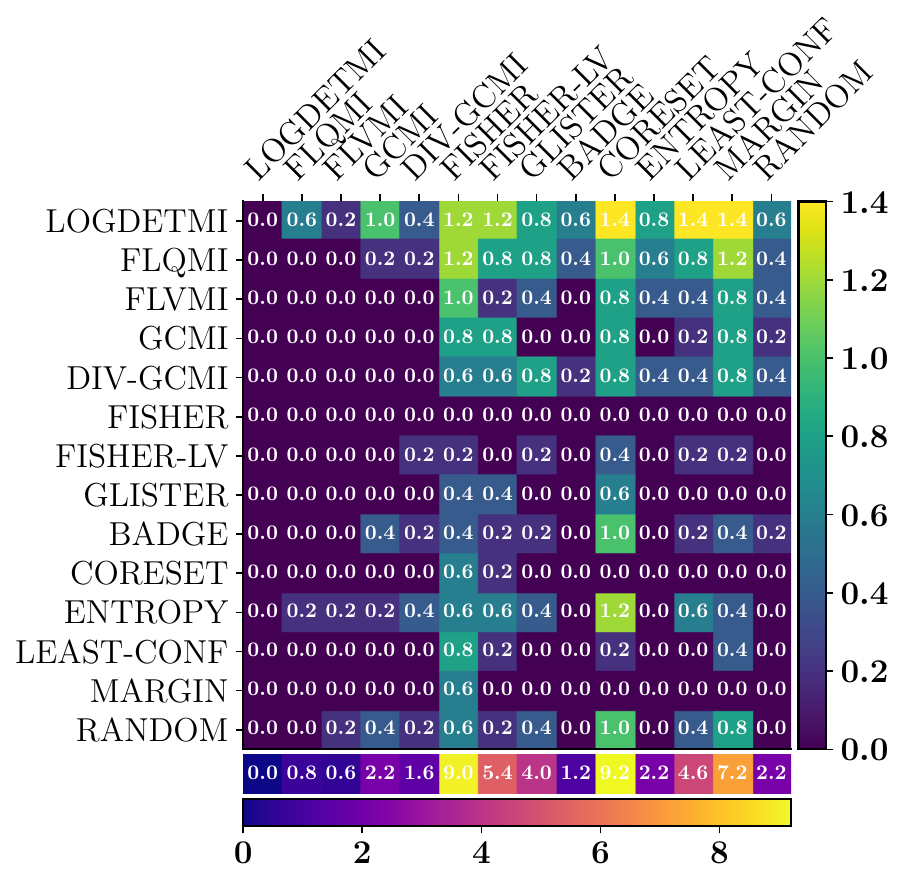}
\caption{Penalty Matrix comparing the average accuracy of rare classes (\textbf{top}) and overall accuracy (\textbf{bottom}) of different AL approaches in the class imbalance scenario. We observe that the SMI functions have a much lower column sum compared to other approaches.
}
\label{pen-matrix}
\end{figure}

\begin{figure*}
\centering
\includegraphics[width = 14cm, height=1cm]{results/cifar10/cifar10_redundancy_legend.pdf}
\centering
\hspace*{-0.6cm}
\begin{subfigure}[t]{0.49\textwidth}
\includegraphics[width = \textwidth]{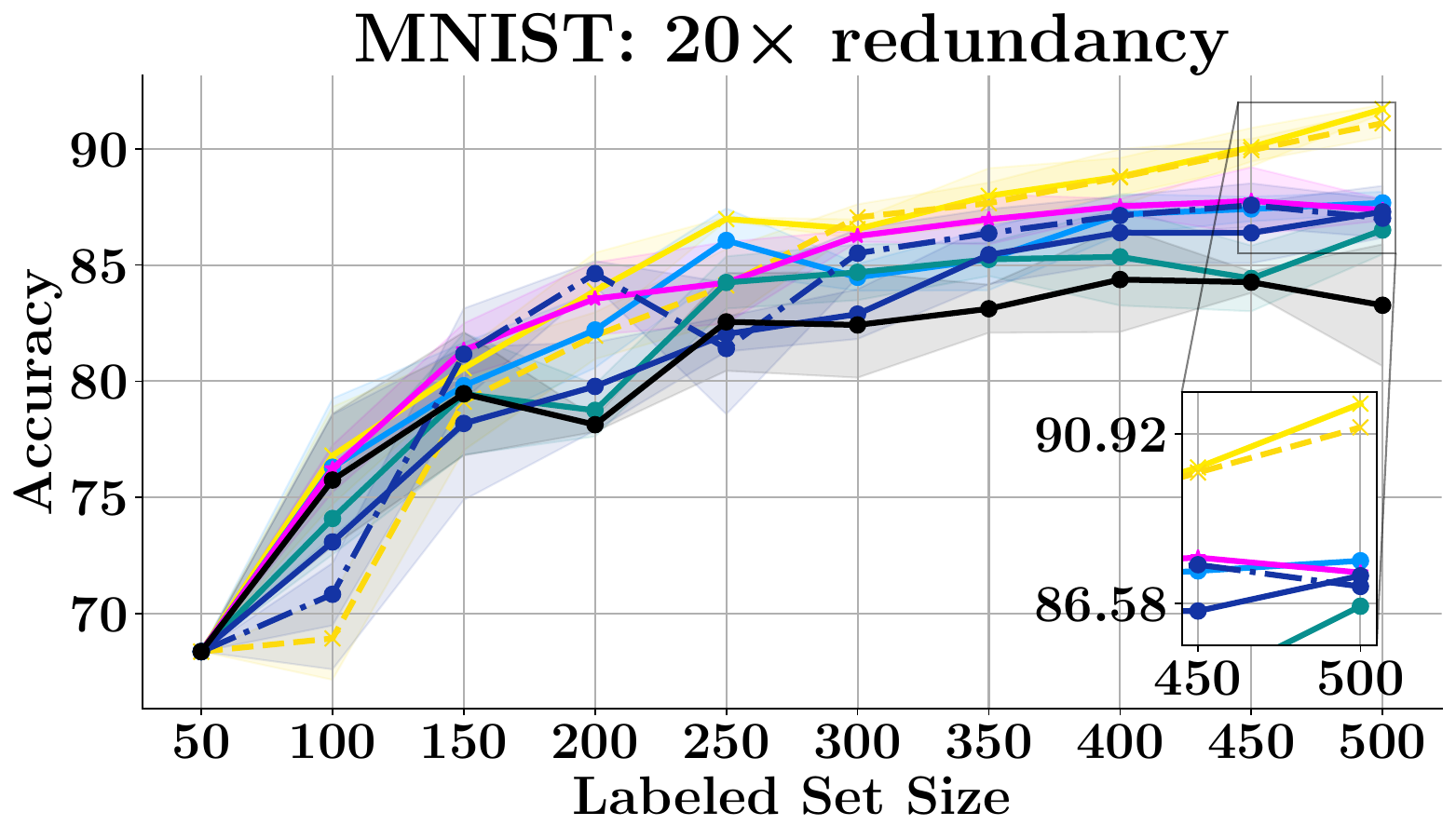}
\end{subfigure}
\begin{subfigure}[t]{0.49\textwidth}
\includegraphics[width = \textwidth]{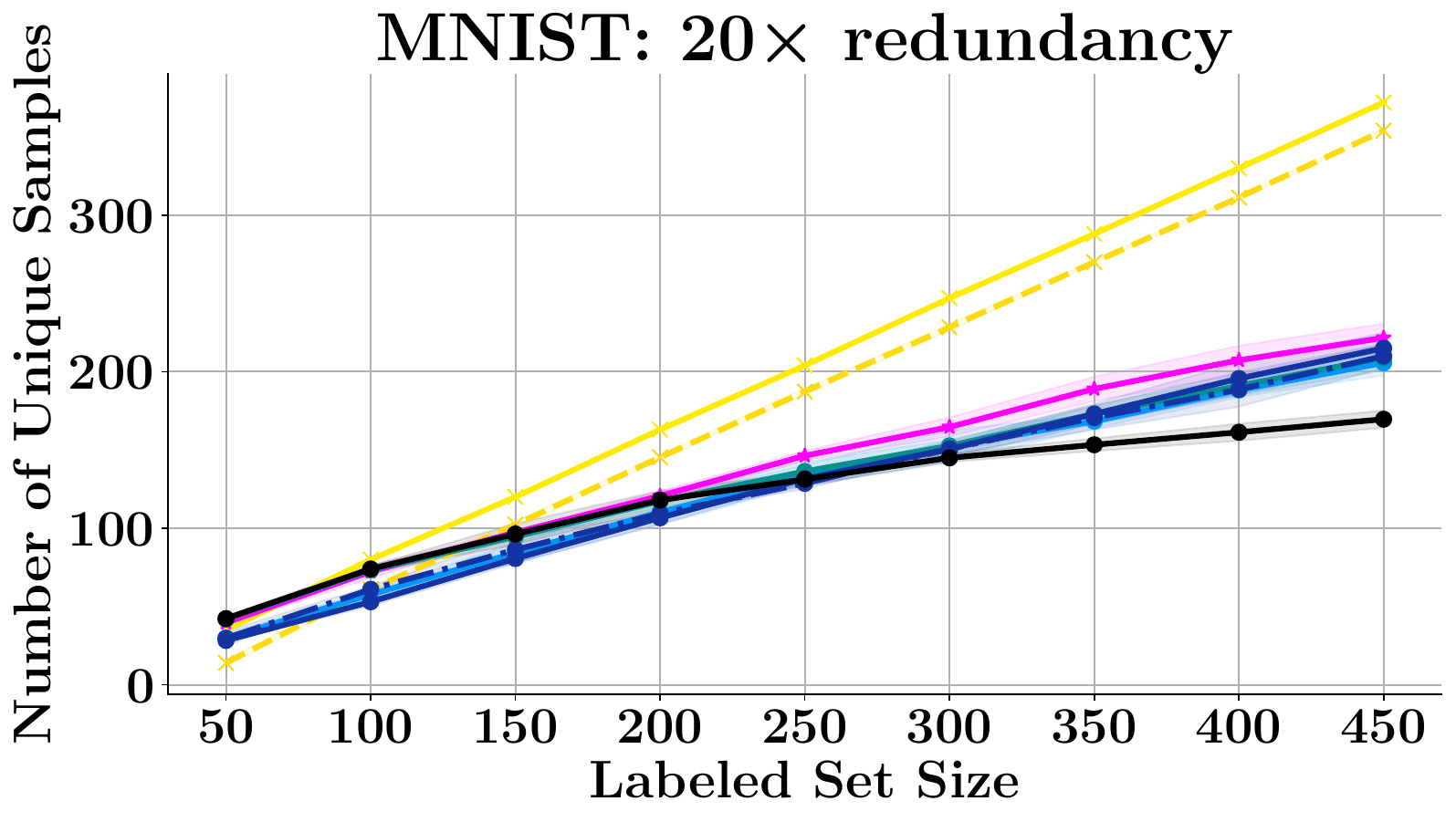}
\end{subfigure}
\begin{subfigure}[t]{0.49\textwidth}
\includegraphics[width = \textwidth]{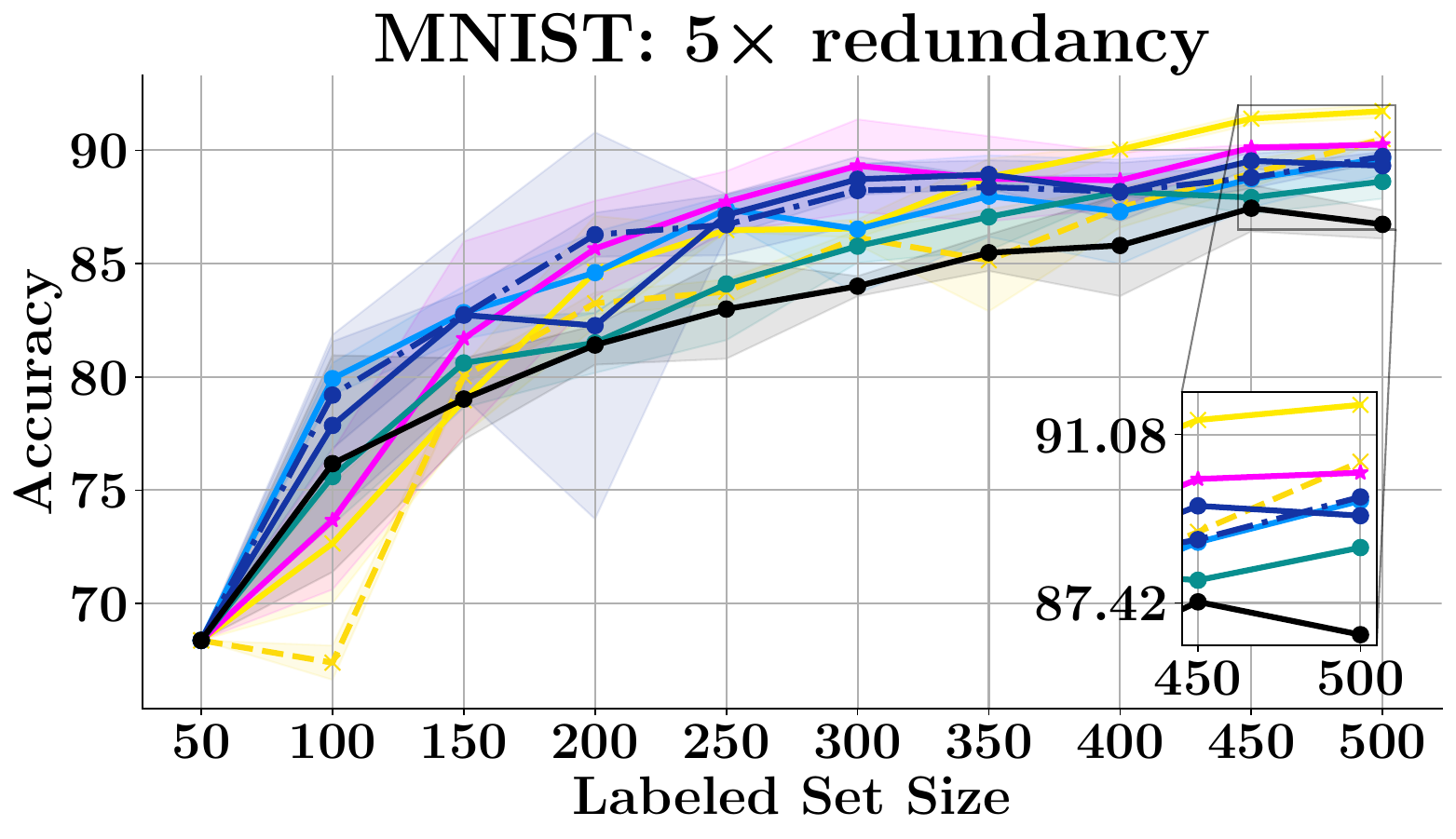}
\end{subfigure}
\begin{subfigure}[t]{0.49\textwidth}
\includegraphics[width = \textwidth]{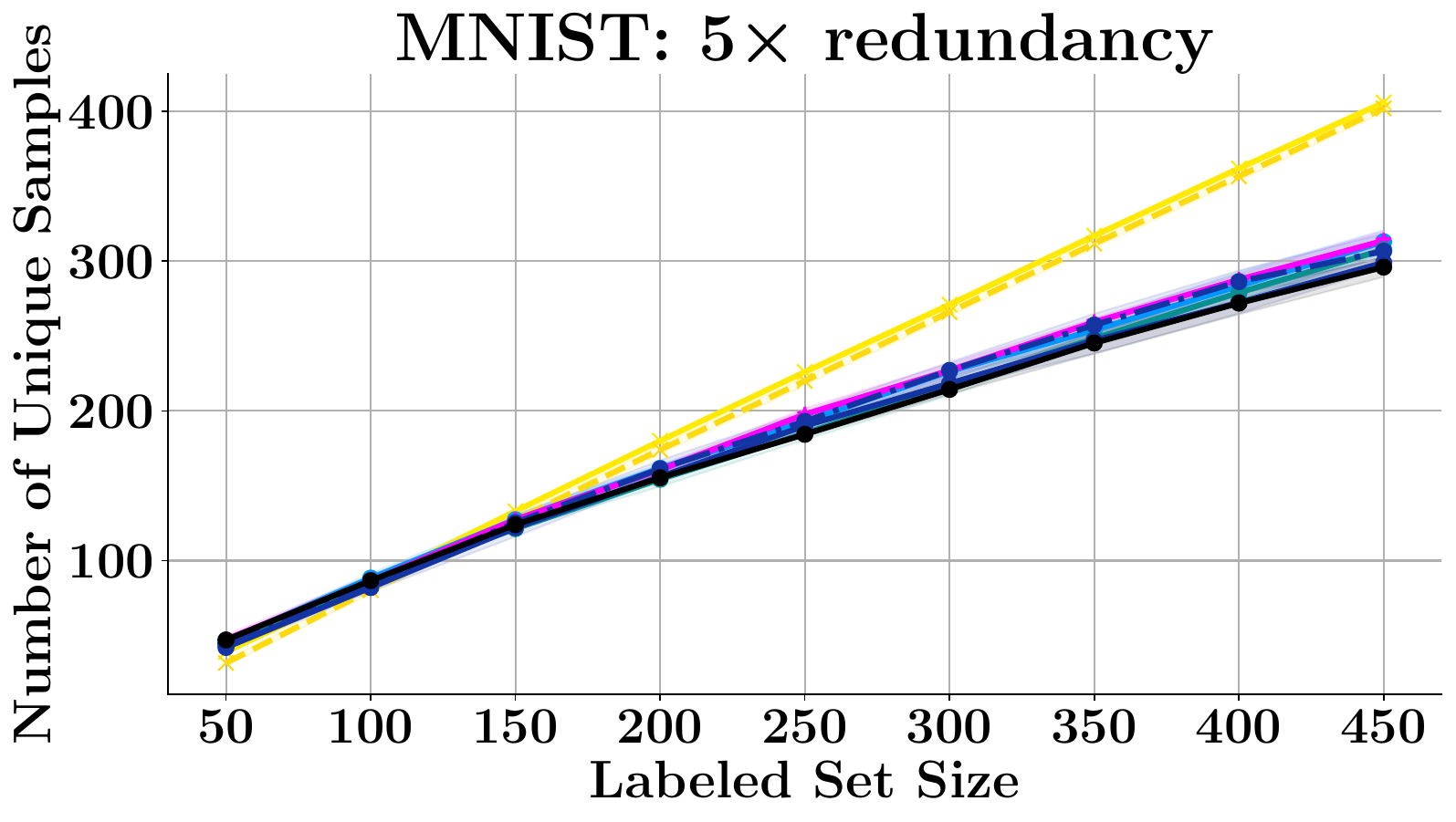}
\end{subfigure}
\begin{subfigure}[t]{0.49\textwidth}
\includegraphics[width = \textwidth]{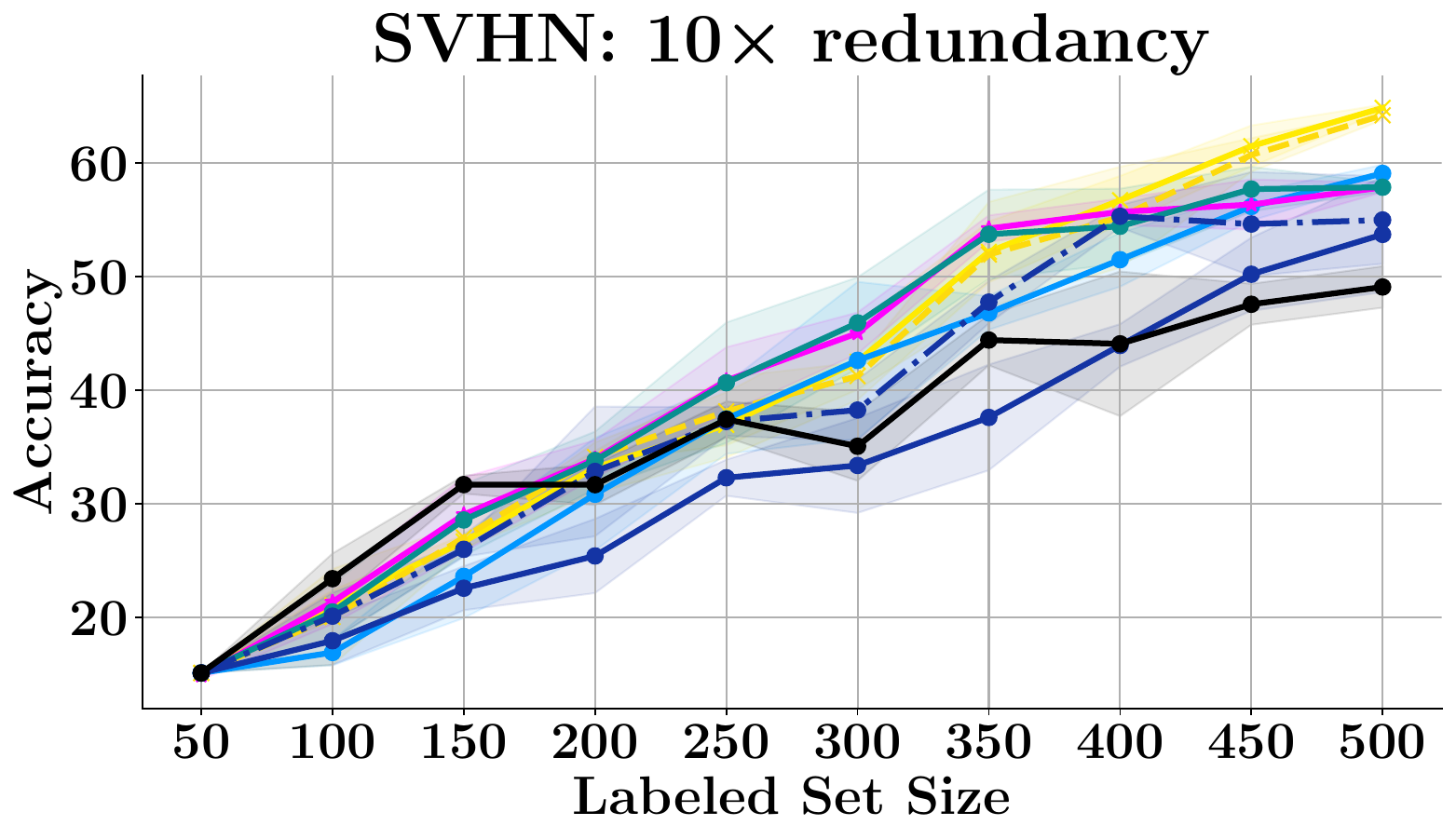}
\end{subfigure}
\begin{subfigure}[t]{0.49\textwidth}
\includegraphics[width = \textwidth]{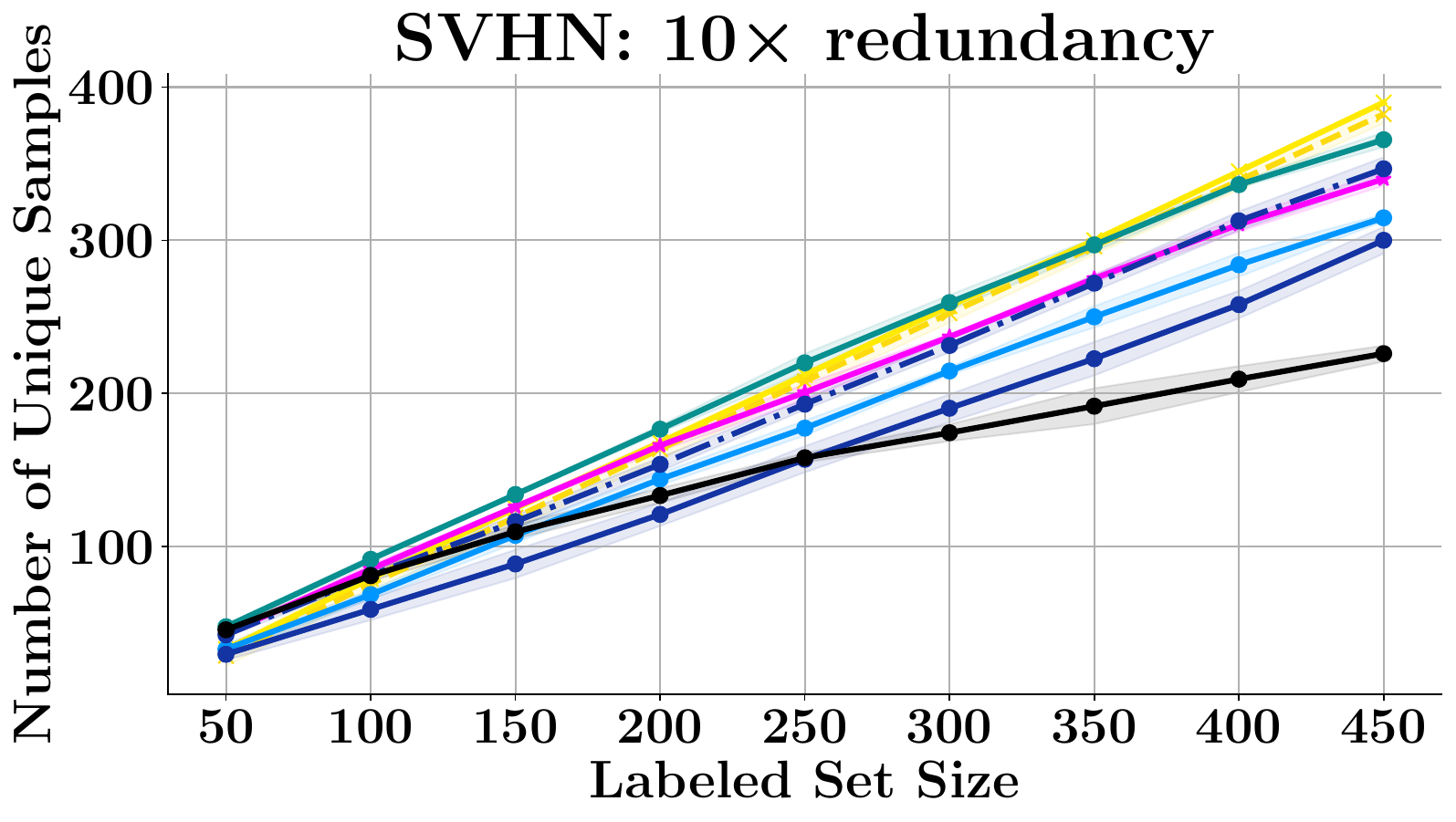}
\end{subfigure}

\caption{Active Learning under $20\times$ redundancy (\textbf{top row}) and $5\times$ redundancy (\textbf{middle row}) on MNIST. \textbf{Bottom row:}  $10 \times$ redundancy on SVHN. The CG functions (\textsc{Logdetcg, Flcg}) pick more unique points and outperform existing algorithms including \textsc{Badge}.
\vspace{-3ex}}
\label{fig:supp_res_redundancy}
\end{figure*}

\begin{figure}[!ht]
\centering
\includegraphics[width = 0.7\textwidth]{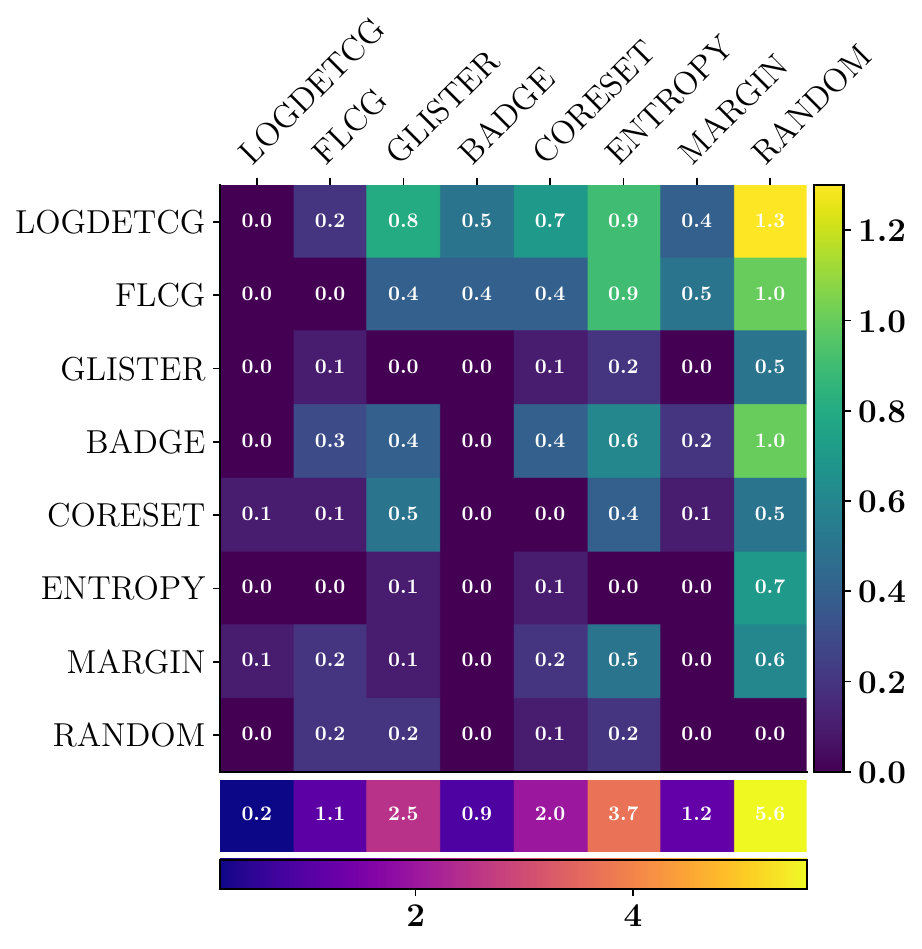}
\caption{Penalty Matrix comparing the different AL approaches in the redundancy scenario. We observe that the SCG functions have a much lower column sum compared to other approaches. 
}
\label{fig:res_redundancy_pen}
\end{figure}

\begin{figure}[!ht]
\centering
\includegraphics[width = 0.7\textwidth]{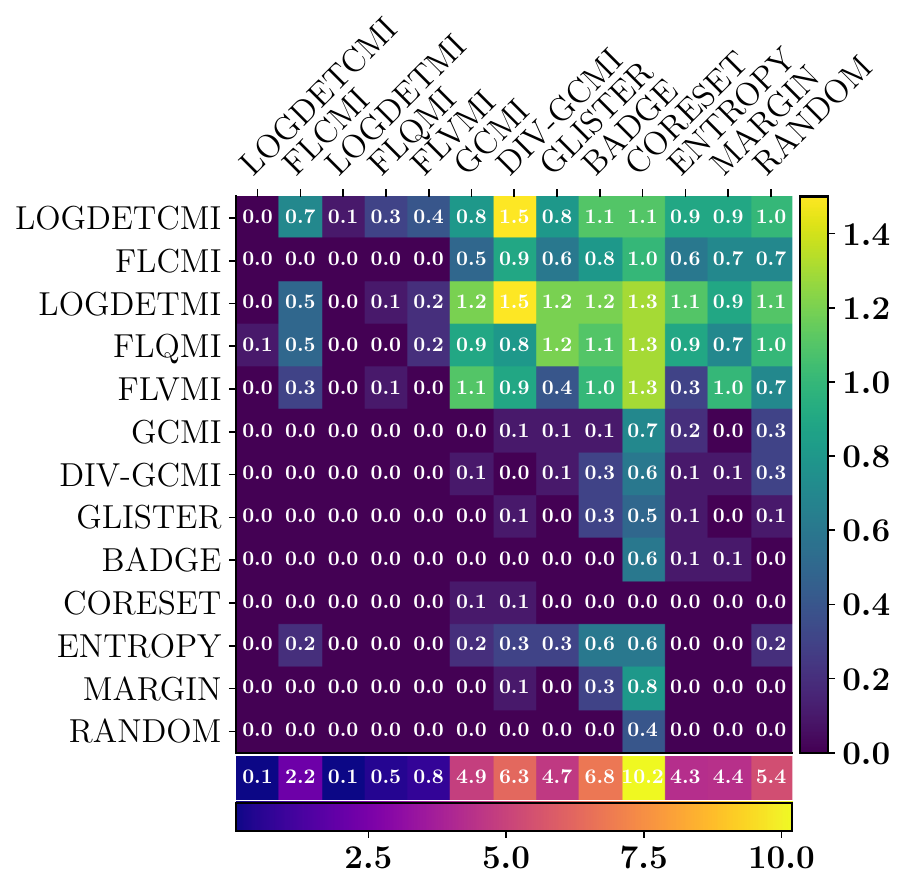}
\caption{Penalty Matrix comparing the different AL approaches in the OOD Scenario. We observe that the SCMI and SMI functions have a much lower column sum compared to other approaches.}
\label{fig:ood_penmat}
\end{figure}

\section{Results with Standard Active Learning}\label{app:standard_al}

In Figure~\ref{standard-al}, we compare the performance of the SFs on standard AL -- i.e., without redundancy, out-of-distribution data, and imbalance. The basic idea here is that we compute the similarity kernels using the gradients of the model (Algorithm~\ref{algo:unifiedAL}) and use just the submodular function -- i.e., setting $\Qcal = \Ucal, \Pcal = \emptyset$. In this work, we use the log-determinant and the facility location functions. We make the following observations: \textbf{1)} Log-determinant functions perform comparable to \textsc{Badge} and entropy sampling, particularly in the beginning. \textbf{2)} The facility location function does not perform as well in the standard AL setting, implying that diversity tends to play a more important role in standard active learning compared to representation.

\section{Additional Experiments and Takeaways for Active Learning with Rare Classes}\label{app:exp_rarecls}


In Figure~\ref{fig:supp_res_rarecls}, we show additional results for MNIST and SVHN for active learning with rare classes. The top row shows the results for the extreme imbalance scenario, i.e., $\rho = 100, B = 25$ (small batch size and extreme imbalance). We observe that \textsc{Logdetmi} significantly outperforms all other techniques, and \textsc{Flqmi} and \textsc{Fisher} come next. Note that the \textsc{Fisher} baseline~\cite{gudovskiy2020deep} was originally presented in this extreme imbalance scenario. The middle row in Figure~\ref{fig:supp_res_rarecls} contains results for $\rho = 20, B = 25$. This is similar to the results presented in the main paper but using a much smaller batch size. Here, \textsc{LogdetMI} and \textsc{Flqmi} again outperform the other baselines. While the average performance of the \textsc{Fisher} baseline~\cite{gudovskiy2020deep} is comparable to \textsc{LogdetMI} and \textsc{Flqmi}, it has a much higher variance compared to others (Figure~\ref{fig:supp_res_rarecls_var}). Finally, the bottom row shows the performance of the different techniques on SVHN. Again, we see that \textsc{LogdetMI} and \textsc{Flqmi} outperform all other techniques.  

\textbf{Takeaways from the Results:} The following are the main takeaways of the experiments in this section and the main paper:
\begin{itemize}
    \item Among the different MI functions, \textsc{LogdetmI} and \textsc{Flqmi} outperform all other MI functions. They also mostly outperform the Fisher Kernel baseline which was also designed for dealing with rare classes~\cite{gudovskiy2020deep}.
    \item \textsc{Logdetmi} particularly outperforms every other method in the high imbalance regime (100x imbalance). This is mainly because it is able to select the highest number of points from the rare classes (top row, right most plot in Figure~\ref{fig:supp_res_rarecls}.
    \item The \textsc{Fisher} baseline also can have a high variance, particularly when the batch size is high.
    \item For a fair comparison, we used a very small validation set in all our experiments. As compared in the main paper, \textsc{Fisher} performance does improve when we use a larger validation set, but doing so is not realistic.
    \item \textsc{Flqmi} is more scalable compared to \textsc{Logdetmi} and other kernel-based approaches; hence, it is the desired choice of approach for very large datasets.
\end{itemize}

\textbf{Penalty Matrix: } Figures~\ref{pen-matrix} shows the penalty matrix results on the rare class accuracy (top) and overall accuracy (bottom).  We see that \textsc{Logdetmi} and \textsc{Flqmi} have the smallest column sum, which indicates that most other baselines are not statistically significantly better than them. Furthermore, they also have the highest row sum (followed by some of the other MI functions), which indicates that they are statistically significantly better than other approaches. These matrices are obtained by combining the results on MNIST and CIFAR-10 for $\rho = 20, B = 125$ (i.e., the results in the main paper).

\section{Additional Experiments and Takeaways from Active Learning with Redundancy}\label{app:exp_redundancy}
In the main paper, we show the results on CIFAR-10 and MNIST with $10\times$ redundancy. In this section, we also add results for $5\times$ and $15\times$ redundancy for MNIST. The results are in Figure~\ref{fig:supp_res_redundancy}. Furthermore, we also run experiments on SVHN (bottom row) with $10\times$ redundancy. The following are the takeaways of the results:

\begin{itemize}
    \item The CG functions (\textsc{Logdetcg} and \textsc{Flcg}) significantly outperform other baselines including \textsc{Badge}, particularly after a few rounds of AL and towards the end. In particular, there is a improvement of 3\% to 5\% using the CG functions compared to \textsc{Badge} and other baselines with a labeled set size of 500. 
    \item The main reason for this is that the CG functions pick more unique points compared to the other techniques.
    \item Amongst the two CG functions, we see that \textsc{Logdetcg} performs better than \textsc{Flcg}. 
    \item From the pairwise penalty matrix in Figure~\ref{fig:res_redundancy_pen}, we see that \textsc{Logdetcg} has the lowest column sum and has the highest row sum, which indicates that it statistically significantly outperforms other techniques. In terms of the row sum, \textsc{Logdetcg} is followed by \textsc{flcg} and \textsc{Badge}.
\end{itemize}


\section{Additional Experiments and Takeaways for Active Learning with OOD Data}\label{app:exp_ood}
In the case of active learning with OOD data, we additionally add the penalty matrix (figure~\ref{fig:ood_penmat}). The following are the main observations and takeaways:
\begin{itemize}
    \item Figure~\ref{fig:ood_penmat} shows the results of the penalty matrix with the different CMI functions. We observe that \textsc{Logdetcmi} has the smallest column sum along with \textsc{Logdetmi}. 
    \item However, as shown in the main paper, the CMI functions have the smallest variance and are hence more stable compared to the SMI variants. Furthermore, the CMI functions generally outperform the SMI counterparts at later rounds.
    \item However, the SMI functions are often comparable (particularly \textsc{Logdetmi} and \textsc{Flqmi}) and hence are a good choice for OOD data as well. 
\end{itemize}

\section{Societal Impacts and Limitations}\label{app:imp_lim}
\textbf{Limitations of this work: } The first limitation of this work is that the MI functions are all graph-based functions. With the exception of \textsc{Flqmi}, all functions have quadratic complexity. The partitioning trick will help, but that comes at the cost of performance. We would like to explore more classes of MI functions (feature-based functions~\cite{wei2015submodularity} in particular) in future work. Secondly, the MI functions depend on good choices of features. In this work, we use gradients which tend to work very well since they inherently also capture uncertainty~\cite{ash2019deep}. However, the approaches do not perform as well in the early stages, which could be mitigated by the use better features, e.g., self-supervised and unsupervised representations~\cite{gudovskiy2020deep}.

\textbf{Societal Impacts: } Negative societal impacts of this work include using \textsc{Similar} to mine through large datasets to perpetuate and amplify certain biases in the data. On the flip side, this work can also have a positive impact through its use for fair active learning, where certain under-represented and minority slices or classes can be improved upon by applying it in the rare class and rare slice experiment setting (\secref{sec:rareClsAL}). We would like to explore the use of \textsc{Similar} in applications like improving the performance of biased slices based on race; for example, we would like to improve inference performance on underrepresented Asian woman using \textsc{Similar} for tasks like face recognition, gender recognition, and age recognition. Importantly, recent work has shown that commercial facial recognition and age/gender classification engines perform poorly on these rare slices~\cite{buolamwini2018gender}. A number of recent papers have been proposed to generate such fair face datasets~\cite{karkkainen2021fairface}, but creating such datasets can take a lot of manual effort to mine the rare slices. We propose to use and study \textsc{Similar} for such scenarios in future work.

\section{Experiments on Real-world Medical Dataset} \label{app:MedicalDatasetRes}
\begin{figure*}
\centering
\includegraphics[width = 14cm, height=1cm]{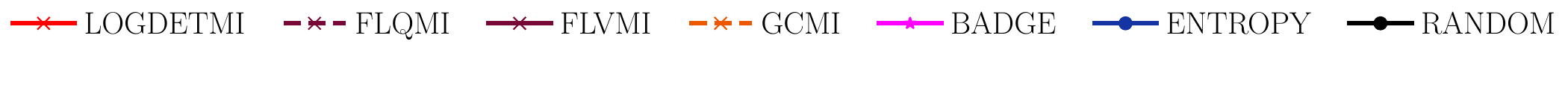}
\centering
\hspace*{-0.6cm}
\begin{subfigure}[t]{0.49\textwidth}
\includegraphics[width = \textwidth]{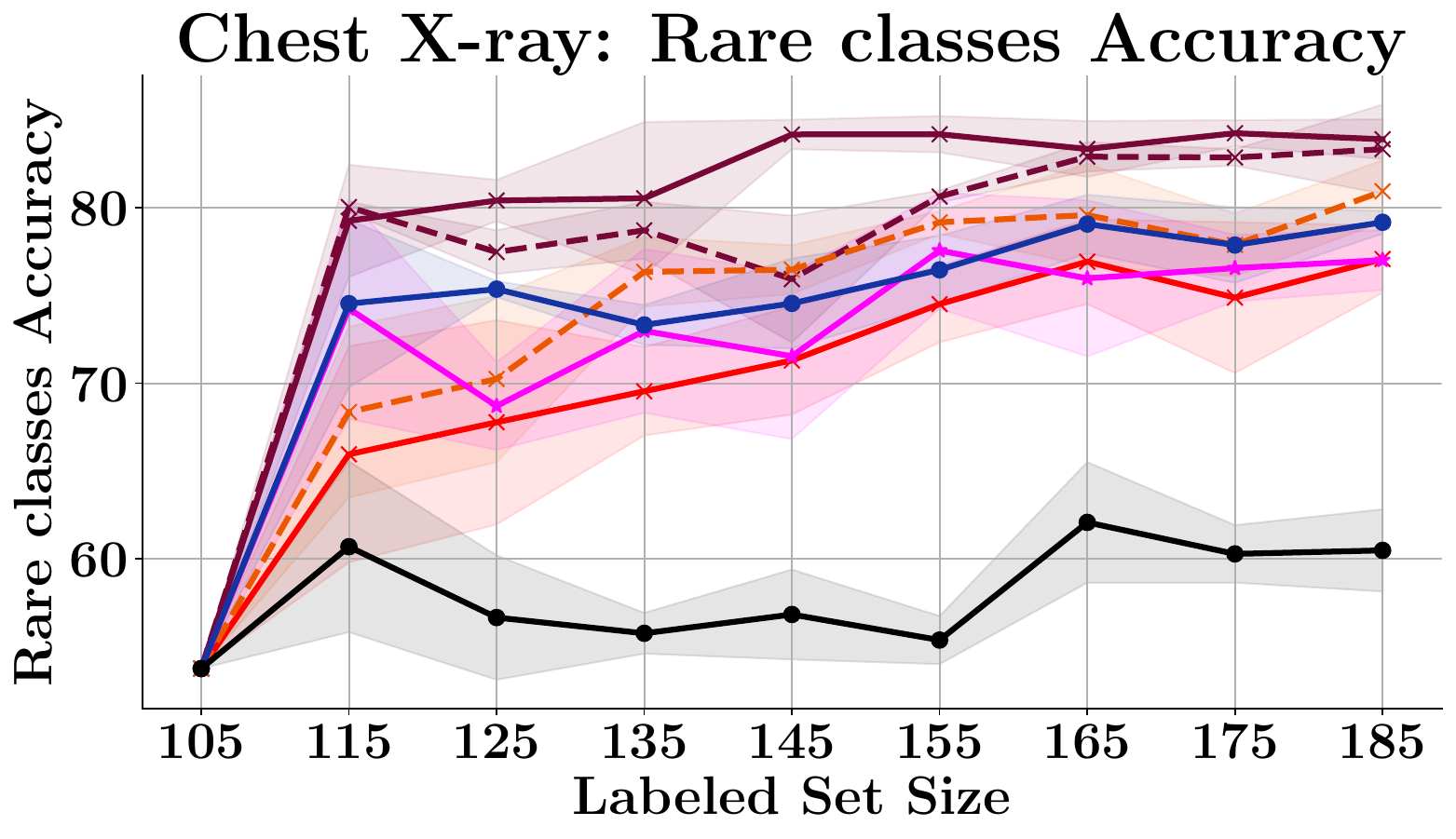}
\end{subfigure}
\begin{subfigure}[t]{0.49\textwidth}
\includegraphics[width = \textwidth]{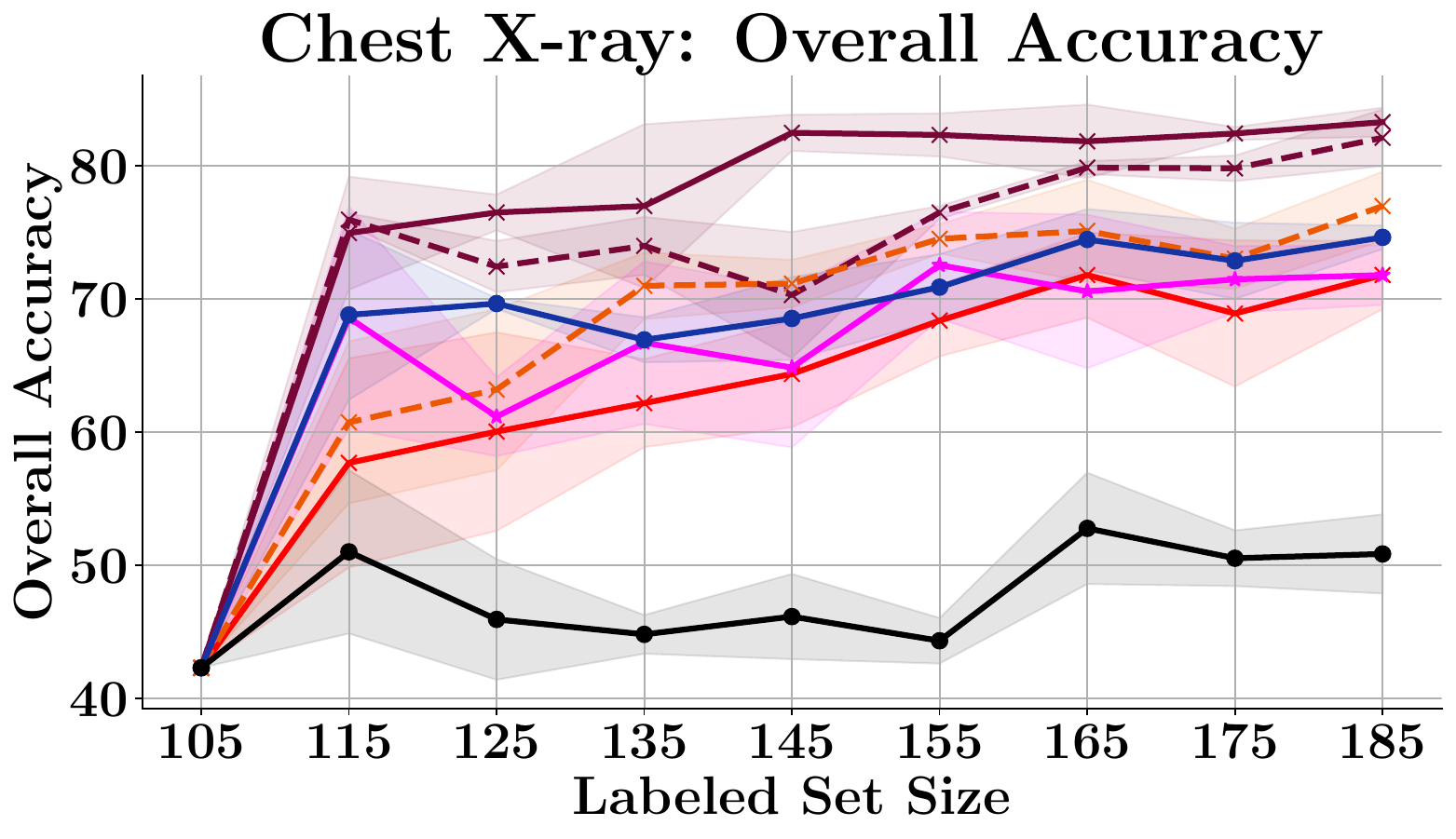}
\end{subfigure}
\caption{Active Learning on real-world medical image classification. The \textsc{Smi} functions outperform the baselines by $\approx 10\% - 12\%$ on the rare classes accuracy and $\approx 8\% - 10\%$ on the overall accuracy.
\vspace{-3ex}}
\label{fig:pneumonia_mnist}
\end{figure*}

In this section, we apply our framework to Pneumonia-MNIST (pediatric chest X-ray) medical image classification dataset. The goal is to classify X-ray images into 'pneumonia' and 'benign'. As done in \secref{sec:expRareCls} and to simulate a real-world scenario, we use an imbalance factor $\rho=20$, such that the 'pneumonia' class is a rare class. We use $|\Ccal_\Lcal| + |\Dcal_\Lcal| = 105$, $|\Ccal_\Ucal| + |\Dcal_\Ucal| = 1100$, $B=10$ (AL batch size) and, $|\Rcal| = 5$. On this dataset, we observed that using misclassified data points in $\Rcal$ is beneficial for acquiring subsets that lead to higher accuracy gains. We observe that the \textsc{Smi} functions outperform the baselines by $\approx 10\% - 12\%$ on the rare classes accuracy and $\approx 8\% - 10\%$ on the overall accuracy (see \figref{fig:pneumonia_mnist}). \looseness-1

\section{Experiments on Multiple Realistic Scenarios} \label{app:resMultipleScenarios}
\begin{figure}[!ht]
\centering
\includegraphics[width = 0.95\textwidth]{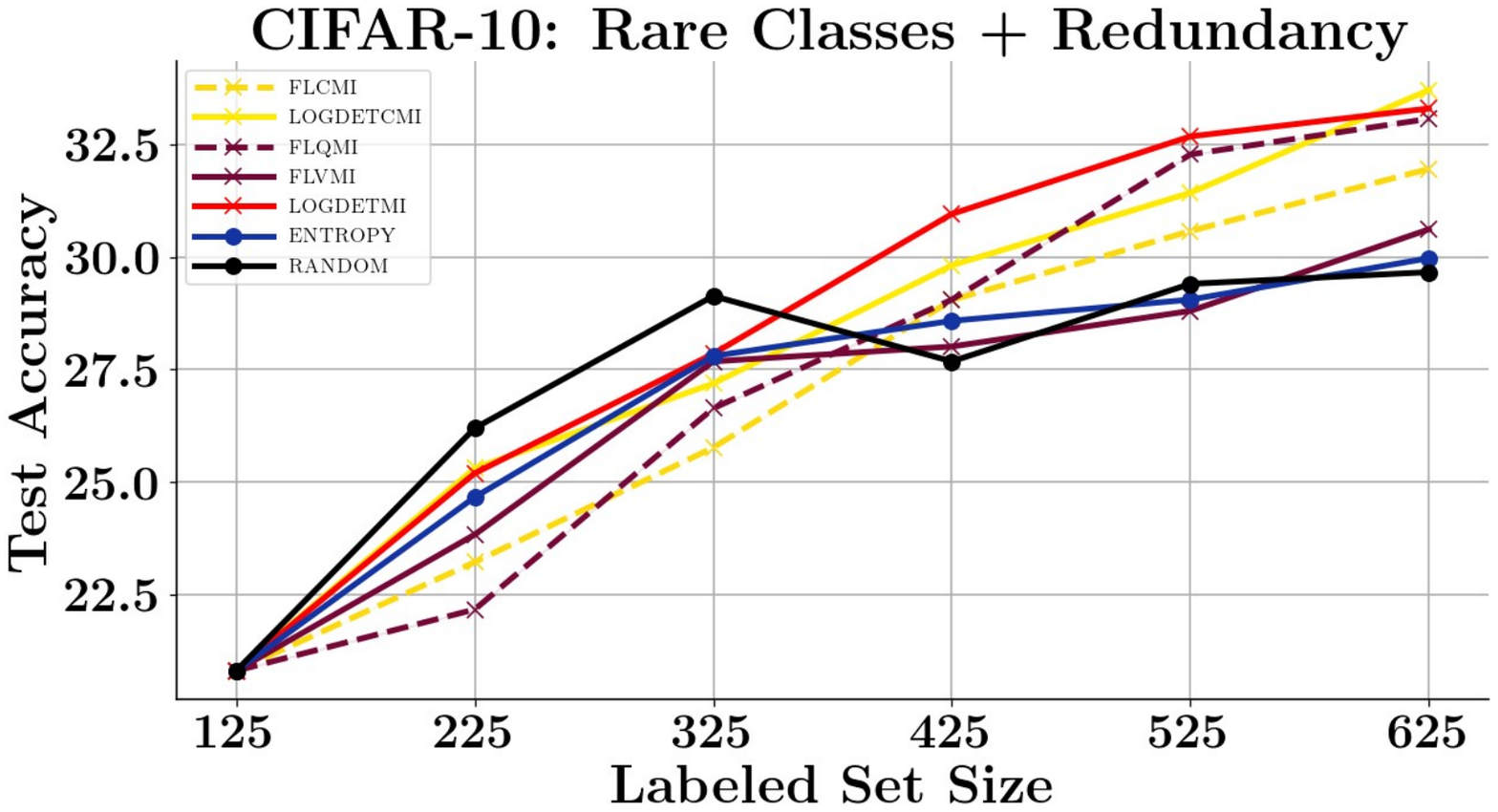}
\caption{Active learning in multiple realistic scenarios (Rare classes + Redundancy).  
}
\label{fig:multipleScenario}
\end{figure}

In this section, we apply our framework to a scenario where redundancy \emph{and} rare classes are co-occurring in the dataset. To do so, we first create an imbalance on CIFAR-10 in a similar fashion as done in \secref{sec:expRareCls}. We use $\rho=10, |\Ccal_\Lcal| + |\Dcal_\Lcal| = 125, |\Ccal_\Ucal| + |\Dcal_\Ucal| = 5.5K, B=100$ and repeat the unlabeled dataset $5\times$ to get $|\Ccal_\Ucal| + |\Dcal_\Ucal| = 27.5K$.
We observe that the SCMI and SMI functions perform better than the baselines (see \figref{fig:multipleScenario}).

\end{document}